\documentclass{article}

\PassOptionsToPackage{numbers, sort&compress}{natbib}

\usepackage[preprint]{neurips_2026}

\input{preamble}

\usepackage[utf8]{inputenc} \usepackage[T1]{fontenc}    \usepackage{hyperref}       \usepackage{url}            \usepackage{booktabs}       \usepackage{amsfonts}       \usepackage{nicefrac}       \usepackage{microtype}      \usepackage{xcolor}         

\title{\ours{}: Improved Surface Geometry in Point Maps}

\author{\textbf{Karim Knaebel$^1$\qquad
Gonzalo Martin Garcia$^1$\qquad
Christian Schmidt$^1$\qquad
Ilya Fradlin$^1$}\\[0.4em]
\textbf{Lucas Nunes$^1$\qquad
Daan de Geus$^2$\qquad
Bastian Leibe$^1$}\\[0.6em]
\small
$^1$RWTH Aachen University\qquad $^2$Eindhoven University of Technology\\
\small
\url{https://vision.rwth-aachen.de/surge}
}

\begin{document}

\maketitle

\begin{abstract}
Recent feedforward 3D reconstruction methods predict point maps and estimate global 3D geometry remarkably well.
    However, their predictions still exhibit inaccurate local surface geometry, which is clearly visible qualitatively but only weakly reflected in common metrics.
    To make these errors more explicit in evaluation, we introduce a point map normal metric that evaluates the local surface orientation induced by neighboring 3D predictions.
    To reduce these errors, we propose two complementary components: a point gradient matching loss that supervises depth-normalized 3D finite differences, and a Neighborhood Attention Decoder (NAD) that progressively upsamples features and uses Neighborhood Attention for local feature mixing.
    Across eight zero-shot monocular geometry benchmarks, our model, \ours{}, achieves the best average rank for global point map \AbsRel{} and consistently improves local point map and point map normal evaluations.
\end{abstract}

\newcommand{\teaserCarSplitTop}{0.66}
\newcommand{\teaserCarSplitBottom}{0.33}
\newcommand{\teaserUmicArrowStartX}{0.05}
\newcommand{\teaserUmicArrowStartY}{0.4}
\newcommand{\teaserUmicArrowEndX}{0.22}
\newcommand{\teaserUmicArrowEndY}{0.33}
\newcommand{\teaserUmicSecondArrowStartX}{0.85}
\newcommand{\teaserUmicSecondArrowStartY}{0.07}
\newcommand{\teaserUmicSecondArrowEndX}{0.9}
\newcommand{\teaserUmicSecondArrowEndY}{0.32}
\newcommand{\teaserIncludeCrop}[5]{\adjincludegraphics[
		width=\linewidth,
		trim={#1\width{} #2\height{} #3\width{} #4\height{}},
		clip
	]{#5}}
\newcommand{\teaserCropImage}[5]{\begin{tikzpicture}[baseline=(current bounding box.center)]
		\node[inner sep=0pt, outer sep=0pt] {\teaserIncludeCrop{#1}{#2}{#3}{#4}{#5}};
	\end{tikzpicture}}
\newcommand{\teaserDrawArrow}[4]{\draw[red, line width=1.25pt, -latex]
		($($(teaserImage.south west)!#1!(teaserImage.south east)$)!#2!($(teaserImage.north west)!#1!(teaserImage.north east)$)$) --
		($($(teaserImage.south west)!#3!(teaserImage.south east)$)!#4!($(teaserImage.north west)!#3!(teaserImage.north east)$)$);
}
\newcommand{\teaserArrowImage}[9]{\begin{tikzpicture}[baseline=(current bounding box.center)]
		\node[inner sep=0pt, outer sep=0pt] (teaserImage) {\teaserIncludeCrop{#1}{#2}{#3}{#4}{#5}};
		\teaserDrawArrow{#6}{#7}{#8}{#9}
	\end{tikzpicture}}
\newcommand{\teaserSplitImage}[8]{\begin{tikzpicture}[baseline=(current bounding box.center)]
		\node[inner sep=0pt, outer sep=0pt] (teaserImage) {\teaserIncludeCrop{#1}{#2}{#3}{#4}{#6}};
		\begin{scope}
			\clip
				(teaserImage.north west) --
				($(teaserImage.north west)!#7!(teaserImage.north east)$) --
				($(teaserImage.south west)!#8!(teaserImage.south east)$) --
				(teaserImage.south west) --
				cycle;
			\node[inner sep=0pt, outer sep=0pt] at (teaserImage.center) {\teaserIncludeCrop{#1}{#2}{#3}{#4}{#5}};
		\end{scope}
	\end{tikzpicture}}

\begin{figure}[h]
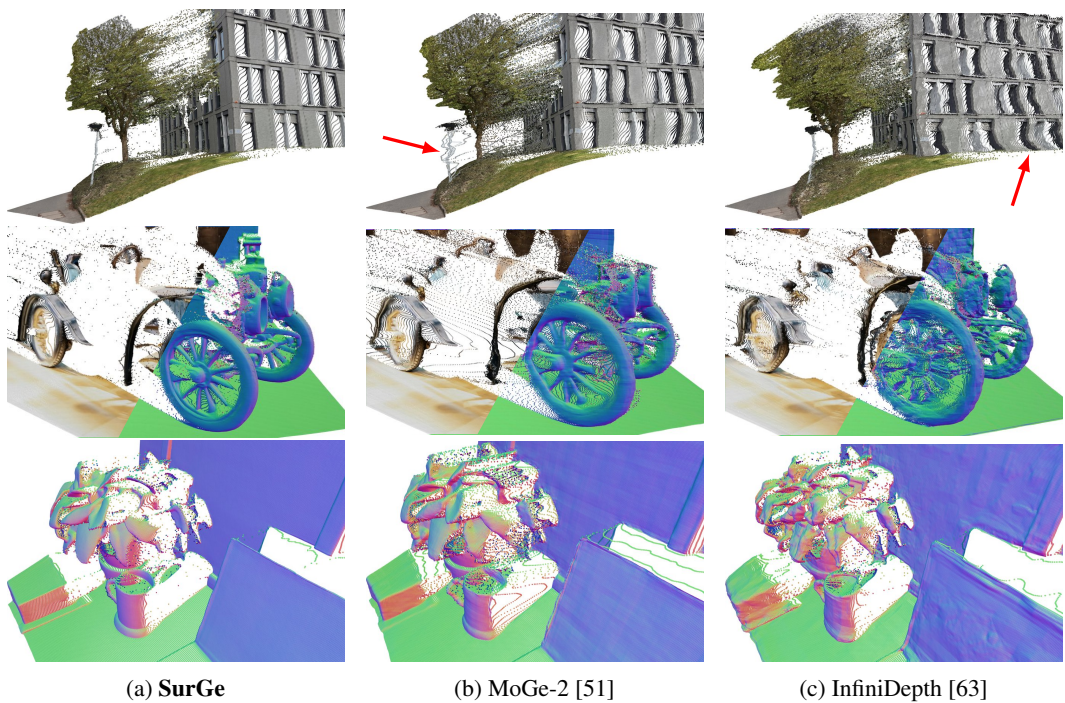

	\centering

	\begin{subfigure}{0.32\linewidth}
		\centering
		\teaserCropImage {0.000000}{0.000000}{0.116750}{0.156154}
			{img/teaser/umic/NAD/points}
\end{subfigure}
	\hfill
	\begin{subfigure}{0.32\linewidth}
		\centering
		\teaserArrowImage {0.000000}{0.000000}{0.116750}{0.161012}
			{img/teaser/umic/moge2/points}
			{\teaserUmicArrowStartX}
			{\teaserUmicArrowStartY}
			{\teaserUmicArrowEndX}
			{\teaserUmicArrowEndY}
\end{subfigure}
	\hfill
	\begin{subfigure}{0.32\linewidth}
		\centering
		\teaserArrowImage {0.000000}{0.000000}{0.220527}{0.245789}
			{img/teaser/umic/infinidepth/points}
			{\teaserUmicSecondArrowStartX}
			{\teaserUmicSecondArrowStartY}
			{\teaserUmicSecondArrowEndX}
			{\teaserUmicSecondArrowEndY}
\end{subfigure}

	\begin{subfigure}{0.32\linewidth}
		\centering
		\teaserSplitImage {0.337277}{0.000000}{0.025944}{0.331797}
			{img/teaser/car2/NAD/points}
			{img/teaser/car2/NAD/point_normals}
			{\teaserCarSplitTop}
			{\teaserCarSplitBottom}
\end{subfigure}
	\hfill
	\begin{subfigure}{0.32\linewidth}
		\centering
		\teaserSplitImage {0.285388}{0.026194}{0.025944}{0.270676}
			{img/teaser/car2/moge2/points}
			{img/teaser/car2/moge2/point_normals}
			{\teaserCarSplitTop}
			{\teaserCarSplitBottom}
\end{subfigure}
	\hfill
	\begin{subfigure}{0.32\linewidth}
		\centering
		\teaserSplitImage {0.285388}{0.026194}{0.025944}{0.270676}
			{img/teaser/car2/infinidepth/points}
			{img/teaser/car2/infinidepth/point_normals}
			{\teaserCarSplitTop}
			{\teaserCarSplitBottom}
\end{subfigure}

	\begin{subfigure}{0.32\linewidth}
		\centering
		\teaserCropImage {0.000000}{0.215554}{0.155666}{0.067361}
			{img/teaser/cup/NAD/point_normals}
		\caption{\textbf{\ours{}}}
	\end{subfigure}
	\hfill
	\begin{subfigure}{0.32\linewidth}
		\centering
		\teaserCropImage {0.000000}{0.234192}{0.155666}{0.094943}
			{img/teaser/cup/moge2/point_normals}
		\caption{MoGe-2~\cite{wang2025moge2}}
	\end{subfigure}
	\hfill
	\begin{subfigure}{0.32\linewidth}
		\centering
		\teaserCropImage {0.000000}{0.236142}{0.155666}{0.095733}
			{img/teaser/cup/infinidepth/point_normals}
		\caption{InfiniDepth~\cite{yu2026infinidepth}}
	\end{subfigure}

	\caption{
        \textbf{Qualitative state-of-the-art comparison.}
        \ours{} predicts noticeably cleaner point maps.
}\label{fig:teaser}
\end{figure}

\section{Introduction}
\label{sec:intro}

Monocular geometry estimation seeks to recover dense 3D scene structure from a single image.
Recent feedforward models~\cite{wang2025moge,wang2025moge2,piccinelli2025unidepthv2} predict a point map, assigning each pixel a 3D point.
This representation was initially proposed for two-view geometry by DUSt3R~\cite{wang2024dust3r}; it was then brought to monocular geometry by MoGe~\cite{wang2025moge}, and was extended to the many-view setting by VGGT~\cite{wang2025vggt}.
Since then, the area has developed rapidly, with a growing number of follow-up models building on the same general paradigm~\cite{keetha2026mapanything,wang2026pi3,lin2026depthanything3}.
While these models perform strongly overall and generalize well to in-the-wild images, their predictions still contain visible local 3D artifacts despite having plausible coarse scene geometry (\cref{fig:teaser}).
In particular, small deviations in the relative positions of nearby predicted points can distort local shape and orientation.
We refer to the relative 3D structure of nearby points as \emph{local surface geometry} and focus on improving it in this work.

This weakness is especially visible on thin structures.
Thin foreground elements such as streetlamps, street signs, chair legs, or faucets often emerge bent or oscillatory in 3D, and the distortion typically becomes more severe as the structure gets thinner and the background is farther away.
We find that this is not a 2D edge-localization problem but a 3D shape problem: the point prediction can look plausible when viewed as a depth map while neighboring 3D points form inconsistent surface patches.

\begin{figure}
    \centering
\resizebox{\linewidth}{!}{
    \begin{tikzpicture}[
        image/.style={
inner sep=0pt,
            outer sep=1pt,
},
        label/.style={
            anchor=south west,
            outer sep=2pt,
inner sep=2pt,
            color=black,
            fill=white,
            opacity=0.8,
            text opacity=1,
            rounded corners=3pt,
text height=1.5ex,
            text depth=0.5ex,
        },
        column label/.style={
            anchor=south,
            text height=1.5ex,
            text depth=0.5ex,
        },
        pics/splitImage/.style 2 args={code={\node[image] (-img1) at (0, 0) {#1};\begin{scope}\clip ($(-img1.north west)!0.333!(-img1.north east)$)
    				-- (-img1.north east)
    				-- (-img1.south east)
    				-- ($(-img1.south west)!0.667!(-img1.south east)$)
    				-- cycle;\node[image] (-img2) at (0, 0) {#2};\end{scope}}},
        box/.style={
            fill=white,
            draw=black,
            rounded corners=2pt,
            font=\footnotesize,
            align=left,
            inner sep=4pt,
        }
    ]
        \newcommand{\imgWidth}{4.5cm}
        \newcommand{\imgSpacing}{8pt}
        \node[image] (pts-gt) {\includegraphics[width=\imgWidth]{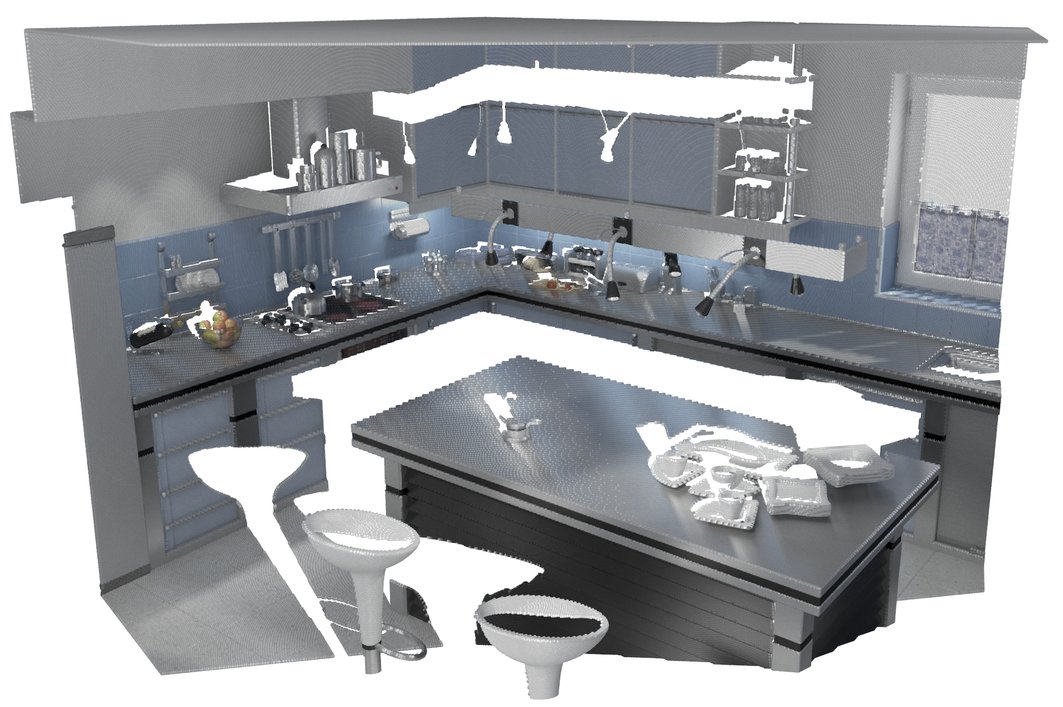}};
        \node[image, anchor=west] (pts-low) at ($(pts-gt.east)+(\imgSpacing,0)$) {\includegraphics[width=\imgWidth]{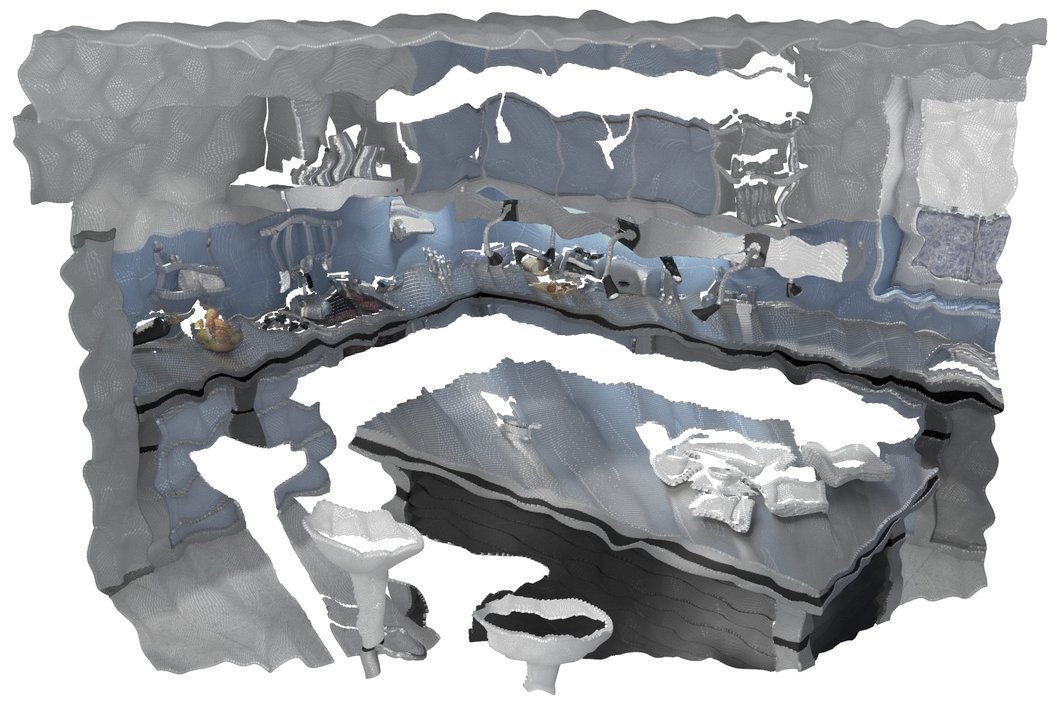}};
        \node[image, anchor=west] (pts-high) at ($(pts-low.east)+(\imgSpacing,0)$) {\includegraphics[width=\imgWidth]{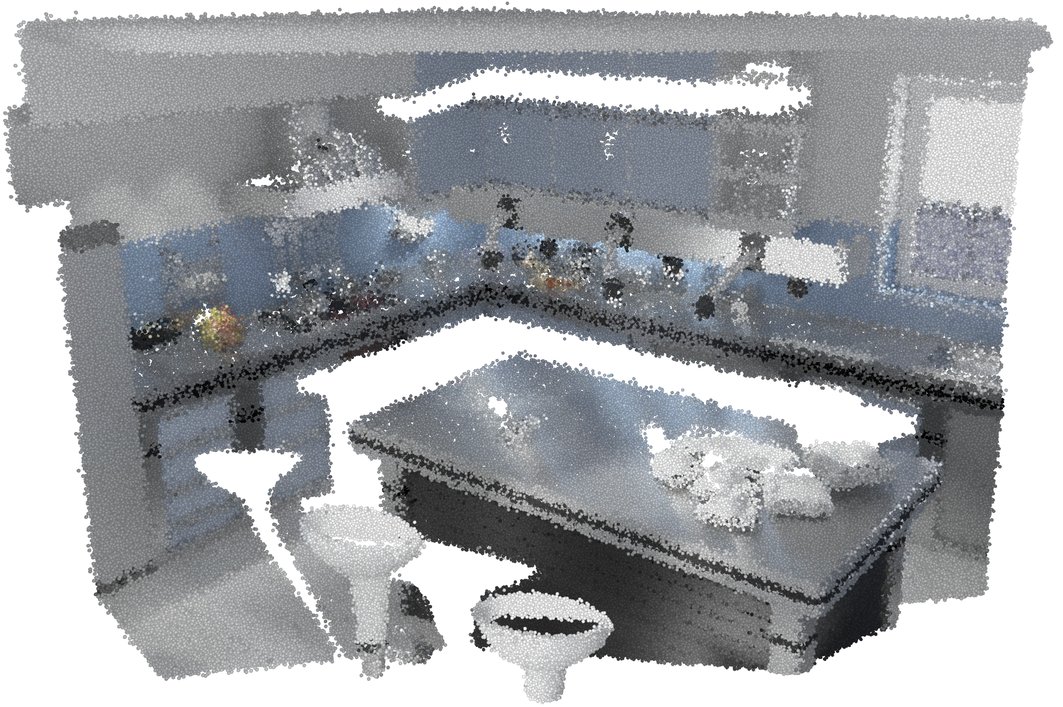}};

        \node[image, anchor=north] (norm-gt) at (pts-gt.south) {\includegraphics[width=\imgWidth]{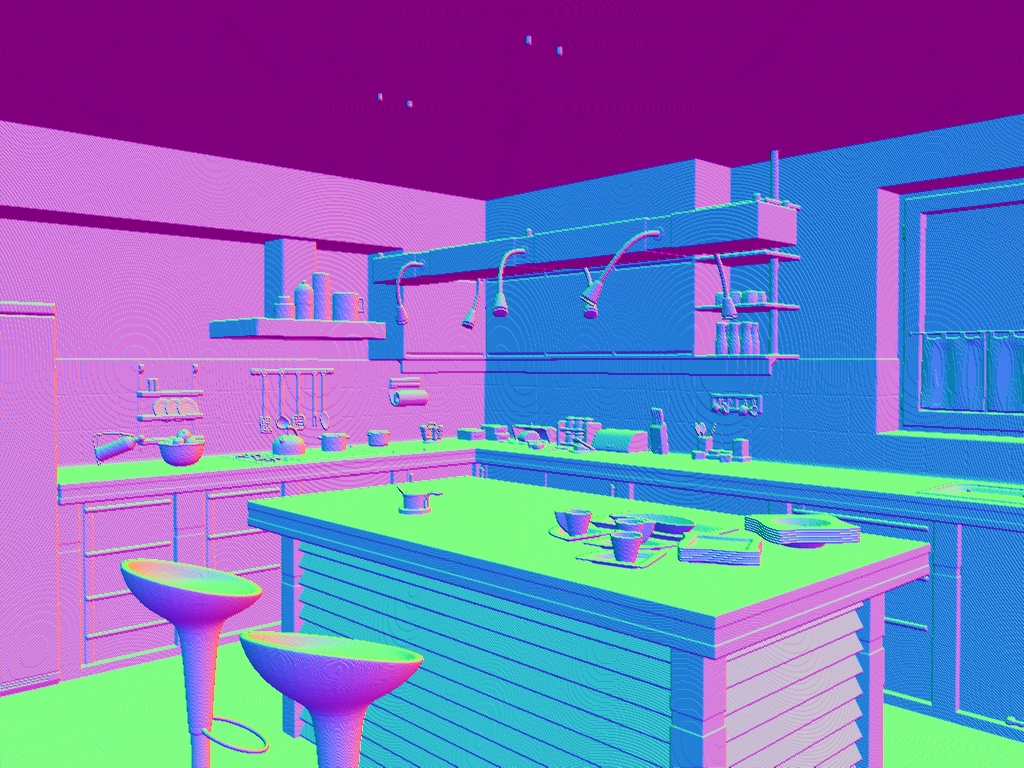}};
        \node[image, anchor=west] (norm-low) at ($(norm-gt.east)+(\imgSpacing,0)$) {\includegraphics[width=\imgWidth]{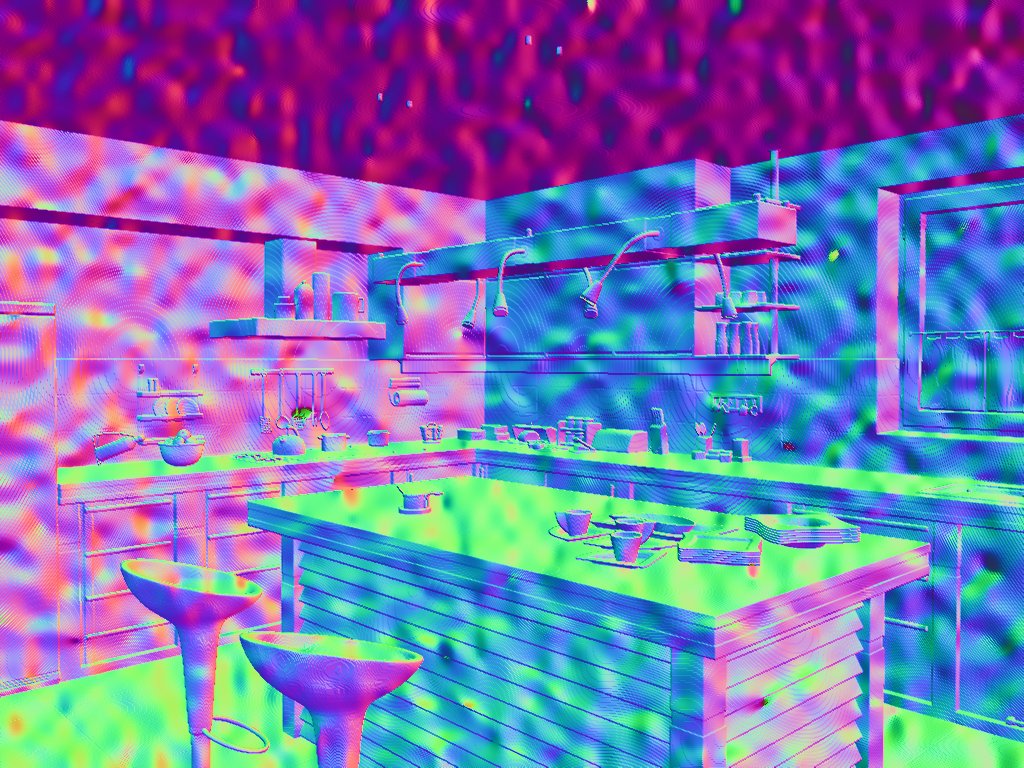}};
        \node[image, anchor=west] (norm-high) at ($(norm-low.east)+(\imgSpacing,0)$) {\includegraphics[width=\imgWidth]{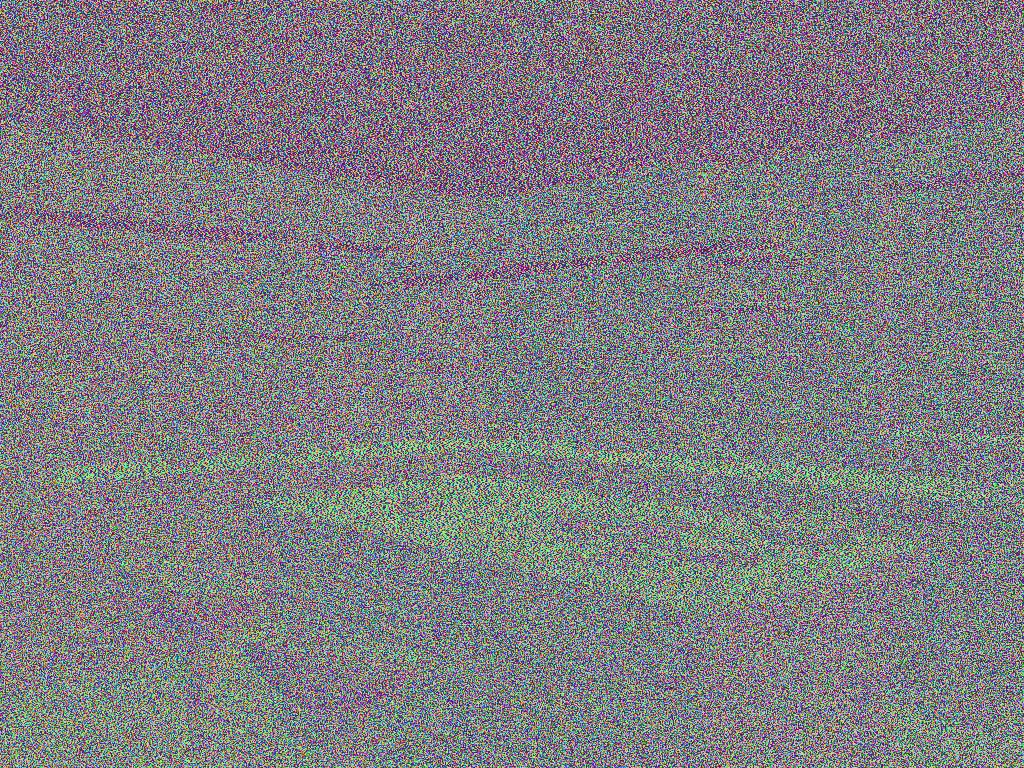}};

        \node[box] at (pts-low.south) {
            \begin{tabular}{@{}l@{\hspace{0.6em}}r@{}}
            \AbsRelGlob{} & \SI{0.62}{\percent}\\
            \AbsRelLoc{} & \SI{2.37}{\percent}\\
            \MAEnormal{} & \SI{19.0}{\degree}
            \end{tabular}
        };

        \node[box] at (pts-high.south) {
            \begin{tabular}{@{}l@{\hspace{0.6em}}r@{}}
            \AbsRelGlob{} & \SI{0.60}{\percent}\\
            \AbsRelLoc{} & \SI{2.39}{\percent}\\
            \MAEnormal{} & \SI{85.1}{\degree}
            \end{tabular}
        };

        \node[rotate=90, anchor=south] at (pts-gt.west) {Point Map};
        \node[rotate=90, anchor=south] at (norm-gt.west) {Point Map Normals};
        \node[column label] at (pts-gt.north) {Ground Truth};
        \node[column label] at (pts-low.north) {Low-Frequency Perturbation};
        \node[column label] at (pts-high.north) {High-Frequency Perturbation};
    \end{tikzpicture}
    }
    \caption{
        \textbf{Pointwise metrics only weakly capture local surface geometry.}
        We add low- and high-frequency perturbations to the same ground-truth point map.
        \AbsRelGlob{} and \AbsRelLoc{} average pointwise position errors, giving nearly identical scores even though the high-frequency perturbation yields much less coherent local surface geometry.
        \MAEnormal{} instead compares point map normals induced by neighboring point differences, and therefore reflects this degradation.
    }
    \label{fig:normal_metric}
\end{figure}
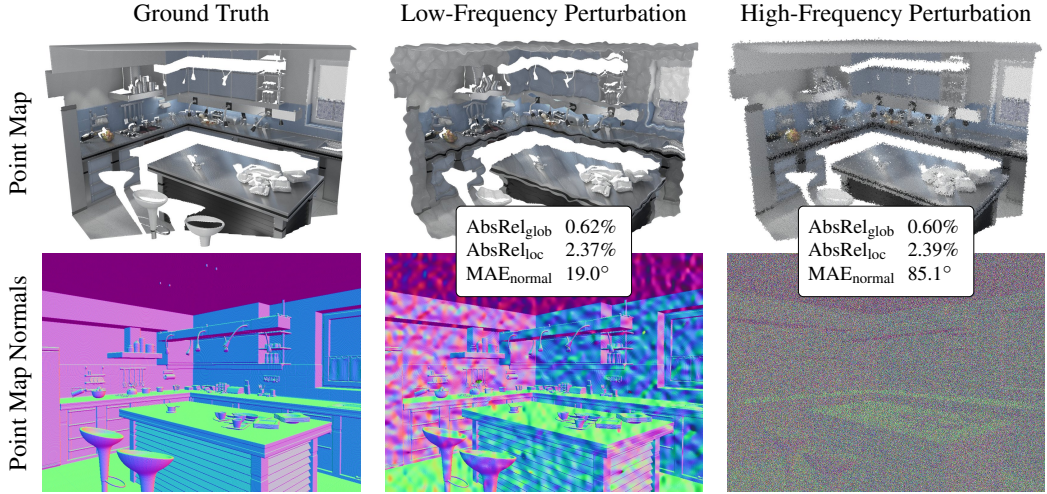

We hypothesize that errors in local surface geometry have received limited attention because standard point map metrics evaluate local surface geometry only indirectly.
They measure whether predicted points have accurate 3D positions on average after global alignment.
However, neighboring points can fail to form a coherent surface even when the individual point errors are small on average.
Surface ripples, blockiness, and distortions can arise from relatively small point displacements, so their contribution to an average point-position error can be weak compared to ordinary placement errors.
The local point map metric of \citet{wang2025moge} makes this evaluation more fine-grained by aligning segmented instances separately, but after alignment it still averages pointwise residuals.
We therefore complement pointwise metrics with a point map normal metric computed from neighboring point differences, which directly evaluates the local surface orientation induced by the predicted point map.
In a normal-based metric, artifacts are measured through the changes they induce in local surface orientation, rather than through the magnitude of the underlying point-position residuals; \cref{fig:normal_metric} illustrates this with low- and high-frequency perturbations of the ground-truth surface.

Equipped with a metric that better tracks local structure, we revisit the loss formulation of recent models.
Like the global point map metrics, commonly used global point losses~\cite{wang2025moge,wang2024dust3r} measure average point positioning error and are dominated by global geometry.
As a result, they require additional losses to enforce local consistency and to suppress oscillations and surface irregularities.
Common surface losses include terms on normals estimated from point maps and edge angles~\cite{wang2025moge,wang2025moge2} as well as several gradient matching variants~\cite{ranftl2022midas,li2018megadepth}.
However, none of these existing losses provides sufficiently strong surface supervision for point maps.
Interestingly, preliminary experiments showed that the log-depth gradient matching loss originally proposed for monocular depth~\cite{li2018megadepth} \emph{does} improve local surface geometry reliably, but tends to harm global geometry as it is not compatible with the point map formulation.
In this work, we adapt this loss to enable its direct application to point maps, while preserving its pairwise scale invariance.

We find that this improved supervision helps, but does not remove the main failure mode.
Therefore, we also investigate the architectural design of the decoder.
Recent point map models typically pair a ViT backbone with a convolutional decoder such as MoGe's ConvStack~\cite{wang2025moge} or a DPT head~\cite{ranftl2021dpt}.
We find that even with improved surface losses, these decoders still struggle to recover the 3D shape of thin structures in front of distant or complex backgrounds.
We hypothesize that convolution-based decoders struggle here because these regions require reconstructing a high-frequency, high-amplitude signal in image space, which is difficult to represent in fixed convolutional kernels.
Scaling the convolutional decoder helps, but quickly reaches diminishing returns.
A ViT decoder such as in $\pi^3$~\cite{wang2026pi3} avoids fixed filters, but it processes all features at a single low-resolution scale determined by the patch size, which leads to patch-aligned artifacts.

Our solution is to keep the progressive multiscale decoding structure of convolution-based decoders, while replacing convolutions with blocks based on Neighborhood Attention~\cite{hassani2023nat}.
This yields a decoder that can selectively aggregate local evidence unlike convolutional decoders, without incurring the cost of full-resolution self-attention and without producing patch-aligned artifacts as a pure ViT decoder.
In practice, it produces more faithful thin structures and more stable local surface geometry than convolutional decoders.
Combined with our point gradient matching loss, the resulting model sets a new state of the art in the local point map and point map normal evaluations, while also achieving the best results in the global point map evaluation on several common zero-shot benchmarks.

Our contributions are as follows:
\begin{enumerate}
    \item An evaluation metric based on point map normals that better reflects local surface quality than standard point map metrics (\MAEnormal{}, \cref{eq:normal_metric}).
    \item A scale-invariant point gradient matching loss, inspired by log-depth gradient matching~\cite{li2018megadepth}, for supervising local surface structure in point maps (\(\mathcal{L}_\mathrm{pgm}\), \cref{eq:pgm_loss}).
    \item A decoder based on Neighborhood Attention for dense point map prediction that improves thin structures and local surface geometry (NAD, \cref{sec:arch}).
\end{enumerate}

\section{Related Work}
\label{sec:related}

\subsection{Feedforward Geometry Estimation}
Geometry estimation is a foundational task for many applications, including robotics, autonomous driving, and virtual reality.
Many state-of-the-art architectures combine a plain ViT initialized from DINOv2~\cite{oquab2023dinov2} with a convolutional decoder.
This decoder is often implemented as a DPT head~\cite{ranftl2021dpt,yang2024depth_anything_v1,yang2024depth_anything_v2,bochkovskii2024depthpro}, a ConvStack~\cite{wang2025moge,wang2025moge2}, or a similar convolutional multi-resolution architecture~\cite{piccinelli2025unidepthv2}.

Notable exceptions to this are InfiniDepth~\cite{yu2026infinidepth}, which uses a query-based decoder that decodes depth for individual continuous image coordinates, and Pixel-Perfect Depth~(PPD)~\cite{xu2025ppd}, which uses a decoder inspired by the DiT architecture~\cite{peebles2023dit}.
InfiniDepth and PPD predict affine-invariant depth in log space instead of point maps.
Unprojecting affine-invariant log-depth to 3D geometry requires a reference point map; both models use MoGe-2 to estimate this reference point map~\cite{yu2026infinidepth,xu2025ppd}.
While InfiniDepth and PPD aim to improve fine-grained geometry estimation, we observe blocky surface artifacts for InfiniDepth and severe surface noise for PPD\@.
We show a quantitative comparison in~\cref{sec:experiments} and qualitative examples in \cref{supp:qualitative}.

Current multi-view methods also achieve impressive results in the monocular setting.
Their architecture follows a similar scheme as monocular models, combining a ViT encoder with an inter-frame feature fusion module and a per-frame decoder.
DUSt3R~\cite{wang2024dust3r} uses a ViT pretrained with cross-view completion~\cite{weinzaepfel2022croco_v1,weinzaepfel2023croco_v2} and a DPT head for point map predictions.
It predicts point maps from two input views in a shared coordinate system.
Follow-up works employ global attention over all frames, enabling faster processing of large image sets~\cite{wang2025vggt,keetha2026mapanything,wang2026pi3,lin2026depthanything3,yang2025fast3r}.
Similar to the previous two-view and monocular methods, these methods often use a DPT head to regress one point map per input image~\cite{wang2025vggt,keetha2026mapanything,lin2026depthanything3,yang2025fast3r}.
A notable exception is $\pi^3$~\cite{wang2026pi3}, which combines a plain Transformer~\cite{vaswani2017attention} with a final pixel shuffle~\cite{shi2016pixelshuffle} for point predictions.

The impressive performance of these methods illustrates the tremendous progress in geometry estimation thanks to strong backbones, improved feature fusion modules, large-scale training, and improved supervision.
However, we still observe poor predictions for local surface geometry, especially around thin structures (see~\cref{fig:teaser}).
Compared to previous methods, our Neighborhood Attention Decoder (NAD) visibly improves in this regard.
Unlike DPT heads and ConvStack decoders, it is not restricted to fixed convolutional kernels, and unlike ViT decoders, it is not plagued by patching artifacts.

\subsection{Evaluation of Local Surface Geometry}

Point map metrics used in recent work remain pointwise after alignment: global affine-invariant AbsRel evaluates all valid pixels, while the local point map metric of \citet{wang2025moge} evaluates instance-mask regions after separate local alignment.
This improves sensitivity to errors inside individual object masks, but still does not measure whether neighboring predictions form a coherent surface.
\Cref{fig:normal_metric} illustrates this limitation and motivates the point map normal metric in \cref{eq:normal_metric}.

\subsection{Point Map Supervision}

DUSt3R~\cite{wang2024dust3r} popularized the point map representation for 3D geometry estimation and introduced the corresponding point map loss as direct regression of aligned or normalized 3D points.
Many follow-up works adapt this point map loss with different output parameterizations~\cite{wang2025vggt,keetha2026mapanything,lin2026depthanything3}.
The residuals, however, remain pointwise and do not directly constrain whether neighboring predictions form a coherent local surface.
Prior methods therefore add surface losses: point normals and edge angle losses supervise local orientation but do not constrain displacement magnitude~\cite{wang2025moge,wang2025moge2,lin2026depthanything3}; other depth and point map methods compare finite differences after one shared alignment or normalization of the whole prediction~\cite{ranftl2022midas,wang2025vggt}, but this does not reward local coherence in globally misplaced regions.
Log-depth gradient matching compares finite differences of log depth, yielding a pairwise scale-invariant residual~\cite{li2018megadepth}.
In \cref{sec:loss}, we adapt this pairwise idea to vector-valued point maps.

\subsection{Neighborhood Attention}
In Neighborhood Attention (NA)~\cite{hassani2023nat}, each query only attends to its local neighborhood.
It can be viewed as a way to introduce inductive bias into attention or to speed up computations due to sparsity~\cite{hassani2023nat,hassani2022dinat,hassani2024faster,hassani2025generalized}.
To introduce inductive bias, NA can be used to implement hierarchical feature encoders~\cite{hassani2023nat,hassani2022dinat}.
As a sparse attention variant, it has been used to replace full attention in isometric Transformer architectures such as video diffusion models~\cite{hassani2025generalized}.

Dilated NA has been used with encoder-decoder architectures as an efficient drop-in for regular attention, for example, in medical imaging~\cite{ding2024catunet,saadati2023dilatedunet} and image restoration~\cite{liu2025dinatir}.
NAF~\cite{chambon2025naf} proposes a decoder that utilizes NA to cross-attend between high-resolution image features of a small CNN and a low-resolution feature map of a vision foundation model in order to compute high-resolution features.
Compared to these works, our Neighborhood Attention Decoder (NAD, \cref{sec:arch}) neither cross-attends between resolutions nor uses dilated NA; instead, it uses plain NA to make attention-based mixing practical in a multi-stage dense geometry decoder, where high-resolution stages make global self-attention prohibitive.
To our knowledge, we are the first to demonstrate the effectiveness of NA as the core mixing mechanism in this setting.

\section{Method}
\label{sec:method}

Our model, \ours{}, predicts a dense point map \(\hat{\mathbf{P}} \in \mathbb{R}^{H\times W\times 3}\), mapping every pixel of a single image \(\mathbf{I}\in\mathbb{R}^{H\times W\times 3}\) to its corresponding 3D point.
It combines a DINOv2~\cite{oquab2023dinov2}-initialized ViT~\cite{dosovitskiy2021vit} encoder with a Neighborhood Attention Decoder (NAD, \cref{sec:arch}).
We train it with global and local point map losses and our proposed point gradient matching loss (\(\mathcal{L}_\mathrm{pgm}\), \cref{eq:pgm_loss}).

\subsection{Architecture}
\label{sec:arch}
\begin{figure}
    \centering
    \includegraphics[width=\linewidth]{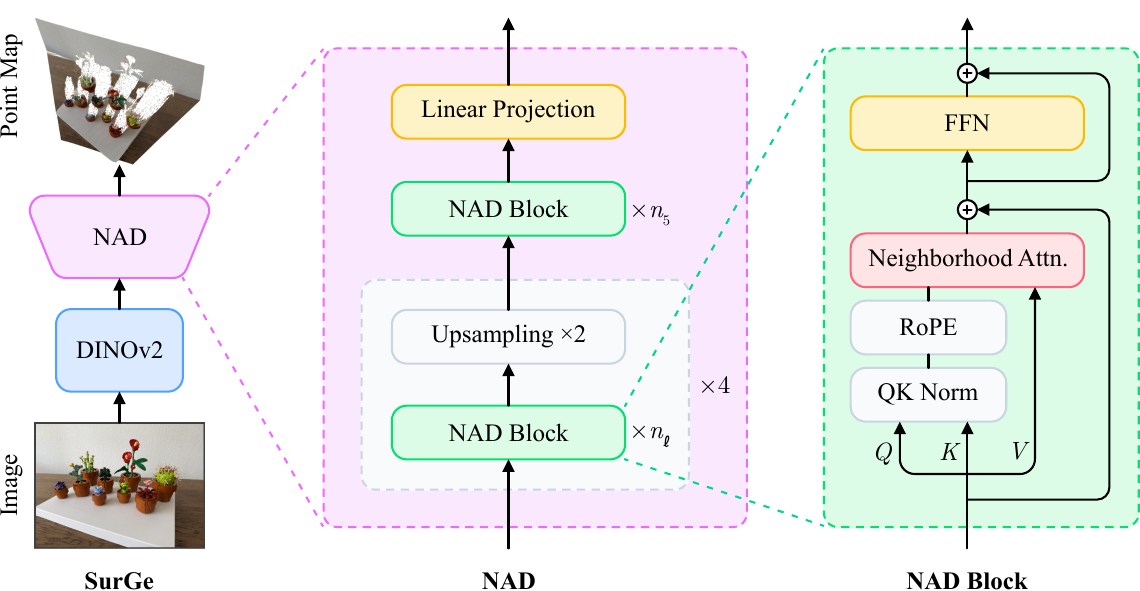}
    \caption{\textbf{\ours{} architecture overview.}
    \ours{} combines a DINOv2~\cite{oquab2023dinov2} encoder with our Neighborhood Attention Decoder (NAD).
    NAD upsamples encoder features through a sequence of stages \(\ell \in \{1, \ldots,5\}\), each built from \(n_\ell\) NAD blocks.
    Compared to standard ViT~\cite{dosovitskiy2021vit} blocks, NAD blocks replace global self-attention with Neighborhood Attention~\cite{hassani2023nat}, use window-matched RoPE~\cite{su2024rope} and only QK normalization~\cite{dehghani2023vit22b} instead of pre-attention and pre-feedforward LayerNorm~\cite{ba2016layernormalization}.}
    \label{fig:arch}
\end{figure}

\Cref{fig:arch} visualizes the architecture of \ours{}.
Following recent feedforward 3D reconstruction methods~\cite{wang2025moge,wang2025vggt}, we use a ViT-Large backbone~\cite{dosovitskiy2021vit} initialized from DINOv2~\cite{oquab2023dinov2}.
Given \(\mathbf{I}\), the encoder produces a low-resolution token grid \(\mathbf{Z}\in\mathbb{R}^{\frac{H}{16}\times \frac{W}{16}\times C}\).
The decoder must then process and upsample this grid to produce per-pixel point predictions $\hat{\mathbf{P}}$.
Recent dense prediction decoders do so with a progressive multi-resolution design~\cite{ranftl2021dpt,wang2025moge,piccinelli2025unidepthv2}, which we keep.
The remaining design choice is how to mix features inside each stage of this decoder.
Existing methods struggle with images containing thin structures, where foreground and background evidence are placed in the same small image neighborhood.
We hypothesize that feature-dependent attention can separate this evidence more easily than spatially shared convolutional kernels.
Therefore, \ours{} adopts attention in its decoder.
In practice, we use Neighborhood Attention rather than self-attention to obtain content-dependent local mixing at linear cost.
Notably, as this progressive decoder processes features at multiple resolutions, it prevents patch-aligned artifacts, unlike single-resolution ViT decoders.

We refer to the resulting decoder as Neighborhood Attention Decoder (NAD).
Starting from the encoder token grid \(\mathbf{Z}\), NAD consists of five stages that produce the predicted point map \(\hat{\mathbf{P}}\), with stages \num{1} through \num{4} upsampling by a factor of \(2\times\).
Following \citet{wang2025moge}, the final output projection predicts \((\xi,\eta,\rho)\) and maps it to points as \((\xi e^\rho,\eta e^\rho,e^\rho)\).
Each stage \(\ell\) operates on feature map \(\mathbf{X}_\ell \in \mathbb{R}^{\frac{H}{2^{5 - \ell}} \times \frac{W}{2^{5 - \ell}} \times C_\ell}\) and contains \(n_\ell\) NAD blocks.
For \ours{}, \(C_1,\ldots,C_5=(1024,512,256,128,64)\) and \(n_1=\cdots=n_5=3\).
For \(\ell<5\), the NAD blocks are followed by an upsampling module that doubles the spatial resolution and halves the channel dimension.
We implement this module as a transposed \(2\times{}2\) convolution with stride \num{2}, followed by a \(3\times{}3\) convolution~\cite{odena2016better_upsampling}.

Each NAD block is a Transformer-style~\cite{vaswani2017attention} residual block with a Neighborhood Attention layer and a pointwise FFN\@.
It uses window-matched RoPE~\cite{su2024rope} on queries and keys, and omits the usual pre-attention and pre-FFN LayerNorm~\cite{ba2016layernormalization} layers.
Across all stages, the attention window is \(k=9\) and the head dimension is \(d_h=64\); at stage \(\ell\), the FFN hidden dimension is \(4C_\ell\).
Removing LayerNorm empirically improves accuracy; we hypothesize that dense point regression benefits when activations can retain magnitude cues related to scene geometry.
To stabilize attention without pre-attention normalization, we use QK~normalization~\cite{zhai2022scalingvisiontransformers,dehghani2023vit22b}.
Additional details on stage-wise positional embeddings, the RoPE temperature, and normalization choices are given in~\cref{supp:arch_details}.

\subsection{Loss}
\label{sec:loss}

Following \citet{wang2025moge}, we use the global affine-invariant point map loss $\mathcal{L}_\mathrm{glob}$ with ROE~\cite{wang2025moge} alignment, as well as local patch losses \(\mathcal{L}_{\mathrm{loc},4}\), \(\mathcal{L}_{\mathrm{loc},16}\), and \(\mathcal{L}_{\mathrm{loc},64}\) with diameters \(\frac{1}{4}\), \(\frac{1}{16}\), and \(\frac{1}{64}\) of the image diagonal.
These losses supervise point positions under global and local affine-invariant alignments, but their residuals are still pointwise after alignment.
They therefore only weakly constrain whether neighboring pixels form a locally consistent 3D surface.
We address this by adding a surface loss on neighboring 3D point differences.

\paragraph{Surface supervision.}
Existing surface losses make different trade-offs.
Losses on normals estimated from point maps and on edge angles~\cite{wang2025moge,wang2025moge2} compare local surface orientation, but discard the magnitude of the underlying 3D displacements.
Some prior depth and point map methods compare finite differences after one shared alignment or normalization of the whole prediction~\cite{ranftl2022midas,lin2026depthanything3,wang2025vggt}.
This asks each local pair to match its target finite difference under the same aligned frame, even when that neighborhood is locally coherent but misplaced relative to the whole scene.
For local surface supervision, we instead want the comparison to be defined by the neighboring pair itself.

\citet{li2018megadepth} define such a pairwise comparison for a ground-truth depth map \(\mathbf{D}\) and predicted depth map \(\hat{\mathbf{D}}\) through the log-depth gradient matching loss:
{\small
\begin{equation}
    \label{eq:log_gm}
	\mathcal{L}_{\mathrm{gm}}(\hat{\mathbf{D}}, \mathbf{D}) =
		\frac{1}{2|\mathcal{V}_x|} \sum_{(i,j) \in \mathcal{V}_x} \abs{\Delta_x \log \hat{D}_{ij} - \Delta_x \log D_{ij}}
		+
        \frac{1}{2|\mathcal{V}_y|} \sum_{(i,j) \in \mathcal{V}_y}
		\abs{\Delta_y \log \hat{D}_{ij} - \Delta_y \log D_{ij}},
\end{equation}
}where \(\Delta_x,\Delta_y\) are forward finite differences, and \(\mathcal{V}_x,\mathcal{V}_y\) contain annotated pixels whose horizontal or vertical forward neighbor is also annotated.
The logarithm makes each residual depend on a neighboring pair's depth ratio rather than a global prediction scale, but this construction is tied to positive scalar depth.
Point maps are vector-valued and coordinate-wise logarithms or ratios are not geometrically meaningful.
Restricting the loss to the \(z\) coordinate, as in MapAnything~\cite{keetha2026mapanything}, preserves this locality for depth, but its corrections are not expressed as 3D point differences, making them poorly aligned with the point map losses.

We therefore extend the pairwise comparison to \emph{point maps} by matching depth-normalized 3D finite differences:
\begin{equation}
	\widetilde{\Delta}_x \mathbf{Q}_{ij} =
	\frac{\Delta_x \mathbf{Q}_{ij}}{\min([\mathbf{Q}_{ij}]_z, [\mathbf{Q}_{i,j+1}]_z)},
\end{equation}
where \(\mathbf{Q}\) is a point map, \([\mathbf{Q}_{ij}]_z\) denotes the \(z\) coordinate of \(\mathbf{Q}_{ij}\), and \(\widetilde{\Delta}_y\) is defined analogously for vertical pairs.
We normalize by \(z\) since our decoder represents predicted points as \((\xi e^\rho,\eta e^\rho,e^\rho)\), where \(z=e^\rho\) is the scalar scale factor shared by all three coordinates.
For a neighboring pair, we use the nearer endpoint depth as a conservative local scale.

Our point gradient matching loss averages Euclidean distances between corresponding predicted and ground-truth normalized finite differences over the same valid-pair sets:
\begin{equation}
    \label{eq:pgm_loss}
	\mathcal{L}_{\mathrm{pgm}}(\hat{\mathbf{P}}, \mathbf{P}) =
        \frac{1}{2|\mathcal{V}_x|} \sum_{(i,j) \in \mathcal{V}_x}
		\norm{\widetilde{\Delta}_x \hat{\mathbf{P}}_{ij} - \widetilde{\Delta}_x \mathbf{P}_{ij}}_2
		+
        \frac{1}{2|\mathcal{V}_y|} \sum_{(i,j) \in \mathcal{V}_y}
		\norm{\widetilde{\Delta}_y \hat{\mathbf{P}}_{ij} - \widetilde{\Delta}_y \mathbf{P}_{ij}}_2.
\end{equation}
A Python-like version is given in~\cref{code:pgm_loss}.

Our proposed \(\mathcal{L}_\mathrm{pgm}\) combines the following useful properties: the residual is computed on full 3D displacements, so it supervises both local direction and magnitude, while the depth normalization makes the comparison scale-invariant for each neighboring pair.

In practice, we evaluate \(\mathcal{L}_\mathrm{pgm}\) only on pairs whose two endpoints are annotated and omit pairs near occlusion boundaries.
At such boundaries, adjacent pixels often belong to different surfaces and the exact ground-truth transition depends on dataset-specific edge conventions and resampling.
Masking these pairs removes largely irreducible residuals and slightly improves training.

We choose the relative weight of \(\mathcal{L}_\mathrm{pgm}\) empirically, finding that a coefficient of \num{10} gives a good balance with the global point loss.
Our full dense-label objective is:
\[
	\mathcal{L}
	=
	\mathcal{L}_\mathrm{glob}
	+ \mathcal{L}_{\mathrm{loc},4}
	+ \mathcal{L}_{\mathrm{loc},16}
	+ \mathcal{L}_{\mathrm{loc},64}
	+ 10\mathcal{L}_\mathrm{pgm}.
\]
For noisier or sparse annotations, we omit high-frequency terms according to label quality; see~\cref{supp:training_details}.

\section{Experiments}
\label{sec:experiments}

\subsection{Experimental Setup}
\label{sec:exp_setup}

\paragraph{Training.}
Unless otherwise stated, we use the DINOv2~\cite{oquab2023dinov2}-Large encoder and NAD described in~\cref{sec:arch}.
We train on a balanced mix of twenty synthetic and real datasets covering outdoor, indoor, in-the-wild, driving, and object-centric domains~\cite{wilson2021argoverse2,baruch2021arkitscenes,yao2020blendedmvs,deitke2022objaverse,qiu2024richdreamer,wang2020gtasfm,roberts2021hypersim,wang2021irs,niklaus2019kenburns,li2023matrixcity,li2018megadepth,fonder2019midair,huang2018mvssynth,zheng2020structured3d,bengar2019synthiaAL,hernandez2017synthiaSF,wang2020tartanairV1,zamir2018taskonomy,tosi2021smdnets,gomez2025urbansyn,sun2020waymo}.
The dataset weights are listed in~\cref{supp:training_data}.
Following common practice~\cite{wang2025moge,bochkovskii2024depthpro}, we adapt supervision to label quality: for synthetic labels we use the full objective in~\cref{sec:method}, for SfM labels we omit \(\mathcal{L}_{\mathrm{loc},64}\) and \(\mathcal{L}_\mathrm{pgm}\), and for LiDAR labels we keep only \(\mathcal{L}_\mathrm{glob}\) and \(\mathcal{L}_{\mathrm{loc},4}\).
We optimize with AdamW~\cite{loshchilov2019adamw} for \num{120}K steps at total batch size \num{128}, using peak learning rates \(3\times10^{-4}\) for the decoder and \(3\times10^{-5}\) for the backbone, and a reciprocal square root schedule~\cite{zhai2022scalingvisiontransformers} with \num{1}K warmup steps and a \(10\%\) cooldown.
The first \(80\%\) of training uses a low-resolution budget of \num{1024} encoder tokens with a fixed image area and sampled target aspect ratios.
The final \(20\%\) uses higher resolutions with encoder-token budgets between \num{1024} and \num{2802}, and the final \(10\%\) samples only synthetic data.
More details are given in~\cref{supp:training_details}.
Every ablation follows the final model's full training protocol and changes only the component under study.
This full-scale protocol is deliberate: several effects observed in smaller proxy runs did not persist at the final scale.
A full run takes roughly \num{31} hours on \num{16} H100 GPUs for the low-resolution phase and another \num{8} hours on \num{32} H100 GPUs for the high-resolution phase.

\paragraph{Evaluation.}
We evaluate zero-shot on eight common monocular geometry benchmarks: NYUv2~\cite{silberman2012nyu}, KITTI~\cite{uhrig2017kitti_depth}, ETH3D~\cite{schops2017eth3d}, iBims-1~\cite{koch2018ibims}, GSO~\cite{downs2022gso}, Sintel~\cite{butler2012sintel}, DDAD~\cite{guizilini2020ddad}, and DIODE~\cite{vasiljevic2019diode}.
Following MoGe~\cite{wang2025moge}, we report ROE-aligned affine-invariant point map AbsRel metrics in global and instance-wise local forms.
Let $\mathcal{V}$ be the set of all valid pixels and let \(\hat{\mathbf{P}}^\star_{ij}=s\hat{\mathbf{P}}_{ij}+\mathbf{t}\) denote aligned predictions, where \(s\) is a shared scale factor and \(\mathbf{t}\) is a 3D translation.
The global absolute relative error is defined as
\[
    \AbsRelGlob
    =
    \frac{1}{|\mathcal{V}|} \sum_{(i,j)\in\mathcal{V}}
    \frac{\norm{\hat{\mathbf{P}}^\star_{ij}-\mathbf{P}_{ij}}_2}
         {\norm{\mathbf{P}_{ij}}_2}.
\]
For \AbsRelLoc{}, we evaluate each instance \(r\in\mathcal{R}\) with segmentation region \(\Omega_r\) over \(\mathcal{V}_r=\mathcal{V}\cap\Omega_r\).
We estimate a separate local scale-and-translation alignment \(\hat{\mathbf{P}}^{\star,r}_{ij}=s_r\hat{\mathbf{P}}_{ij}+\mathbf{t}_r\), and normalize residuals by the ground-truth instance diameter \(d_r=\norm{\max_{(i,j)\in\mathcal{V}_r}\mathbf{P}_{ij}-\min_{(i,j)\in\mathcal{V}_r}\mathbf{P}_{ij}}_\infty\), where \(\max\) and \(\min\) are taken component-wise:
\[
    \AbsRelLoc
    =
    \frac{1}{|\mathcal{R}|}\sum_{r\in\mathcal{R}}
    \frac{1}{|\mathcal{V}_r|}\sum_{(i,j)\in\mathcal{V}_r}
    \frac{\norm{\hat{\mathbf{P}}^{\star,r}_{ij}-\mathbf{P}_{ij}}_2}
         {d_r}.
\]
For our proposed point map normal mean angular error (\MAEnormal{}), we first form four local normals around each pixel from cross products of adjacent point map differences.
The valid local normals are averaged, giving normal maps \(\mathbf{N}_{ij}\) and \(\hat{\mathbf{N}}_{ij}\) for the ground truth and prediction.
We report
\begin{equation}
    \label{eq:normal_metric}
    \MAEnormal
    =
    \frac{1}{|\mathcal{V}_N|}
    \sum_{(i,j)\in\mathcal{V}_N}
    \angle(\hat{\mathbf{N}}_{ij}, \mathbf{N}_{ij})
\end{equation}
in degrees, where \(\mathcal{V}_N\) contains pixels whose annotated neighborhood defines at least one valid local normal.
We report \AbsRelLoc{} on benchmarks with instance masks, and \MAEnormal{} where dense ground truth supports reliable normal estimation.
Lower is better for all metrics.

\subsection{Comparison to the State of the Art}
\setcounter{topnumber}{3}
\begin{table}[t]
\centering
\begin{minipage}{0.485\textwidth}
\centering
\scriptsize
\setlength{\tabcolsep}{2.5pt}
\sisetup{table-column-width=23pt}
\caption{
    \textbf{SotA comparison for \AbsRelLoc{} (\%).}
}
\label{tab:main:local-points-rel}
\renewcommand{\arraystretch}{1.2}
\begin{tabularx}{\linewidth}{@{}L *{2}{S[table-format=1.2]} S[table-format=2.2] *{2}{S[table-format=1.2]}@{}}
\toprule
Method & {ETH3D} & {iBims-1} & {Sintel} & {DDAD} & {DIODE} \\
\midrule
DUSt3R~\cite{wang2024dust3r} & 6.31 & 5.43 & 11.79 & 9.23 & 7.31 \\
Depth Pro~\cite{bochkovskii2024depthpro} & 4.77 & 4.11 & 10.54 & 8.09 & 6.80 \\
MapAnything~\cite{keetha2026mapanything} & 5.80 & 5.16 & 12.93 & 9.10 & 6.92 \\
VGGT~\cite{wang2025vggt} & 3.89 & 5.09 & 9.62 & 8.11 & 5.90 \\
InfiniDepth~\cite{yu2026infinidepth}\textsuperscript{*} & 4.65 & 4.29 & 10.56 & 8.57 & 6.77 \\
VGGT-$\Omega$~\cite{wang2026vggtomega}\textsuperscript{\textdagger} & 4.12 & 5.23 & 11.49 & 7.83 & 7.30 \\
Depth~Any.~3~\cite{lin2026depthanything3} & 5.03 & 4.18 & 11.10 & 8.89 & 6.14 \\
PPD~\cite{xu2025ppd}\textsuperscript{*} & 4.09 & 4.11 & 11.22 & 8.48 & 6.13 \\
UniDepthV2~\cite{piccinelli2025unidepthv2} & 4.00 & 4.02 & 9.35 & 8.32 & 6.15 \\
$\pi^3$~\cite{wang2026pi3} & 4.10 & 4.52 & 10.58 & 7.24 & 5.76 \\
MoGe-2~\cite{wang2025moge2} & \cellcolor{tablethird}3.27 & \cellcolor{tablesecond}3.61 & \cellcolor{tablesecond}8.13 & \cellcolor{tablesecond}6.57 & \cellcolor{tablethird}5.09 \\
MoGe~\cite{wang2025moge} & \cellcolor{tablesecond}3.16 & \cellcolor{tablethird}3.79 & \cellcolor{tablethird}8.45 & \cellcolor{tablethird}6.70 & \cellcolor{tablesecond}4.77 \\
\textbf{\ours{}} & \cellcolor{tablefirst}\bfseries 2.66 & \cellcolor{tablefirst}\bfseries 3.33 & \cellcolor{tablefirst}\bfseries 7.66 & \cellcolor{tablefirst}\bfseries 6.29 & \cellcolor{tablefirst}\bfseries 4.63 \\
\bottomrule
\end{tabularx}
\end{minipage}
\hfill
\begin{minipage}{0.485\textwidth}
\centering
\scriptsize
\setlength{\tabcolsep}{2.5pt}
\sisetup{table-column-width=23pt}
\caption{
    \textbf{SotA comparison for \MAEnormal{} ($^\circ$).}
}
\label{tab:main:points-normal-angle-mae}
\renewcommand{\arraystretch}{1.2}
\begin{tabularx}{\linewidth}{@{}L *{5}{S[table-format=2.1]}@{}}
\toprule
Method & {ETH3D} & {iBims-1} & {GSO} & {Sintel} & {DIODE} \\
\midrule
DUSt3R~\cite{wang2024dust3r} & 26.2 & 24.0 & 28.7 & 43.6 & 21.1 \\
Depth Pro~\cite{bochkovskii2024depthpro} & 24.0 & 20.3 & 18.0 & 33.8 & 21.9 \\
MapAnything~\cite{keetha2026mapanything} & 25.2 & 25.1 & 18.2 & 40.3 & 22.7 \\
VGGT~\cite{wang2025vggt} & \cellcolor{tablethird}20.1 & 21.6 & \cellcolor{tablesecond}10.6 & 29.7 & \cellcolor{tablethird}14.7 \\
InfiniDepth~\cite{yu2026infinidepth}\textsuperscript{*} & 24.6 & 21.9 & 19.3 & 36.4 & 22.4 \\
VGGT-$\Omega$~\cite{wang2026vggtomega}\textsuperscript{\textdagger} & 22.6 & 23.0 & 16.1 & 33.6 & 20.4 \\
Depth~Any.~3~\cite{lin2026depthanything3} & 22.2 & 20.7 & 14.7 & 33.6 & 17.5 \\
PPD~\cite{xu2025ppd}\textsuperscript{*} & 60.4 & 38.2 & 26.6 & 53.8 & 49.6 \\
UniDepthV2~\cite{piccinelli2025unidepthv2} & 24.2 & 20.7 & 19.0 & 31.2 & 21.3 \\
$\pi^3$~\cite{wang2026pi3} & 22.3 & 21.1 & 13.3 & 34.0 & 18.0 \\
MoGe-2~\cite{wang2025moge2} & 21.0 & \cellcolor{tablethird}18.4 & 11.5 & \cellcolor{tablethird}27.9 & 15.2 \\
MoGe~\cite{wang2025moge} & \cellcolor{tablesecond}19.5 & \cellcolor{tablesecond}17.6 & \cellcolor{tablethird}11.0 & \cellcolor{tablesecond}25.2 & \cellcolor{tablesecond}12.4 \\
\textbf{\ours{}} & \cellcolor{tablefirst}\bfseries 18.3 & \cellcolor{tablefirst}\bfseries 16.5 & \cellcolor{tablefirst}\bfseries 10.5 & \cellcolor{tablefirst}\bfseries 24.5 & \cellcolor{tablefirst}\bfseries 12.0 \\
\bottomrule
\end{tabularx}
\end{minipage}
\par
\addvspace{\floatsep}
\nointerlineskip
\begin{minipage}{\textwidth}
\centering
\small
\setlength{\tabcolsep}{2.25pt}
\sisetup{table-column-width=29.5pt}
\caption{
    \textbf{State-of-the-art comparison for \AbsRelGlob{} (\%).}
    Avg.\ rank is computed over the eight datasets.
    \textsuperscript{*}~Uses MoGe-2 for alignment and unprojection.
    \textsuperscript{\textdagger}~September 2026 checkpoint.
}
\label{tab:main:points-affine-invariant-rel}
\renewcommand{\arraystretch}{1.2}
\begin{tabularx}{\linewidth}{@{}L S[table-format=1.2] S[table-format=2.2] *{3}{S[table-format=1.2]}
    S[table-format=2.2] S[table-format=2.2] S[table-format=1.2] S[table-format=2.2,table-column-width=36pt]@{}}
\toprule
Method & {NYUv2} & {KITTI} & {ETH3D} & {iBims-1} & {GSO} & {Sintel} & {DDAD} & {DIODE} & {Avg.\ rank} \\
\midrule
DUSt3R~\cite{wang2024dust3r} & 4.42 & 12.81 & 7.35 & 5.06 & 4.56 & 30.33 & 19.62 & 9.04 & 12.38 \\
Depth Pro~\cite{bochkovskii2024depthpro} & 4.36 & 9.16 & 7.73 & 4.34 & 3.16 & 19.57 & 14.44 & 6.29 & 9.75 \\
MapAnything~\cite{keetha2026mapanything} & 4.58 & 8.72 & 5.23 & 4.49 & 1.98 & 20.88 & 14.85 & 6.43 & 9.75 \\
VGGT~\cite{wang2025vggt} & 3.88 & 8.91 & 4.59 & 4.83 & \cellcolor{tablefirst}\bfseries 1.04 & 21.64 & 17.34 & 6.65 & 9.12 \\
InfiniDepth~\cite{yu2026infinidepth}\textsuperscript{*} & 4.51 & 6.23 & 7.12 & 4.43 & 3.03 & 19.39 & 14.24 & 6.62 & 9.12 \\
VGGT-$\Omega$~\cite{wang2026vggtomega}\textsuperscript{\textdagger} & 5.60 & 7.09 & 4.20 & 5.31 & 2.23 & 18.92 & 12.82 & 6.34 & 8.38 \\
Depth Anything 3~\cite{lin2026depthanything3} & 3.78 & 6.19 & 5.47 & 3.75 & 1.70 & 21.50 & 14.52 & 6.44 & 7.62 \\
PPD~\cite{xu2025ppd}\textsuperscript{*} & 3.81 & 6.89 & 4.55 & 4.06 & 1.48 & 17.96 & 12.84 & 5.86 & 6.25 \\
UniDepthV2~\cite{piccinelli2025unidepthv2} & 3.66 & \cellcolor{tablesecond}4.75 & 4.38 & 4.05 & 2.91 & 17.93 & 11.96 & 7.45 & 6.12 \\
$\pi^3$~\cite{wang2026pi3} & \cellcolor{tablethird}3.41 & 8.16 & 4.91 & 3.93 & 1.49 & 17.68 & 11.88 & 5.24 & 5.38 \\
MoGe-2~\cite{wang2025moge2} & \cellcolor{tablesecond}3.33 & 6.47 & \cellcolor{tablethird}3.89 & \cellcolor{tablethird}3.65 & 1.16 & \cellcolor{tablethird}17.45 & \cellcolor{tablesecond}10.14 & \cellcolor{tablethird}5.13 & \cellcolor{tablethird}3.25 \\
MoGe~\cite{wang2025moge} & 3.69 & \cellcolor{tablefirst}\bfseries 4.67 & \cellcolor{tablesecond}3.54 & \cellcolor{tablesecond}3.34 & \cellcolor{tablethird}1.14 & \cellcolor{tablefirst}\bfseries 16.87 & \cellcolor{tablethird}10.44 & \cellcolor{tablefirst}\bfseries 4.44 & \cellcolor{tablesecond}2.25 \\
\textbf{\ours{}} & \cellcolor{tablefirst}\bfseries 3.31 & \cellcolor{tablethird}4.80 & \cellcolor{tablefirst}\bfseries 3.51 & \cellcolor{tablefirst}\bfseries 3.31 & \cellcolor{tablesecond}1.11 & \cellcolor{tablesecond}17.13 & \cellcolor{tablefirst}\bfseries 9.05 & \cellcolor{tablesecond}4.88 & \cellcolor{tablefirst}\bfseries 1.62 \\
\bottomrule
\end{tabularx}
\end{minipage}
\end{table}

We compare \ours{} against recent feedforward 3D reconstruction methods.
The clearest gains appear in the local evaluations, \AbsRelLoc{} and \MAEnormal{}.
These evaluations are complementary: \AbsRelLoc{} measures pointwise point map accuracy after instance-level alignment, whereas \MAEnormal{} evaluates the local surface orientation induced by neighboring point predictions.
In \cref{tab:main:local-points-rel}, \ours{} gives the lowest \AbsRelLoc{} on every evaluated dataset.
The same pattern holds for \MAEnormal{} in \cref{tab:main:points-normal-angle-mae}, indicating that \ours{}'s point maps form more accurate local surfaces rather than only lower pointwise error.

\Cref{tab:main:points-affine-invariant-rel} shows that global point map accuracy is already highly competitive among the strongest recent models, especially MoGe, MoGe-2~\cite{wang2025moge2}, and \(\pi^3\)~\cite{wang2026pi3}.
Within this tighter regime, \ours{} remains consistently strong: it ranks first on four of the eight datasets and obtains the best average rank overall.
Thus, gains in the local evaluations do not come at the expense of global scene geometry.

The qualitative examples in \cref{fig:teaser} show the same trend.
MoGe-2 and InfiniDepth~\cite{yu2026infinidepth} recover plausible coarse shape but produce irregular local surfaces, visible in both the rendered 3D geometry and the point map normals.
In contrast, \ours{} preserves thin structures and produces smoother surfaces with sharper geometric detail.
Together, the quantitative and qualitative results show that \ours{} improves local surface quality while preserving state-of-the-art global point map accuracy.

\subsection{Ablation Study}

\paragraph{Decoder design.}
\begin{table}[t]
\centering
\begin{minipage}{0.485\textwidth}
\centering
\scriptsize
\setlength{\tabcolsep}{3pt}
\sisetup{table-column-width=24pt}
\caption{
\textbf{Decoder ablation for \AbsRelLoc{} (\%).}
}
\label{tab:decoder:local-points-rel}
\renewcommand{\arraystretch}{1.2}
\begin{tabularx}{\linewidth}{@{}L *{5}{S[table-format=1.2]}@{}}
\toprule
Decoder & {ETH3D} & {iBims-1} & {Sintel} & {DDAD} & {DIODE} \\
\midrule
ViT & 4.12 & 4.13 & 9.29 & 7.76 & 6.25 \\
DPT head & 3.25 & 3.58 & 8.42 & 6.62 & 5.06 \\
ConvStack & \cellcolor{tablethird}3.04 & \cellcolor{tablethird}3.49 & \cellcolor{tablesecond}8.11 & \cellcolor{tablethird}6.55 & \cellcolor{tablethird}4.93 \\
ConvStack-L & \cellcolor{tablesecond}2.94 & \cellcolor{tablesecond}3.44 & \cellcolor{tablethird}8.15 & \cellcolor{tablesecond}6.43 & \cellcolor{tablesecond}4.82 \\
\textbf{NAD (ours)} & \cellcolor{tablefirst}\bfseries 2.66 & \cellcolor{tablefirst}\bfseries 3.33 & \cellcolor{tablefirst}\bfseries 7.66 & \cellcolor{tablefirst}\bfseries 6.29 & \cellcolor{tablefirst}\bfseries 4.63 \\
\bottomrule
\end{tabularx}
\end{minipage}
\hfill
\begin{minipage}{0.485\textwidth}
\centering
\scriptsize
\setlength{\tabcolsep}{3pt}
\sisetup{table-column-width=24pt}
\caption{
\textbf{Decoder ablation for \MAEnormal{} ($^\circ$).}
}
\label{tab:decoder:points-normal-angle-mae}
\renewcommand{\arraystretch}{1.2}
\begin{tabularx}{\linewidth}{@{}L *{5}{S[table-format=2.1]}@{}}
\toprule
Decoder & {ETH3D} & {iBims-1} & {GSO} & {Sintel} & {DIODE} \\
\midrule
ViT & 23.6 & 21.5 & 17.0 & 32.3 & 18.7 \\
DPT head & 20.3 & 18.5 & 11.6 & 28.3 & 14.9 \\
ConvStack & \cellcolor{tablethird}19.5 & \cellcolor{tablethird}17.3 & \cellcolor{tablethird}11.1 & \cellcolor{tablethird}26.9 & \cellcolor{tablethird}13.3 \\
ConvStack-L & \cellcolor{tablesecond}18.8 & \cellcolor{tablesecond}16.9 & \cellcolor{tablesecond}10.9 & \cellcolor{tablesecond}25.4 & \cellcolor{tablesecond}12.5 \\
\textbf{NAD (ours)} & \cellcolor{tablefirst}\bfseries 18.3 & \cellcolor{tablefirst}\bfseries 16.5 & \cellcolor{tablefirst}\bfseries 10.5 & \cellcolor{tablefirst}\bfseries 24.5 & \cellcolor{tablefirst}\bfseries 12.0 \\
\bottomrule
\end{tabularx}
\end{minipage}
\par
\addvspace{\floatsep}
\nointerlineskip
\begin{minipage}{\textwidth}
\centering
\small
\setlength{\tabcolsep}{3.5pt}
\sisetup{table-column-width=31pt}
\caption{
\textbf{Decoder ablation for \AbsRelGlob{} (\%).} Avg.\ rank is computed over the eight datasets.
}
\label{tab:decoder:points-affine-invariant-rel}
\renewcommand{\arraystretch}{1.2}
\begin{tabularx}{\linewidth}{@{}L *{5}{S[table-format=1.2]} S[table-format=2.2] S[table-format=2.2]
    S[table-format=1.2] S[table-format=1.2,table-column-width=36pt]@{}}
\toprule
Decoder & {NYUv2} & {KITTI} & {ETH3D} & {iBims-1} & {GSO} & {Sintel} & {DDAD} & {DIODE} & {Avg.\ rank} \\
\midrule
ViT & 3.90 & 5.65 & 3.86 & 3.66 & 1.32 & \cellcolor{tablesecond}17.88 & 10.12 & \cellcolor{tablefirst}\bfseries 4.80 & 4.12 \\
DPT head & \cellcolor{tablesecond}3.44 & \cellcolor{tablethird}5.08 & \cellcolor{tablefirst}\bfseries 3.51 & \cellcolor{tablesecond}3.37 & \cellcolor{tablesecond}1.14 & 18.18 & 9.42 & 5.02 & \cellcolor{tablesecond}3.12 \\
ConvStack & 3.54 & 5.15 & \cellcolor{tablethird}3.53 & \cellcolor{tablesecond}3.37 & \cellcolor{tablethird}1.17 & 18.07 & \cellcolor{tablethird}9.40 & \cellcolor{tablesecond}4.82 & 3.25 \\
ConvStack-L & \cellcolor{tablethird}3.49 & \cellcolor{tablesecond}4.99 & 3.65 & 3.43 & \cellcolor{tablethird}1.17 & \cellcolor{tablethird}17.93 & \cellcolor{tablesecond}9.25 & 4.93 & \cellcolor{tablethird}3.19 \\
\textbf{NAD (ours)} & \cellcolor{tablefirst}\bfseries 3.31 & \cellcolor{tablefirst}\bfseries 4.80 & \cellcolor{tablefirst}\bfseries 3.51 & \cellcolor{tablefirst}\bfseries 3.31 & \cellcolor{tablefirst}\bfseries 1.11 & \cellcolor{tablefirst}\bfseries 17.13 & \cellcolor{tablefirst}\bfseries 9.05 & \cellcolor{tablethird}4.88 & \cellcolor{tablefirst}\bfseries 1.31 \\
\bottomrule
\end{tabularx}
\end{minipage}
\end{table}

We compare \ours{}'s Neighborhood Attention Decoder (NAD) against a DPT head following VGGT~\cite{wang2025vggt}, MoGe's ConvStack~\cite{wang2025moge}, a larger \enquote{ConvStack-L} baseline that matches our stage layout, widths, and depth while keeping convolutional residual blocks, and a ViT~\cite{dosovitskiy2021vit} decoder following \(\pi^3\)~\cite{wang2026pi3}.
To isolate the effect of the decoder architecture, we attach each decoder to the same DINOv2 ViT-Large backbone and use the same point map output parameterization and loss, including \(\mathcal{L}_\mathrm{pgm}\).
\Cref{tab:decoder:local-points-rel,tab:decoder:points-normal-angle-mae,tab:decoder:points-affine-invariant-rel} show that increasing convolutional decoder capacity helps: ConvStack-L moderately improves over ConvStack on most datasets in both local and global evaluations.
However, our NAD performs even better, achieving the lowest error on every dataset in the \AbsRelLoc{} and \MAEnormal{} evaluations, and on seven of eight datasets for global point map accuracy.
The ViT decoder is competitive on some datasets for the global evaluation, but has the highest error on every dataset in the local evaluations; patch-level artifacts are one visible failure mode.

\newcommand{\decoderDrawArrow}[4]{\draw[red, line width=1.25pt, -latex]
		($($(decoderImageBase.south west)!#1!(decoderImageBase.south east)$)!#2!($(decoderImageBase.north west)!#1!(decoderImageBase.north east)$)$) --
		($($(decoderImageBase.south west)!#3!(decoderImageBase.south east)$)!#4!($(decoderImageBase.north west)!#3!(decoderImageBase.north east)$)$);
}
\newcommand{\decoderImage}[2][]{\begin{tikzpicture}[baseline=(decoderImageBase.center)]
		\node[inner sep=0pt, outer sep=0pt] (decoderImageBase) {\includegraphics[
				width=\linewidth
			]{#2}};
		#1
	\end{tikzpicture}}

\begin{figure}[t]
	\centering

	\begin{subfigure}{0.49\linewidth}
		\centering
		\decoderImage{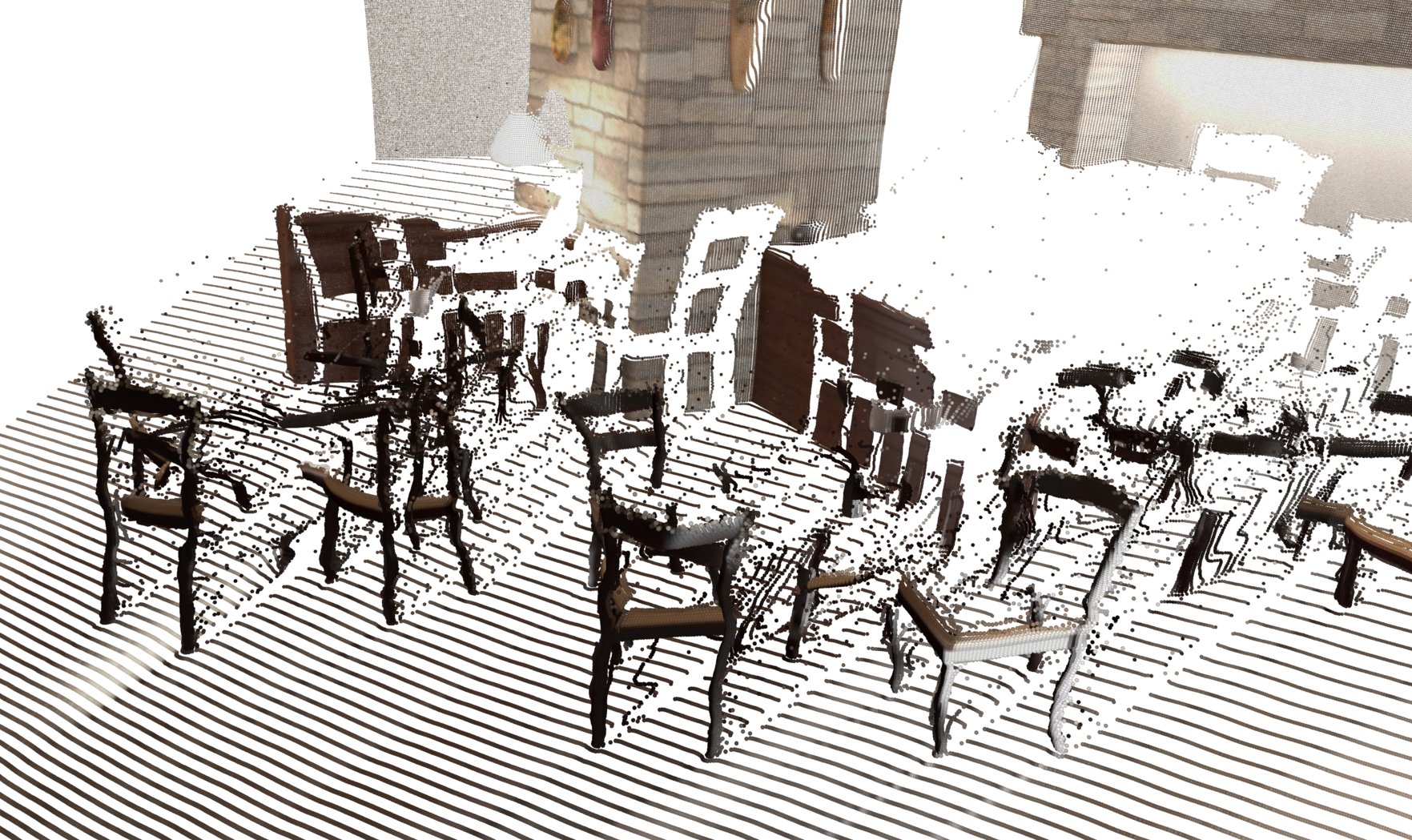}
		\caption{\textbf{NAD (ours)}}
	\end{subfigure}
	\hfill
	\begin{subfigure}{0.49\linewidth}
		\centering
		\decoderImage[
    		\decoderDrawArrow{0.005}{0.33}{0.07}{0.39}
    		\decoderDrawArrow{0.78}{0.68}{0.68}{0.58}
    		\decoderDrawArrow{0.3}{0.06}{0.40}{0.16}
		]
		{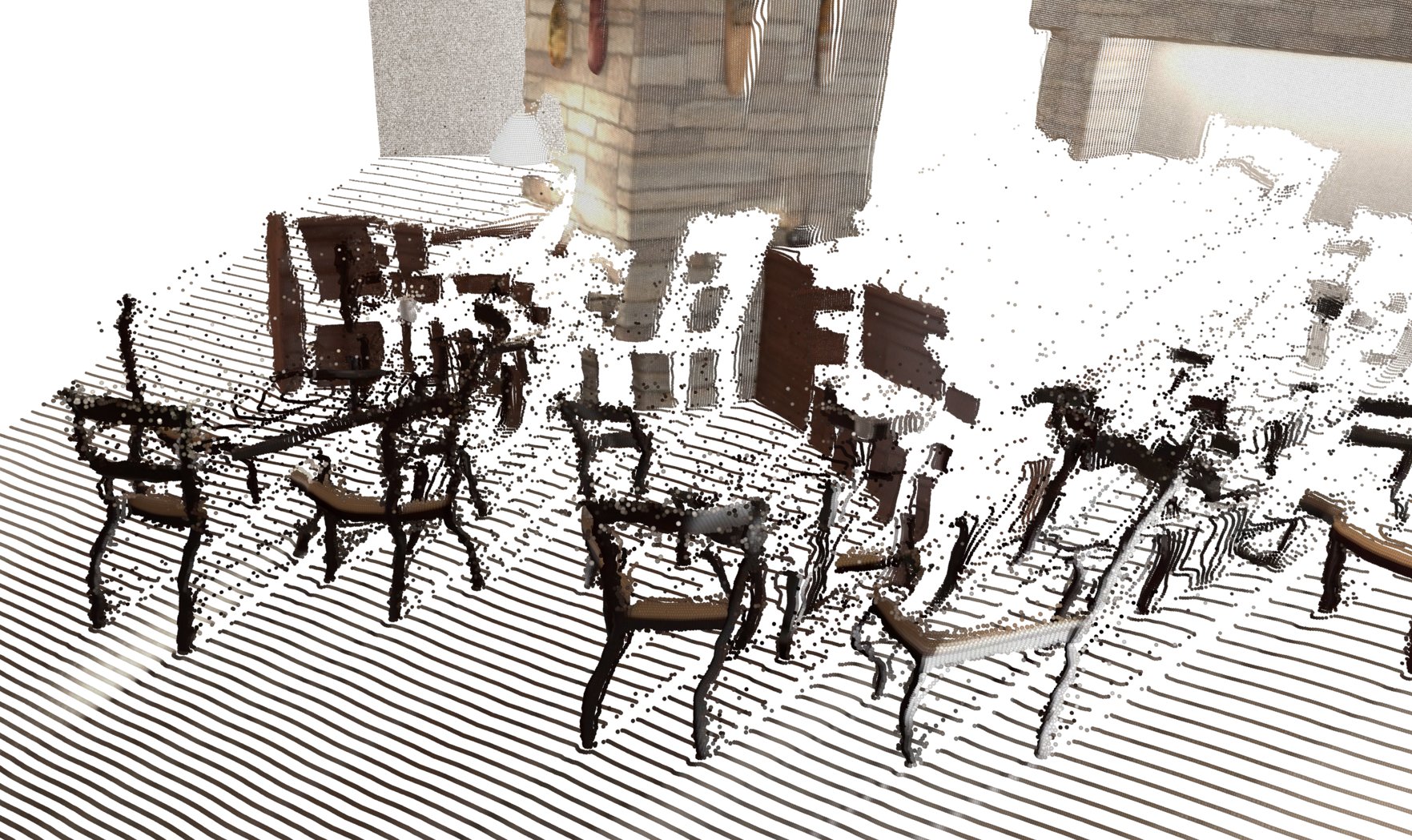}
		\caption{ConvStack-L}
	\end{subfigure}

	\caption{
        \textbf{Qualitative decoder ablation.}
        Our NAD produces less warped geometry than a convolutional decoder, visible in the chair legs and the wall to the right.
    }
	\label{fig:decoder_ablation}
\end{figure}

\Cref{fig:decoder_ablation} shows a qualitative comparison between \ours{} and the ConvStack-L configuration.
Compared to \ours{}, the model with the ConvStack-L decoder slightly warps thin structures such as the chair legs and misplaces parts of the wall on the right side.

\paragraph{Surface loss.}
\begin{table}[t]
\centering
\begin{minipage}{0.485\textwidth}
\centering
\scriptsize
\setlength{\tabcolsep}{3pt}
\sisetup{table-column-width=24pt}
\caption{
\textbf{Surface loss abl.\ for \AbsRelLoc{} (\%).}
}
\label{tab:loss:local-points-rel}
\renewcommand{\arraystretch}{1.2}
\begin{tabularx}{\linewidth}{@{}L *{5}{S[table-format=1.2]}@{}}
\toprule
Surface loss & {ETH3D} & {iBims-1} & {Sintel} & {DDAD} & {DIODE} \\
\midrule
None & \cellcolor{tablethird}2.81 & \cellcolor{tablethird}3.40 & \cellcolor{tablethird}7.97 & 6.32 & 4.66 \\
$\mathcal{L}_\mathrm{normal}$ & 2.87 & 3.42 & \cellcolor{tablethird}7.97 & \cellcolor{tablethird}6.30 & \cellcolor{tablesecond}4.63 \\
$\mathcal{L}_\mathrm{gm}$ & \cellcolor{tablesecond}2.78 & \cellcolor{tablesecond}3.35 & \cellcolor{tablesecond}7.86 & \cellcolor{tablefirst}\bfseries 6.23 & \cellcolor{tablefirst}\bfseries 4.59 \\
\textbf{$\mathcal{L}_\mathrm{pgm}$ (ours)} & \cellcolor{tablefirst}\bfseries 2.66 & \cellcolor{tablefirst}\bfseries 3.33 & \cellcolor{tablefirst}\bfseries 7.66 & \cellcolor{tablesecond}6.29 & \cellcolor{tablesecond}4.63 \\
\bottomrule
\end{tabularx}
\end{minipage}
\hfill
\begin{minipage}{0.485\textwidth}
\centering
\scriptsize
\setlength{\tabcolsep}{3pt}
\sisetup{table-column-width=24pt}
\caption{
\textbf{Surface loss abl.\ for \MAEnormal{} ($^\circ$).}
}
\label{tab:loss:points-normal-angle-mae}
\renewcommand{\arraystretch}{1.2}
\begin{tabularx}{\linewidth}{@{}L *{5}{S[table-format=2.1]}@{}}
\toprule
Surface loss & {ETH3D} & {iBims-1} & {GSO} & {Sintel} & {DIODE} \\
\midrule
None & 19.8 & 17.5 & 11.1 & 27.4 & 13.5 \\
$\mathcal{L}_\mathrm{normal}$ & \cellcolor{tablethird}19.1 & \cellcolor{tablethird}17.0 & \cellcolor{tablethird}10.9 & \cellcolor{tablethird}25.8 & \cellcolor{tablethird}12.6 \\
$\mathcal{L}_\mathrm{gm}$ & \cellcolor{tablesecond}18.4 & \cellcolor{tablefirst}\bfseries 16.5 & \cellcolor{tablesecond}10.8 & \cellcolor{tablesecond}24.8 & \cellcolor{tablefirst}\bfseries 11.8 \\
\textbf{$\mathcal{L}_\mathrm{pgm}$ (ours)} & \cellcolor{tablefirst}\bfseries 18.3 & \cellcolor{tablefirst}\bfseries 16.5 & \cellcolor{tablefirst}\bfseries 10.5 & \cellcolor{tablefirst}\bfseries 24.5 & \cellcolor{tablesecond}12.0 \\
\bottomrule
\end{tabularx}
\end{minipage}
\par
\addvspace{\floatsep}
\nointerlineskip
\begin{minipage}{\textwidth}
\centering
\small
\setlength{\tabcolsep}{3.5pt}
\sisetup{table-column-width=31pt}
\caption{
\textbf{Surface loss ablation for \AbsRelGlob{} (\%).} All rows use the same global and local point losses. Avg.\ rank is computed over the eight datasets.
}
\label{tab:loss:points-affine-invariant-rel}
\renewcommand{\arraystretch}{1.2}
\begin{tabularx}{\linewidth}{@{}L *{5}{S[table-format=1.2]} S[table-format=2.2] *{2}{S[table-format=1.2]}
    S[table-format=1.2,table-column-width=36pt]@{}}
\toprule
Surface loss & {NYUv2} & {KITTI} & {ETH3D} & {iBims-1} & {GSO} & {Sintel} & {DDAD} & {DIODE} & {Avg.\ rank} \\
\midrule
None & 3.46 & 5.10 & \cellcolor{tablethird}3.58 & \cellcolor{tablesecond}3.39 & \cellcolor{tablefirst}\bfseries 1.11 & \cellcolor{tablesecond}17.31 & 9.17 & 5.16 & 3.12 \\
$\mathcal{L}_\mathrm{normal}$ & \cellcolor{tablesecond}3.43 & \cellcolor{tablethird}5.01 & \cellcolor{tablefirst}\bfseries 3.51 & \cellcolor{tablethird}3.40 & \cellcolor{tablefirst}\bfseries 1.11 & 18.02 & \cellcolor{tablethird}9.09 & \cellcolor{tablethird}5.14 & \cellcolor{tablesecond}2.69 \\
$\mathcal{L}_\mathrm{gm}$ & \cellcolor{tablethird}3.44 & \cellcolor{tablesecond}5.00 & 3.72 & 3.46 & 1.17 & \cellcolor{tablethird}17.78 & \cellcolor{tablesecond}9.06 & \cellcolor{tablesecond}4.96 & \cellcolor{tablethird}3.00 \\
\textbf{$\mathcal{L}_\mathrm{pgm}$ (ours)} & \cellcolor{tablefirst}\bfseries 3.31 & \cellcolor{tablefirst}\bfseries 4.80 & \cellcolor{tablefirst}\bfseries 3.51 & \cellcolor{tablefirst}\bfseries 3.31 & \cellcolor{tablefirst}\bfseries 1.11 & \cellcolor{tablefirst}\bfseries 17.13 & \cellcolor{tablefirst}\bfseries 9.05 & \cellcolor{tablefirst}\bfseries 4.88 & \cellcolor{tablefirst}\bfseries 1.19 \\
\bottomrule
\end{tabularx}
\end{minipage}
\end{table}

We next ablate the surface loss while keeping the NAD and the pointwise global and local losses fixed.
We compare supervision from point map normals~\cite{wang2025moge} (\(\mathcal{L}_\mathrm{normal}\)), log-depth gradient matching on the \(z\) coordinate~\cite{li2018megadepth,keetha2026mapanything} (\(\mathcal{L}_\mathrm{gm}\)), and our point gradient matching loss (\(\mathcal{L}_\mathrm{pgm}\)).
As shown in \cref{tab:loss:points-normal-angle-mae,tab:loss:local-points-rel}, both gradient-based losses improve the local evaluations compared to \(\mathcal{L}_\mathrm{normal}\), and our proposed \(\mathcal{L}_\mathrm{pgm}\) improves over \(\mathcal{L}_\mathrm{gm}\) in most cases.
However, in the global evaluation shown in \cref{tab:loss:points-affine-invariant-rel}, \(\mathcal{L}_\mathrm{gm}\) generally performs worse than~\(\mathcal{L}_\mathrm{normal}\), while \(\mathcal{L}_\mathrm{pgm}\) noticeably outperforms both, which we attribute to the fact that the \(\mathcal{L}_\mathrm{pgm}\) signal remains in the same space as the global loss (\ie, 3D point displacements).
Overall, \(\mathcal{L}_\mathrm{pgm}\) provides a practical replacement for prior surface losses, improving local surface quality without sacrificing global point map accuracy.

\section{Conclusion}
\label{sec:conclusion}
We presented \ours{}, a monocular point map model designed to improve local surface geometry rather than only average point accuracy.
\ours{} improves the local point map and point map normal performance across eight zero-shot benchmarks while preserving strong global geometry; ablations show that both the NAD head and \(\mathcal{L}_\mathrm{pgm}\) contribute to these gains.
Our ablations further suggest that current decoder designs are a limiting factor in predicting high-quality local surface geometry.
Since the NAD head increases computational cost relative to smaller convolutional decoders, more efficient variants are a useful direction for future work.
Remaining failure cases are discussed in~\cref{supp:failure_cases}.
Beyond the model, our point map normal metric is a step toward improved geometric evaluation: it exposes surface artifacts that are only weakly captured by pointwise metrics.
We hope that this will make local surface quality a more explicit target for future geometry models.

{
\small
\paragraph{Acknowledgements.}
This work has received funding from the Jupiter AI Factory (JAIF), jointly funded by the European High Performance Computing Joint Undertaking (JU), the German Federal Ministry of Research, Technology and Space (BMFTR), and the Ministry of Culture and Science of North Rhine-Westphalia (MKW NRW) under grant agreement No.\ \texttt{101250682}.
The authors gratefully acknowledge the Gauss Centre for Supercomputing e.V.\ (\texttt{www.gauss-centre.eu}) for providing computing time through the John von Neumann Institute for Computing (NIC) on the GCS Supercomputers JUWELS and JUPITER at Jülich Supercomputing Centre (JSC).
}

{
    \small
    \bibliographystyle{abbrvnat}  \bibliography{main}

@string(PAMI={IEEE TPAMI})

@string(CVPR={CVPR})

@string(ICCV={ICCV})

@string(ECCV={ECCV})

@string(NIPS={NeurIPS})

@string(NIPSDB={NeurIPS Datasets and Benchmarks Track})

@string(BMVC={BMVC})

@string(TOG={ACM TOG})

@string(ICME={ICME})

@string(ICLR={ICLR})

@string(CVPRW={CVPR Workshops})

@string(ECCVW={ECCV Workshops})

@string(ICCVW={ICCV Workshops})

@string(ICRA={ICRA})

@string(TMLR={TMLR})

@string(ICML={ICML})

@string(WACV={WACV})

@string(IROS={IROS})

@string(RAL={IEEE Robotics and Automation Letters})

@inproceedings{silberman2012nyu,
    title         = {Indoor Segmentation and Support Inference from {RGBD} Images},
    author        = {Nathan Silberman and Derek Hoiem and Pushmeet Kohli and Rob Fergus},
    year          = 2012,
    booktitle     = ECCV,
}

@inproceedings{butler2012sintel,
    title         = {A Naturalistic Open Source Movie for Optical Flow Evaluation},
    author        = {Daniel J. Butler and Jonas Wulff and Garrett B. Stanley and Michael J. Black},
    year          = 2012,
    booktitle     = ECCV,
}

@article{ba2016layernormalization,
    title         = {Layer Normalization},
    author        = {Jimmy Lei Ba and Jamie Ryan Kiros and Geoffrey E. Hinton},
    year          = 2016,
    journal       = {arXiv preprint arXiv:1607.06450},
}

@article{odena2016better_upsampling,
    title         = {Deconvolution and Checkerboard Artifacts},
    author        = {Odena, Augustus and Dumoulin, Vincent and Olah, Chris},
    year          = 2016,
    journal       = {Distill},
}

@inproceedings{shi2016pixelshuffle,
    title         = {Real-time single image and video super-resolution using an efficient sub-pixel convolutional neural network},
    author        = {Shi, Wenzhe and Caballero, Jose and Husz{\'a}r, Ferenc and Totz, Johannes and Aitken, Andrew P and Bishop, Rob and Rueckert, Daniel and Wang, Zehan},
    year          = 2016,
    booktitle     = CVPR,
}

@inproceedings{uhrig2017kitti_depth,
    title         = {Sparsity Invariant {CNN}s},
    author        = {Jonas Uhrig and Nick Schneider and Lukas Schneider and Uwe Franke and Thomas Brox and Andreas Geiger},
    year          = 2017,
    booktitle     = \3DV,
}

@inproceedings{schops2017eth3d,
    title         = {A Multi-View Stereo Benchmark With High-Resolution Images and Multi-Camera Videos},
    author        = {Thomas Sch\"{o}ps and Johannes L. Sch\"{o}nberger and Silvano Galliani and Torsten Sattler and Konrad Schindler and Marc Pollefeys and Andreas Geiger},
    year          = 2017,
    booktitle     = CVPR,
}

@inproceedings{hernandez2017synthiaSF,
    title         = {Slanted Stixels: Representing San Francisco’s Steepest Streets},
    author        = {Hernandez-Juarez, Daniel and Schneider, Lukas and Espinosa, Antonio and Vazquez, David and Lopez, Antonio M. and Franke, Uwe and Pollefeys, Marc and Moure, Juan Carlos},
    year          = 2017,
    booktitle     = BMVC,
}

@inproceedings{vaswani2017attention,
    title         = {Attention is all you need},
    author        = {Vaswani, Ashish and Shazeer, Noam and Parmar, Niki and Uszkoreit, Jakob and Jones, Llion and Gomez, Aidan N and Kaiser, {\L}ukasz and Polosukhin, Illia},
    year          = 2017,
    booktitle     = NIPS,
}

@inproceedings{li2018megadepth,
    title         = {{MegaDepth}: Learning single-view depth prediction from internet photos},
    author        = {Li, Zhengqi and Snavely, Noah},
    year          = 2018,
    booktitle     = CVPR,
}

@inproceedings{koch2018ibims,
    title         = {Evaluation of {CNN}-based Single-Image Depth Estimation Methods},
    author        = {Tobias Koch and Lukas Liebel and Friedrich Fraundorfer and Marco K\"{o}rner},
    year          = 2018,
    booktitle     = ECCVW,
}

@inproceedings{huang2018mvssynth,
    title         = {DeepMVS: Learning Multi-View Stereopsis},
    author        = {Huang, Po-Han and Matzen, Kevin and Kopf, Johannes and Ahuja, Narendra and Huang, Jia-Bin},
    year          = 2018,
    booktitle     = CVPR,
}

@inproceedings{zamir2018taskonomy,
    title         = {Taskonomy: Disentangling Task Transfer Learning},
    author        = {Zamir, Amir R and Sax, Alexander and Shen, William B and Guibas, Leonidas and Malik, Jitendra and Savarese, Silvio},
    year          = 2018,
    booktitle     = CVPR,
}

@inproceedings{loshchilov2019adamw,
    title         = {Decoupled Weight Decay Regularization},
    author        = {Ilya Loshchilov and Frank Hutter},
    year          = 2019,
    booktitle     = ICLR,
}

@article{vasiljevic2019diode,
    title         = {{DIODE}: A Dense Indoor and Outdoor {DE}pth Dataset},
    author        = {Igor Vasiljevic and Nicholas I. Kolkin and Shanyi Zhang and Ruotian Luo and Haochen Wang and Falcon Z. Dai and Andrea F. Daniele and Mohammadreza Mostajabi and Steven Basart and Matthew R. Walter and Gregory Shakhnarovich},
    year          = 2019,
    journal       = {CoRR},
    volume        = {abs/1908.00463},
}

@article{niklaus2019kenburns,
    title         = {3D Ken Burns Effect from a Single Image},
    author        = {Simon Niklaus and Long Mai and Jimei Yang and Feng Liu},
    year          = 2019,
    journal       = TOG,
}

@inproceedings{fonder2019midair,
    title         = {Mid-Air: A multi-modal dataset for extremely low altitude drone flights},
    author        = {Michael Fonder and Marc Van Droogenbroeck},
    year          = 2019,
    booktitle     = CVPRW,
}

@inproceedings{bengar2019synthiaAL,
    title         = {Temporal Coherence for Active Learning in Videos},
    author        = {Zolfaghari Bengar, Javad and Gonzalez-Garcia, Abel and Villalonga, Gabriel and Raducanu, Bogdan and Aghdam, Hamed H and Mozerov, Mikhail and Lopez, Antonio M and van de Weijer, Joost},
    year          = 2019,
    booktitle     = ICCVW,
}

@inproceedings{zhang2019rmsnorm,
    title         = {Root mean square layer normalization},
    author        = {Zhang, Biao and Sennrich, Rico},
    year          = 2019,
    booktitle     = NIPS,
}

@inproceedings{guizilini2020ddad,
    title         = {{3D} Packing for Self-Supervised Monocular Depth Estimation},
    author        = {Vitor Guizilini and Rares Ambrus and Sudeep Pillai and Allan Raventos and Adrien Gaidon},
    year          = 2020,
    booktitle     = CVPR,
}

@inproceedings{zheng2020structured3d,
    title         = {{Structured3D}: A Large Photo-realistic Dataset for Structured 3D Modeling},
    author        = {Jia Zheng and Junfei Zhang and Jing Li and Rui Tang and Shenghua Gao and Zihan Zhou},
    year          = 2020,
    booktitle     = ECCV,
}

@inproceedings{sun2020waymo,
    title         = {Scalability in Perception for Autonomous Driving: Waymo Open Dataset},
    author        = {Sun, Pei and Kretzschmar, Henrik and Dotiwalla, Xerxes and Chouard, Aurelien and Patnaik, Vijaysai and Tsui, Paul and Guo, James and Zhou, Yin and Chai, Yuning and Caine, Benjamin and Vasudevan, Vijay and Han, Wei and Ngiam, Jiquan and Zhao, Hang and Timofeev, Aleksei and Ettinger, Scott and Krivokon, Maxim and Gao, Amy and Joshi, Aditya and Zhang, Yu and Shlens, Jonathon and Chen, Zhifeng and Anguelov, Dragomir},
    year          = 2020,
    booktitle     = CVPR,
}

@inproceedings{yao2020blendedmvs,
    title         = {{BlendedMVS}: A Large-scale Dataset for Generalized Multi-view Stereo Networks},
    author        = {Yao, Yao and Luo, Zixin and Li, Shiwei and Zhang, Jingyang and Ren, Yufan and Zhou, Lei and Fang, Tian and Quan, Long},
    year          = 2020,
    booktitle     = CVPR,
}

@inproceedings{wang2020tartanairV1,
    title         = {TartanAir: A Dataset to Push the Limits of Visual SLAM},
    author        = {Wang, Wenshan and Zhu, Delong and Wang, Xiangwei and Hu, Yaoyu and Qiu, Yuheng and Wang, Chen and Hu, Yafei and Kapoor, Ashish and Scherer, Sebastian},
    year          = 2020,
    booktitle     = IROS,
}

@article{wang2020gtasfm,
    title         = {Flow-motion and depth network for monocular stereo and beyond},
    author        = {Wang, Kaixuan and Shen, Shaojie},
    year          = 2020,
    journal       = RAL,
}

@inproceedings{dosovitskiy2021vit,
    title         = {An Image is Worth 16x16 Words: Transformers for Image Recognition at Scale},
    author        = {Alexey Dosovitskiy and Lucas Beyer and Alexander Kolesnikov and Dirk Weissenborn and Xiaohua Zhai and Thomas Unterthiner and Mostafa Dehghani and Matthias Minderer and Georg Heigold and Sylvain Gelly and Jakob Uszkoreit and Neil Houlsby},
    year          = 2021,
    booktitle     = ICLR,
}

@inproceedings{tosi2021smdnets,
    title         = {{SMD-Nets}: Stereo Mixture Density Networks},
    author        = {Fabio Tosi and Yiyi Liao and Carolin Schmitt and Andreas Geiger},
    year          = 2021,
    booktitle     = CVPR,
}

@inproceedings{baruch2021arkitscenes,
    title         = {{ARK}itScenes - A Diverse Real-World Dataset for 3D Indoor Scene Understanding Using Mobile {RGB}-D Data},
    author        = {Gilad Baruch and Zhuoyuan Chen and Afshin Dehghan and Tal Dimry and Yuri Feigin and Peter Fu and Thomas Gebauer and Brandon Joffe and Daniel Kurz and Arik Schwartz and Elad Shulman},
    year          = 2021,
    booktitle     = NIPSDB,
}

@inproceedings{ranftl2021dpt,
    title         = {Vision transformers for dense prediction},
    author        = {Ranftl, Ren{\'e} and Bochkovskiy, Alexey and Koltun, Vladlen},
    year          = 2021,
    booktitle     = ICCV,
}

@inproceedings{wilson2021argoverse2,
    title         = {{Argoverse 2}: Next Generation Datasets for Self-driving Perception and Forecasting},
    author        = {Benjamin Wilson and William Qi and Tanmay Agarwal and John Lambert and Jagjeet Singh and Siddhesh Khandelwal and Bowen Pan and Ratnesh Kumar and Andrew Hartnett and Jhony Kaesemodel Pontes and Deva Ramanan and Peter Carr and James Hays},
    year          = 2021,
    booktitle     = NIPSDB,
}

@inproceedings{roberts2021hypersim,
    title         = {{Hypersim}: {A} Photorealistic Synthetic Dataset for Holistic Indoor Scene Understanding},
    author        = {Mike Roberts and Jason Ramapuram and Anurag Ranjan and Atulit Kumar and Miguel Angel Bautista and Nathan Paczan and Russ Webb and Joshua M. Susskind},
    year          = 2021,
    booktitle     = ICCV,
}

@inproceedings{wang2021irs,
    title         = {Irs: A large naturalistic indoor robotics stereo dataset to train deep models for disparity and surface normal estimation},
    author        = {Wang, Qiang and Zheng, Shizhen and Yan, Qingsong and Deng, Fei and Zhao, Kaiyong and Chu, Xiaowen},
    year          = 2021,
    booktitle     = ICME,
}

@inproceedings{touvron2021layerscale,
    title         = {Going deeper with image transformers},
    author        = {Touvron, Hugo and Cord, Matthieu and Sablayrolles, Alexandre and Synnaeve, Gabriel and J{\'e}gou, Herv{\'e}},
    year          = 2021,
    booktitle     = ICCV,
}

@article{hassani2022dinat,
    title         = {Dilated Neighborhood Attention Transformer},
    author        = {Ali Hassani and Humphrey Shi},
    year          = 2022,
    journal       = {arXiv preprint arXiv:2209.15001},
}

@article{deitke2022objaverse,
    title         = {Objaverse: A Universe of Annotated 3D Objects},
    author        = {Matt Deitke and Dustin Schwenk and Jordi Salvador and Luca Weihs and Oscar Michel and Eli VanderBilt and Ludwig Schmidt and Kiana Ehsani and Aniruddha Kembhavi and Ali Farhadi},
    year          = 2022,
    journal       = {arXiv preprint arXiv:2212.08051},
}

@inproceedings{downs2022gso,
    title         = {Google Scanned Objects: A High-Quality Dataset of {3D} Scanned Household Items},
    author        = {Laura Downs and Anthony Francis and Nate Koenig and Brandon Kinman and Ryan Hickman and Krista Reymann and Thomas B. McHugh and Vincent Vanhoucke},
    year          = 2022,
    booktitle     = ICRA,
}

@article{ranftl2022midas,
    title         = {Towards Robust Monocular Depth Estimation: Mixing Datasets for Zero-Shot Cross-Dataset Transfer},
    author        = {Ren\'{e} Ranftl and Katrin Lasinger and David Hafner and Konrad Schindler and Vladlen Koltun},
    year          = 2022,
    journal       = PAMI,
}

@inproceedings{zhai2022scalingvisiontransformers,
    title         = {Scaling Vision Transformers},
    author        = {Xiaohua Zhai and Alexander Kolesnikov and Neil Houlsby and Lucas Beyer},
    year          = 2022,
    booktitle     = CVPR,
}

@article{weinzaepfel2022croco_v1,
    title         = {Croco: Self-supervised pre-training for 3d vision tasks by cross-view completion},
    author        = {Weinzaepfel, Philippe and Leroy, Vincent and Lucas, Thomas and Br{\'e}gier, Romain and Cabon, Yohann and Arora, Vaibhav and Antsfeld, Leonid and Chidlovskii, Boris and Csurka, Gabriela and Revaud, J{\'e}r{\^o}me},
    year          = 2022,
    journal       = NIPS,
}

@inproceedings{hassani2023nat,
    title         = {Neighborhood Attention Transformer},
    author        = {Ali Hassani and Steven Walton and Jiachen Li and Shen Li and Humphrey Shi},
    year          = 2023,
    booktitle     = CVPR,
}

@inproceedings{dehghani2023vit22b,
    title         = {Scaling vision transformers to 22 billion parameters},
    author        = {Dehghani, Mostafa and Djolonga, Josip and Mustafa, Basil and Padlewski, Piotr and Heek, Jonathan and Gilmer, Justin and Steiner, Andreas Peter and Caron, Mathilde and Geirhos, Robert and Alabdulmohsin, Ibrahim and others},
    year          = 2023,
    booktitle     = ICML,
}

@article{saadati2023dilatedunet,
    title         = {Dilated-UNet: A Fast and Accurate Medical Image Segmentation Approach using a Dilated Transformer and U-Net Architecture},
    author        = {Saadati, Davoud and Manzari, Omid Nejati and Mirzakuchaki, Sattar},
    year          = 2023,
    journal       = {arXiv preprint arXiv:2304.11450},
}

@inproceedings{li2023matrixcity,
    title         = {MatrixCity: A Large-scale City Dataset for City-scale Neural Rendering and Beyond},
    author        = {Li, Yixuan and Jiang, Lihan and Xu, Linning and Xiangli, Yuanbo and Wang, Zhenzhi and Lin, Dahua and Dai, Bo},
    year          = 2023,
    booktitle     = ICCV,
}

@inproceedings{peebles2023dit,
    title         = {Scalable diffusion models with transformers},
    author        = {Peebles, William and Xie, Saining},
    year          = 2023,
    booktitle     = ICCV,
}

@inproceedings{weinzaepfel2023croco_v2,
    title         = {Croco v2: Improved cross-view completion pre-training for stereo matching and optical flow},
    author        = {Weinzaepfel, Philippe and Lucas, Thomas and Leroy, Vincent and Cabon, Yohann and Arora, Vaibhav and Br{\'e}gier, Romain and Csurka, Gabriela and Antsfeld, Leonid and Chidlovskii, Boris and Revaud, J{\'e}r{\^o}me},
    year          = 2023,
    booktitle     = ICCV,
}

@inproceedings{jung2023hammer,
    title         = {On the Importance of Accurate Geometry Data for Dense {3D} Vision Tasks},
    author        = {Jung, HyunJun and Ruhkamp, Patrick and Zhai, Guangyao and Brasch, Nikolas and Li, Yitong and Verdie, Yannick and Song, Jifei and Zhou, Yiren and Armagan, Anil and Ilic, Slobodan and Leonardis, Ale\v{s} and Navab, Nassir and Busam, Benjamin},
    year          = 2023,
    booktitle     = CVPR,
}

@inproceedings{mehl2023spring,
    title         = {{Spring}: A High-Resolution High-Detail Dataset and Benchmark for Scene Flow, Optical Flow and Stereo},
    author        = {Mehl, Lukas and Schmalfuss, Jenny and Jahedi, Azin and Nalivayko, Yaroslava and Bruhn, Andr\'es},
    year          = 2023,
    booktitle     = CVPR,
}

@inproceedings{hassani2024faster,
    title         = {Faster Neighborhood Attention: Reducing the {$O(n^2)$} Cost of Self Attention at the Threadblock Level},
    author        = {Ali Hassani and Wen-Mei Hwu and Humphrey Shi},
    year          = 2024,
    booktitle     = {Advances in Neural Information Processing Systems},
}

@article{oquab2023dinov2,
    title         = {{DINOv2}: Learning Robust Visual Features without Supervision},
    author        = {Maxime Oquab and Timoth{\'e}e Darcet and Th{\'e}o Moutakanni and Huy V. Vo and Marc Szafraniec and Vasil Khalidov and Pierre Fernandez and Daniel Haziza and Francisco Massa and Alaaeldin El-Nouby and Mido Assran and Nicolas Ballas and Wojciech Galuba and Russell Howes and Po-Yao Huang and Shang-Wen Li and Ishan Misra and Michael Rabbat and Vasu Sharma and Gabriel Synnaeve and Hu Xu and Herve Jegou and Julien Mairal and Patrick Labatut and Armand Joulin and Piotr Bojanowski},
    year          = 2024,
    journal       = TMLR,
}

@inproceedings{qiu2024richdreamer,
    title         = {Richdreamer: A generalizable normal-depth diffusion model for detail richness in text-to-3d},
    author        = {Qiu, Lingteng and Chen, Guanying and Gu, Xiaodong and Zuo, Qi and Xu, Mutian and Wu, Yushuang and Yuan, Weihao and Dong, Zilong and Bo, Liefeng and Han, Xiaoguang},
    year          = 2024,
    booktitle     = CVPR,
}

@inproceedings{wang2024dust3r,
    title         = {{DUSt3R}: Geometric 3D Vision Made Easy},
    author        = {Wang, Shuzhe and Leroy, Vincent and Cabon, Yohann and Chidlovskii, Boris and Revaud, Jerome},
    year          = 2024,
    booktitle     = CVPR,
}

@inproceedings{yang2024depth_anything_v1,
    title         = {Depth Anything: Unleashing the Power of Large-Scale Unlabeled Data},
    author        = {Yang, Lihe and Kang, Bingyi and Huang, Zilong and Xu, Xiaogang and Feng, Jiashi and Zhao, Hengshuang},
    year          = 2024,
    booktitle     = CVPR,
}

@inproceedings{yang2024depth_anything_v2,
    title         = {Depth Anything V2},
    author        = {Yang, Lihe and Kang, Bingyi and Huang, Zilong and Zhao, Zhen and Xu, Xiaogang and Feng, Jiashi and Zhao, Hengshuang},
    year          = 2024,
    booktitle     = NIPS,
}

@article{ding2024catunet,
    title         = {CAT-Unet: An enhanced U-Net architecture with coordinate attention and skip-neighborhood attention transformer for medical image segmentation},
    author        = {Zhiquan Ding and Yuejin Zhang and Chenxin Zhu and Guolong Zhang and Xiong Li and Nan Jiang and Yue Que and Yuanyuan Peng and Xiaohui Guan},
    year          = 2024,
    journal       = {Information Sciences},
}

@article{su2024rope,
    title         = {Roformer: Enhanced transformer with rotary position embedding},
    author        = {Su, Jianlin and Ahmed, Murtadha and Lu, Yu and Pan, Shengfeng and Bo, Wen and Liu, Yunfeng},
    year          = 2024,
    journal       = {Neurocomputing},
}

@inproceedings{bae2024dsine,
    title         = {Rethinking Inductive Biases for Surface Normal Estimation},
    author        = {Bae, Gwangbin and Davison, Andrew J.},
    year          = 2024,
    booktitle     = CVPR,
}

@inproceedings{bochkovskii2024depthpro,
    title         = {{Depth Pro}: Sharp Monocular Metric Depth in Less Than a Second},
    author        = {Aleksei Bochkovskii and Ama\"{e}l Delaunoy and Hugo Germain and Marcel Santos and Yichao Zhou and Stephan R. Richter and Vladlen Koltun},
    year          = 2025,
    booktitle     = ICLR,
}

@article{hassani2025generalized,
    title         = {Generalized Neighborhood Attention: Multi-dimensional Sparse Attention at the Speed of Light},
    author        = {Hassani, Ali and Zhou, Fengzhe and Kane, Aditya and Huang, Jiannan and Chen, Chieh-Yun and Shi, Min and Walton, Steven and Hoehnerbach, Markus and Thakkar, Vijay and Isaev, Michael and others},
    year          = 2025,
    journal       = {arXiv preprint arXiv:2504.16922},
}

@article{liu2025dinatir,
    title         = {DiNAT-IR: Exploring Dilated Neighborhood Attention for High-Quality Image Restoration},
    author        = {Liu, Hanzhou and Li, Binghan and Liu, Chengkai and Lu, Mi},
    year          = 2025,
    journal       = {arXiv preprint arXiv:2507.17892},
}

@misc{chambon2025naf,
    title         = {NAF: Zero-Shot Feature Upsampling via Neighborhood Attention Filtering},
    author        = {Loick Chambon and Paul Couairon and Eloi Zablocki and Alexandre Boulch and Nicolas Thome and Matthieu Cord},
    year          = 2025,
    url           = {https://arxiv.org/abs/2511.18452},
}

@inproceedings{martingarcia2024diffusione2eft,
    title         = {Fine-Tuning Image-Conditional Diffusion Models is Easier than You Think},
    author        = {Martin Garcia, Gonzalo and Knaebel, Karim and Schmidt, Christian and de Geus, Daan and Hermans, Alexander and Leibe, Bastian},
    year          = 2025,
    booktitle     = WACV,
}

@inproceedings{wang2025moge2,
    title         = {{MoGe-2}: Accurate Monocular Geometry with Metric Scale and Sharp Details},
    author        = {Ruicheng Wang and Sicheng Xu and Yue Dong and Yu Deng and Jianfeng Xiang and Zelong Lv and Guangzhong Sun and Xin Tong and Jiaolong Yang},
    year          = 2025,
    booktitle     = NIPS,
}

@inproceedings{wang2025vggt,
    title         = {{VGGT}: Visual Geometry Grounded Transformer},
    author        = {Wang, Jianyuan and Chen, Minghao and Karaev, Nikita and Vedaldi, Andrea and Rupprecht, Christian and Novotny, David},
    year          = 2025,
    booktitle     = CVPR,
}

@inproceedings{wang2025moge,
    title         = {{MoGe}: Unlocking accurate monocular geometry estimation for open-domain images with optimal training supervision},
    author        = {Wang, Ruicheng and Xu, Sicheng and Dai, Cassie and Xiang, Jianfeng and Deng, Yu and Tong, Xin and Yang, Jiaolong},
    year          = 2025,
    booktitle     = CVPR,
}

@article{xu2025ppd,
    title         = {Pixel-perfect depth with semantics-prompted diffusion transformers},
    author        = {Xu, Gangwei and Lin, Haotong and Luo, Hongcheng and Wang, Xianqi and Yao, Jingfeng and Zhu, Lianghui and Pu, Yuechuan and Chi, Cheng and Sun, Haiyang and Wang, Bing and others},
    year          = 2025,
    journal       = {arXiv preprint arXiv:2510.07316},
}

@article{gomez2025urbansyn,
    title         = {All for one, and one for all: UrbanSyn Dataset, the third Musketeer of synthetic driving scenes},
    author        = {Jose L. Gómez and Manuel Silva and Antonio Seoane and Agnés Borràs and Mario Noriega and German Ros and Jose A. Iglesias-Guitian and Antonio M. López},
    year          = 2025,
    journal       = {Neurocomputing},
}

@inproceedings{zhu2025dyt,
    title         = {Transformers without normalization},
    author        = {Zhu, Jiachen and Chen, Xinlei and He, Kaiming and LeCun, Yann and Liu, Zhuang},
    year          = 2025,
    booktitle     = CVPR,
}

@inproceedings{yang2025fast3r,
    title         = {Fast3R: Towards 3D Reconstruction of 1000+ Images in One Forward Pass},
    author        = {Jianing Yang and Alexander Sax and Kevin J. Liang and Mikael Henaff and Hao Tang and Ang Cao and Joyce Chai and Franziska Meier and Matt Feiszli},
    year          = 2025,
    booktitle     = CVPR,
}

@inproceedings{wang2026vggtomega,
    title         = {{VGGT-$\Omega$}},
    author        = {Jianyuan Wang and Minghao Chen and Shangzhan Zhang and Nikita Karaev and Johannes Sch{\"o}nberger and Patrick Labatut and Piotr Bojanowski and David Novotny and Andrea Vedaldi and Christian Rupprecht},
    year          = 2026,
    booktitle     = CVPR,
}

@article{piccinelli2025unidepthv2,
    title         = {{U}ni{D}epth{V2}: Universal Monocular Metric Depth Estimation Made Simpler},
    author        = {Luigi Piccinelli and Christos Sakaridis and Yung-Hsu Yang and Mattia Segu and Siyuan Li and Wim Abbeloos and Luc Van Gool},
    year          = 2026,
    journal       = PAMI,
}

@inproceedings{yu2026infinidepth,
    title         = {{InfiniDepth}: Arbitrary-Resolution and Fine-Grained Depth Estimation with Neural Implicit Fields},
    author        = {Hao Yu and Haotong Lin and Jiawei Wang and Jiaxin Li and Yida Wang and Xueyang Zhang and Yue Wang and Xiaowei Zhou and Ruizhen Hu and Sida Peng},
    year          = 2026,
    booktitle     = CVPR,
}

@inproceedings{lin2026depthanything3,
    title         = {{Depth Anything 3}: recovering the visual space from any views},
    author        = {Haotong Lin and Sili Chen and Jun Hao Liew and Donny Y. Chen and Zhenyu Li and Guang Shi and Jiashi Feng and Bingyi Kang},
    year          = 2026,
    booktitle     = ICLR,
}

@inproceedings{keetha2026mapanything,
    title         = {{MapAnything}: Universal Feed-Forward Metric {3D} Reconstruction},
    author        = {Nikhil Keetha and Norman M\"{u}ller and Johannes Sch\"{o}nberger and Lorenzo Porzi and Yuchen Zhang and Tobias Fischer and Arno Knapitsch and Duncan Zauss and Ethan Weber and Nelson Antunes and Jonathon Luiten and Manuel Lopez-Antequera and Samuel Rota Bul\`{o} and Christian Richardt and Deva Ramanan and Sebastian Scherer and Peter Kontschieder},
    year          = 2026,
    booktitle     = \3DV,
}

@inproceedings{wang2026pi3,
    title         = {{$\pi^3$}: Permutation-Equivariant Visual Geometry Learning},
    author        = {Wang, Yifan and Zhou, Jianjun and Zhu, Haoyi and Chang, Wenzheng and Zhou, Yang and Li, Zizun and Chen, Junyi and Pang, Jiangmiao and Shen, Chunhua and He, Tong},
    year          = 2026,
    booktitle     = ICLR,
}

@inproceedings{fang2026dens3r,
    title         = {{Dens3R}: A Foundation Model for {3D} Geometry Prediction},
    author        = {Fang, Xianze and Gao, Jingnan and Wang, Zhe and Chen, Zhuo and Ren, Xingyu and Lyu, Jiangjing and Ren, Qiaomu and Yang, Zhonglei and Yang, Xiaokang and Yan, Yichao and Lyu, Chengfei},
    year          = 2026,
    booktitle     = ICLR,
}

@article{bian2026mda,
    title         = {Modeling Depth Ambiguity: A Mixture-Density Representation for Flying-Point-Free Depth Estimation},
    author        = {Siyuan Bian and Congrong Xu and Jun Gao},
    year          = 2026,
    journal       = {arXiv preprint arXiv:2606.02552},
}
}

\clearpage
\appendix
\section*{Appendix}
\label{appendix}

\renewcommand{\thetable}{\Alph{table}}
\setcounter{table}{0}
\renewcommand{\thefigure}{\Alph{figure}}
\setcounter{figure}{0}
\renewcommand{\thepseudocode}{\Alph{pseudocode}}
\setcounter{pseudocode}{0}

\section{Details on the Point Gradient Matching Loss}
\label{supp:pgm_loss_details}
\Cref{code:pgm_loss} gives a Python-like specification of $\mathcal{L}_{\mathrm{pgm}}$ in \cref{eq:pgm_loss}.
The loss optimizes the orientation and magnitude of local point gradients by matching 3D displacements; normalization by \(z\) makes the loss scale-invariant.
We compute \(\mathcal{L}_{\mathrm{pgm}}\) only on point pairs where all involved points are valid.

\begin{pseudocode}[\label{code:pgm_loss}]{Point gradient matching loss \(\mathcal{L}_{\mathrm{pgm}}\)}
def point_gradient_loss(pred, gt, valid):
  # pred, gt: [B, H, W, 3]
  # valid: [B, H, W]

  def pair_loss(pred0, pred1, gt0, gt1, valid0, valid1):
      valid_pair = valid0 & valid1

      pred_grad = (pred1 - pred0) / minimum(pred0[..., 2:], pred1[..., 2:])
      gt_grad = (gt1 - gt0) / minimum(gt0[..., 2:], gt1[..., 2:])

      loss = (pred_grad - gt_grad).norm(dim=-1)
      return (loss * valid_pair).sum() / valid_pair.sum()

  loss_dx = pair_loss(
      pred[:, :, :-1, :], pred[:, :, 1:, :],
      gt[:, :, :-1, :],   gt[:, :, 1:, :],
      valid[:, :, :-1],   valid[:, :, 1:],
  )
  loss_dy = pair_loss(
      pred[:, :-1, :, :], pred[:, 1:, :, :],
      gt[:, :-1, :, :],   gt[:, 1:, :, :],
      valid[:, :-1, :],   valid[:, 1:, :],
  )
  return (loss_dx + loss_dy) / 2
\end{pseudocode}

\section{Architectural Details}
\label{supp:arch_details}

\paragraph{Window-matched RoPE.}
In addition to the embedded UV coordinates added at the beginning of each decoder stage, we apply RoPE~\cite{su2024rope} to the queries and keys in each NAD block.
The UV grid provides absolute coordinates, while RoPE gives each attention head a consistent representation of relative offsets inside local windows, making local attention patterns easier to reuse across neighborhoods.
We adapt the RoPE base frequency to the kernel size where attention is computed.
For a Neighborhood Attention window size \(k\), we set the base frequency \(\tau=\sfrac{k}{\pi}\).
For head dimension \(d_h\), axial 2D RoPE uses \(M=\sfrac{d_h}{4}\) geometrically spaced frequency bands per spatial axis, with frequencies \(\omega_m=\tau^{-\sfrac{m}{M}}\) for \(m=0,\ldots,M-1\).
With this choice, the lowest-frequency band changes by \(\pi\) radians, \ie, half a rotation, over one neighborhood width, providing meaningful phase variation within each local window.
In preliminary experiments, RoPE with this window-matched base frequency improves accuracy over using only the UV embedding.

\paragraph{Normalization in NAD blocks.}
We remove the usual normalization layers before the attention and FFN blocks.
\citet{wang2025moge2} remove normalization layers from the ConvStack without observing instabilities or reduced accuracy.
In our attention decoder, removing LayerNorm~\cite{ba2016layernormalization} \emph{improves} accuracy, but occasionally leads to irrecoverable training instabilities.
We tried to replace LayerNorm with several alternative normalization approaches (RMSNorm~\cite{zhang2019rmsnorm}, DyT~\cite{zhu2025dyt}, and LayerScale~\cite{touvron2021layerscale}), and found that each of them recovers training stability, but none of them improves accuracy over LayerNorm.
We therefore suspect that per-token normalization can interfere with the unbounded regression task of 3D geometry estimation.
To stabilize training without compromising task performance, we keep the norm-free block structure and apply QK normalization~\cite{zhai2022scalingvisiontransformers,dehghani2023vit22b} in the NAD blocks.

\paragraph{Stage-wise UV embedding.}
Following \citet{wang2025moge}, we add a learned positional embedding to the features at the beginning of each stage in the NAD head.
This embedding is produced by linearly projecting a UV grid at the stage resolution, with coordinates normalized by the image diagonal.
It gives the decoder aspect-ratio-aware absolute position information for the current feature map.

\section{Training Details}
\label{supp:training_details}

\begin{figure}
    \centering
    \includegraphics[width=\linewidth]{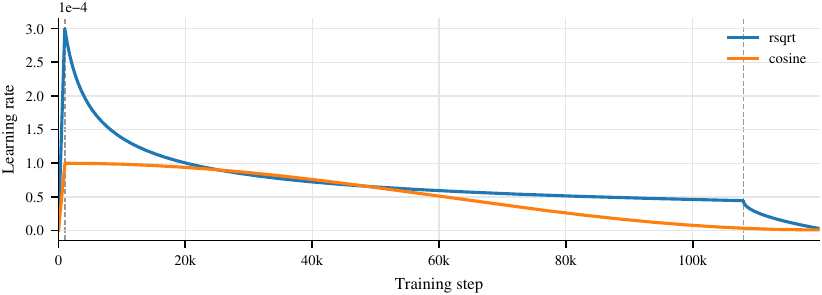}
    \caption{\textbf{Learning rate schedule.} We use a reciprocal square root learning rate schedule~\cite{zhai2022scalingvisiontransformers} (rsqrt) with a timescale of \num{2400}, a linear warmup of \num{1000} steps, and a \(10\%\) cooldown phase with schedule \(1-\sqrt{p} + \mathrm{lr}_\mathrm{min}\), where $p \in (0,1]$ is the progress of the cooldown phase and \(\mathrm{lr}_\mathrm{min}\) is the peak learning rate multiplied by \(10^{-2}\). Early training is stable enough to support a much larger peak learning rate than we can use with standard cosine decay. The rsqrt schedule uses this large initial rate to make rapid progress, then quickly decays to a moderate rate, whereas a cosine schedule at the same peak becomes unstable later in training. In our setting, rsqrt tends to outperform cosine decay even when the total number of training steps is known in advance, while still allowing training to continue indefinitely before cooldown.}
    \label{fig:lr_schedule}
\end{figure}

All main models and ablations are trained for \num{120000} optimizer steps with total batch size \num{128}.
We use AdamW~\cite{loshchilov2019adamw} with peak learning rate \(3\times10^{-4}\) for the decoder and \(3\times10^{-5}\) for the DINOv2~\cite{oquab2023dinov2} backbone, gradient clipping at \num{1.0}, and the reciprocal square root schedule shown in~\cref{fig:lr_schedule}.
The decoder learning rate is linearly warmed up for \num{1000} steps, during which the backbone remains frozen.
After unfreezing, the backbone learning rate is also warmed up for \num{1000} steps, ending at \num{0.1} times the decoder schedule evaluated at step \num{2000}; after this point, it remains phase-aligned with the decoder schedule at the same \num{0.1} scale.
Training uses bf16 mixed precision for the backbone, while the point decoder, output remapping, and losses are evaluated in fp32.

For the first \(80\%\) of training, we sample images at a fixed area of \(512^2\) pixels with target aspect ratios in \([0.5,2.0]\), and use a low-resolution encoder-token budget of \num{1024}.
The final \(20\%\) of steps sample image areas uniformly from \([512^2,960^2]\) pixels and encoder-token budgets uniformly from \(\{1024,\ldots,2802\}\).
The cooldown phase in the final \(10\%\) samples only high-quality synthetic data: all synthetic training datasets except Ken~Burns~\cite{niklaus2019kenburns} and G-Objaverse~\cite{deitke2022objaverse,qiu2024richdreamer}.

Following common practice~\cite{wang2025moge,bochkovskii2024depthpro}, we adapt supervision to label quality by selecting loss terms by label type:
\[
\begin{aligned}
    \mathcal{L}_\mathrm{synthetic}
    &=
    \mathcal{L}_\mathrm{glob}
    + \mathcal{L}_{\mathrm{loc},4}
    + \mathcal{L}_{\mathrm{loc},16}
    + \mathcal{L}_{\mathrm{loc},64}
    + 10\mathcal{L}_\mathrm{pgm},\\
    \mathcal{L}_\mathrm{sfm}
    &=
    \mathcal{L}_\mathrm{glob}
    + \mathcal{L}_{\mathrm{loc},4}
    + \mathcal{L}_{\mathrm{loc},16},\\
    \mathcal{L}_\mathrm{lidar}
    &=
    \mathcal{L}_\mathrm{glob}
    + \mathcal{L}_{\mathrm{loc},4}.
\end{aligned}
\]

\section{Training Data}
\label{supp:training_data}
We train on a subset of the datasets used by MoGe-2~\cite{wang2025moge2}; the full list is provided in~\cref{tab:training_data}.
All datasets used are publicly available for academic use.

\begin{table}
    \centering
    \caption{\textbf{List of training datasets.} The mix and weighting are similar to MoGe-2~\cite{wang2025moge2}. Sampling probabilities are proportional to the listed weights.}
    \renewcommand{\arraystretch}{1.2}
    \begin{tabularx}{\linewidth}{@{}Llcl S[table-format=2.1]@{}}
        \toprule
        Name & Domain & Frames & Type & {Weight} \\
        \midrule
        Argoverse2~\cite{wilson2021argoverse2} & Outdoor/Driving & \num{2.2}M & LiDAR & 7.4\\

        ARKitScenes~\cite{baruch2021arkitscenes} & Indoor & \num{449}K & SfM & 8.6\\

        BlendedMVS~\cite{yao2020blendedmvs} & In-the-wild & \num{112}K & SfM & 12.0 \\

        G-Objaverse~\cite{deitke2022objaverse,qiu2024richdreamer} & Object & \num{894}K & Synthetic & 4.8\\

        GTA-SfM~\cite{wang2020gtasfm} & Outdoor/In-the-wild & \num{18}K & Synthetic & 2.8\\

        Hypersim~\cite{roberts2021hypersim} & Indoor & \num{67}K & Synthetic & 5.0\\

        IRS~\cite{wang2021irs} & Indoor & \num{48}K & Synthetic & 5.6\\

        Ken~Burns~\cite{niklaus2019kenburns} & In-the-wild & \num{304}K & Synthetic & 1.6\\

        MatrixCity~\cite{li2023matrixcity} & Outdoor/Driving & \num{392}K & Synthetic & 1.3\\

        MegaDepth~\cite{li2018megadepth} & Outdoor/In-the-wild & \num{103}K & SfM & 5.6\\

        MidAir~\cite{fonder2019midair} & Outdoor/In-the-wild & \num{391}K & Synthetic & 4.0\\

        MVS-Synth~\cite{huang2018mvssynth} & Outdoor/Driving & \num{12}K & Synthetic & 1.2\\

        Structured3D~\cite{zheng2020structured3d} & Indoor & \num{71}K & Synthetic & 4.8\\

        Synthia-AL~\cite{bengar2019synthiaAL} & Outdoor/Driving & \num{142}K & Synthetic & 1.1\\

        Synthia-SF~\cite{hernandez2017synthiaSF} & Outdoor/Driving & \num{2}K & Synthetic & 0.1\\

        TartanAir~\cite{wang2020tartanairV1} & In-the-wild & \num{586}K & Synthetic & 5.0\\

        Taskonomy~\cite{zamir2018taskonomy} & Indoor & \num{4.5}M & SfM & 14.1\\

        UnrealStereo4K~\cite{tosi2021smdnets} & In-the-wild & \num{14}K & Synthetic & 1.6 \\

        UrbanSyn~\cite{gomez2025urbansyn} & Outdoor/Driving & \num{8}K & Synthetic & 2.1\\

        Waymo~\cite{sun2020waymo} & Outdoor/Driving & \num{1}M & LiDAR & 6.4\\
        \bottomrule
    \end{tabularx}
    \label{tab:training_data}
\end{table}

\section{Decoder Runtime Tradeoff}
\label{supp:decoder_runtime}
NAD improves local surface quality over the convolutional decoder baselines at the cost of higher inference latency.
\Cref{tab:nad_convl_speed} quantifies this runtime tradeoff at \(512{\times}512\) resolution.
The decoder alone is \(2.28{\times}\) slower than ConvStack-L.
In the full model, the slowdown is smaller: \(1.30{\times}\) with DINOv2-giant and \(1.46{\times}\) with DINOv2-Large, since the shared encoder accounts for part of total inference time.
Peak memory increases modestly with NAD.
Thus, NAD is an accuracy-oriented decoder design, and improving the efficiency of Neighborhood Attention decoding is an important direction for future work.

\begin{table}
	\centering
	\caption{
		\textbf{Runtime comparison with ConvStack-L.}
		Inference latency is reported on an NVIDIA H100 GPU as the median over \num{50} timed iterations at $512{\times}512$ resolution and batch size \num{1}, after \num{10} warmup iterations and with CUDA synchronization.
		All benchmarks use \texttt{torch.compile} with dynamic shapes.
		The encoders are run with \texttt{bf16} autocast, while decoders are evaluated in \texttt{fp32} to preserve output quality.
		Peak memory is measured with PyTorch \texttt{max\_memory\_allocated}.
	}
	\label{tab:nad_convl_speed}
	\small
	\setlength{\tabcolsep}{3.5pt}
	\sisetup{table-column-width=70pt}
	\renewcommand{\arraystretch}{1.2}
	\begin{tabularx}{\linewidth}{@{}LL
		S[table-format=2.2]
		S[table-format=1.2]
		S[table-format=1.2]@{}}
		\toprule
		Encoder & Decoder
		& {\begin{tabular}[c]{@{}c@{}}Latency\\(ms) $\downarrow$\end{tabular}}
		& {\begin{tabular}[c]{@{}c@{}}Relative latency\\($\times$) $\downarrow$\end{tabular}}
		& {\begin{tabular}[c]{@{}c@{}}Peak memory\\(GiB) $\downarrow$\end{tabular}} \\
		\midrule
		\multirow{2}{*}{DINOv2-giant} & ConvStack-L & 20.89 & 1.00 & 5.02 \\
		& \textbf{NAD (ours)} & 27.15 & 1.30 & 5.23 \\
		\midrule
		\multirow{2}{*}{DINOv2-Large} & ConvStack-L & 14.53 & 1.00 & 1.87 \\
		& \textbf{NAD (ours)} & 21.15 & 1.46 & 2.02 \\
		\midrule
		\multirow{2}{*}{None} & ConvStack-L & 5.94 & 1.00 & 0.72 \\
		& \textbf{NAD (ours)} & 13.56 & 2.28 & 0.88 \\
		\bottomrule
	\end{tabularx}
\end{table}

\clearpage

\section{Additional Quantitative Results}
\label{supp:additional_quantitative}

For direct comparison with depth estimation methods, \cref{tab:supp:depth-affine-invariant-rel,tab:supp:depth-affine-invariant-delta} report affine-invariant depth \AbsRel{} and $\delta_1$ on the same eight benchmarks as the main tables.
\Cref{tab:supp:points-affine-invariant-delta} reports affine-invariant point map accuracy $\delta_{0.05}$, complementing \AbsRelGlob{} in \cref{tab:main:points-affine-invariant-rel}.
Method order and footnote markers in the full comparisons follow \cref{tab:main:points-affine-invariant-rel}.

\Cref{tab:supp:boundary-f1} reports depth boundary $F_1$~\cite{bochkovskii2024depthpro} on iBims-1, Sintel, HAMMER~\cite{jung2023hammer}, and Spring~\cite{mehl2023spring}.
Because boundary $F_1$ is highly sensitive to output resolution, \cref{tab:supp:boundary-f1-matched} also compares MoGe, MoGe-2, and \ours{} at a matched resolution of \num{2551} encoder tokens.

Finally, \cref{tab:supp:direct-normal} reports mean angular error for three direct normal estimators alongside \ours{} to put the magnitude of our errors in context.
The results are not directly comparable because the direct estimators use dataset normal annotations as ground truth, while \ours{} uses normals derived from ground-truth point maps (\MAEnormal{}).

\begin{table}[htbp]
\centering
\small
\setlength{\tabcolsep}{2.25pt}
\sisetup{table-column-width=34pt}
\caption{\textbf{SotA comparison for affine-invariant depth \AbsRel{} (\%, $\downarrow$).}}
\label{tab:supp:depth-affine-invariant-rel}
\renewcommand{\arraystretch}{1.2}
\begin{tabularx}{\linewidth}{@{}L *{8}{S[table-format=2.2]}@{}}
\toprule
Method & {NYUv2} & {KITTI} & {ETH3D} & {iBims-1} & {GSO} & {Sintel} & {DDAD} & {DIODE} \\
\midrule
DUSt3R~\cite{wang2024dust3r} & 3.72 & 7.28 & 5.08 & 3.94 & 3.30 & 25.44 & 16.85 & 6.68 \\
Depth Pro~\cite{bochkovskii2024depthpro} & 3.67 & 5.12 & 4.97 & 3.23 & 1.46 & 15.82 & 12.61 & 4.66 \\
MapAnything~\cite{keetha2026mapanything} & 3.67 & 4.85 & 4.09 & 3.37 & 1.44 & 16.58 & 13.04 & 4.46 \\
VGGT~\cite{wang2025vggt} & 3.03 & 5.36 & 3.43 & 3.37 & \cellcolor{tablefirst}\bfseries 0.81 & 18.15 & 13.90 & 5.07 \\
InfiniDepth~\cite{yu2026infinidepth}\textsuperscript{*} & 3.98 & 5.09 & 4.65 & 3.14 & 1.85 & 14.27 & 12.16 & 4.90 \\
VGGT-$\Omega$~\cite{wang2026vggtomega}\textsuperscript{\textdagger} & 3.45 & 4.73 & 2.82 & 3.26 & 1.20 & 15.05 & 10.67 & 4.27 \\
Depth Anything 3~\cite{lin2026depthanything3} & 3.28 & 5.24 & 3.94 & 2.81 & 1.33 & 17.28 & 12.23 & 4.78 \\
PPD~\cite{xu2025ppd}\textsuperscript{*} & 4.29 & 10.29 & 6.75 & 4.43 & 1.32 & 16.20 & 16.78 & 6.12 \\
UniDepthV2~\cite{piccinelli2025unidepthv2} & 2.96 & \cellcolor{tablesecond}3.86 & 2.95 & 2.64 & 1.37 & \cellcolor{tablesecond}13.33 & 10.48 & 4.05 \\
$\pi^3$~\cite{wang2026pi3} & 2.95 & 4.29 & \cellcolor{tablethird}2.70 & 2.83 & 1.08 & 14.04 & 9.76 & 3.63 \\
MoGe-2~\cite{wang2025moge2} & \cellcolor{tablesecond}2.89 & \cellcolor{tablefirst}\bfseries 3.75 & 2.80 & \cellcolor{tablesecond}2.36 & 0.94 & \cellcolor{tablethird}13.37 & \cellcolor{tablesecond}8.26 & \cellcolor{tablefirst}\bfseries 3.14 \\
MoGe~\cite{wang2025moge} & \cellcolor{tablethird}2.91 & \cellcolor{tablethird}3.92 & \cellcolor{tablesecond}2.69 & \cellcolor{tablethird}2.53 & \cellcolor{tablethird}0.94 & \cellcolor{tablefirst}\bfseries 13.03 & \cellcolor{tablethird}8.40 & \cellcolor{tablesecond}3.16 \\
\textbf{\ours{}} & \cellcolor{tablefirst}\bfseries 2.85 & 4.00 & \cellcolor{tablefirst}\bfseries 2.64 & \cellcolor{tablefirst}\bfseries 2.34 & \cellcolor{tablesecond}0.87 & 13.41 & \cellcolor{tablefirst}\bfseries 7.77 & \cellcolor{tablethird}3.60 \\
\bottomrule
\end{tabularx}
\end{table}

\begin{table}[htbp]
\centering
\small
\setlength{\tabcolsep}{2.25pt}
\sisetup{table-column-width=34pt}
\caption{\textbf{SotA comparison for affine-invariant depth $\delta_1$ (\%, $\uparrow$).}}
\label{tab:supp:depth-affine-invariant-delta}
\renewcommand{\arraystretch}{1.2}
\begin{tabularx}{\linewidth}{@{}L *{8}{S[table-format=2.2]}@{}}
\toprule
Method & {NYUv2} & {KITTI} & {ETH3D} & {iBims-1} & {GSO} & {Sintel} & {DDAD} & {DIODE} \\
\midrule
DUSt3R~\cite{wang2024dust3r} & 97.86 & 91.57 & 96.24 & 96.57 & 99.23 & 64.05 & 76.35 & 92.52 \\
Depth Pro~\cite{bochkovskii2024depthpro} & 98.20 & 96.76 & 96.43 & 98.33 & 99.99 & 80.12 & 84.08 & 95.60 \\
MapAnything~\cite{keetha2026mapanything} & 98.20 & 97.01 & 97.54 & 97.82 & 99.95 & 78.72 & 84.77 & 95.93 \\
VGGT~\cite{wang2025vggt} & 98.40 & 95.06 & 97.81 & 97.18 & 99.99 & 76.09 & 81.23 & 94.63 \\
InfiniDepth~\cite{yu2026infinidepth}\textsuperscript{*} & 97.77 & 96.31 & 96.42 & \cellcolor{tablethird}98.48 & 99.95 & 81.88 & 86.15 & 95.28 \\
VGGT-$\Omega$~\cite{wang2026vggtomega}\textsuperscript{\textdagger} & 98.17 & 96.81 & 98.81 & 97.64 & 99.95 & 80.65 & 86.92 & 95.83 \\
Depth Anything 3~\cite{lin2026depthanything3} & 98.14 & 96.28 & 96.62 & 98.09 & 99.98 & 76.95 & 84.81 & 94.46 \\
PPD~\cite{xu2025ppd}\textsuperscript{*} & 97.97 & 85.04 & 93.14 & 97.02 & 99.99 & 77.92 & 74.76 & 92.76 \\
UniDepthV2~\cite{piccinelli2025unidepthv2} & \cellcolor{tablefirst}\bfseries 98.64 & \cellcolor{tablesecond}98.08 & 98.53 & 98.36 & 99.99 & \cellcolor{tablethird}83.18 & 90.89 & 96.49 \\
$\pi^3$~\cite{wang2026pi3} & 98.48 & 97.79 & 98.82 & 98.11 & 99.98 & 81.30 & 89.49 & 96.80 \\
MoGe-2~\cite{wang2025moge2} & 98.60 & \cellcolor{tablefirst}\bfseries 98.10 & \cellcolor{tablesecond}99.05 & \cellcolor{tablefirst}\bfseries 98.75 & \cellcolor{tablesecond}99.99 & 82.53 & \cellcolor{tablefirst}\bfseries 92.49 & \cellcolor{tablesecond}97.44 \\
MoGe~\cite{wang2025moge} & \cellcolor{tablesecond}98.64 & \cellcolor{tablethird}98.03 & \cellcolor{tablefirst}\bfseries 99.17 & 98.43 & \cellcolor{tablethird}99.99 & \cellcolor{tablesecond}83.23 & \cellcolor{tablethird}92.20 & \cellcolor{tablefirst}\bfseries 97.47 \\
\textbf{\ours{}} & \cellcolor{tablethird}98.63 & 97.88 & \cellcolor{tablethird}99.04 & \cellcolor{tablesecond}98.71 & \cellcolor{tablefirst}\bfseries 99.99 & \cellcolor{tablefirst}\bfseries 83.47 & \cellcolor{tablesecond}92.27 & \cellcolor{tablethird}97.02 \\
\bottomrule
\end{tabularx}
\end{table}

\begin{table}[htbp]
\centering
\begin{minipage}{\textwidth}
\centering
\small
\setlength{\tabcolsep}{2.25pt}
\sisetup{table-column-width=34pt}
\caption{\textbf{SotA comparison for affine-invariant point map $\delta_{0.05}$ (\%, $\uparrow$).}}
\label{tab:supp:points-affine-invariant-delta}
\renewcommand{\arraystretch}{1.2}
\begin{tabularx}{\linewidth}{@{}L *{8}{S[table-format=2.2]}@{}}
\toprule
Method & {NYUv2} & {KITTI} & {ETH3D} & {iBims-1} & {GSO} & {Sintel} & {DDAD} & {DIODE} \\
\midrule
DUSt3R~\cite{wang2024dust3r} & 75.14 & 22.72 & 59.12 & 72.67 & 69.30 & 21.64 & 22.91 & 64.16 \\
Depth Pro~\cite{bochkovskii2024depthpro} & 75.08 & 37.79 & 57.56 & 75.01 & 79.75 & 26.88 & 32.97 & 67.80 \\
MapAnything~\cite{keetha2026mapanything} & 73.55 & 38.57 & 67.94 & 75.55 & 92.71 & 26.86 & 31.67 & 69.49 \\
VGGT~\cite{wang2025vggt} & 81.02 & 44.47 & 72.39 & 72.66 & \cellcolor{tablefirst}\bfseries 98.06 & 29.85 & 22.80 & 73.11 \\
InfiniDepth~\cite{yu2026infinidepth}\textsuperscript{*} & 75.81 & 62.53 & 56.09 & 74.63 & 85.51 & 16.06 & 28.41 & 68.41 \\
VGGT-$\Omega$~\cite{wang2026vggtomega}\textsuperscript{\textdagger} & 60.97 & 51.84 & 75.83 & 66.72 & 94.16 & \cellcolor{tablethird}32.42 & 34.52 & 69.87 \\
Depth Anything 3~\cite{lin2026depthanything3} & 81.48 & 62.77 & 74.20 & 80.30 & 93.22 & 28.31 & 29.09 & 73.17 \\
PPD~\cite{xu2025ppd}\textsuperscript{*} & 80.46 & 52.02 & 74.51 & 77.33 & 96.02 & 31.48 & 38.68 & 71.80 \\
UniDepthV2~\cite{piccinelli2025unidepthv2} & 82.62 & \cellcolor{tablefirst}\bfseries 73.61 & \cellcolor{tablethird}76.73 & 77.87 & 81.90 & 27.75 & \cellcolor{tablesecond}44.90 & 61.98 \\
$\pi^3$~\cite{wang2026pi3} & \cellcolor{tablethird}83.51 & 47.20 & 68.92 & 79.65 & 94.70 & 30.45 & 35.33 & \cellcolor{tablethird}75.15 \\
MoGe-2~\cite{wang2025moge2} & \cellcolor{tablesecond}84.20 & 53.69 & 76.68 & \cellcolor{tablethird}80.47 & \cellcolor{tablesecond}97.82 & \cellcolor{tablesecond}33.34 & \cellcolor{tablethird}43.74 & 74.00 \\
MoGe~\cite{wang2025moge} & 81.68 & \cellcolor{tablesecond}72.74 & \cellcolor{tablesecond}79.63 & \cellcolor{tablefirst}\bfseries 83.95 & 97.61 & 32.16 & 38.35 & \cellcolor{tablefirst}\bfseries 78.24 \\
\textbf{\ours{}} & \cellcolor{tablefirst}\bfseries 84.51 & \cellcolor{tablethird}70.56 & \cellcolor{tablefirst}\bfseries 80.12 & \cellcolor{tablesecond}83.08 & \cellcolor{tablethird}97.78 & \cellcolor{tablefirst}\bfseries 34.16 & \cellcolor{tablefirst}\bfseries 48.05 & \cellcolor{tablesecond}77.66 \\
\bottomrule
\end{tabularx}
\end{minipage}
\par
\addvspace{\floatsep}
\nointerlineskip
\begin{minipage}{0.85\textwidth}
\centering
\small
\setlength{\tabcolsep}{2.25pt}
\sisetup{table-column-width=50pt}
\caption{\textbf{SotA comparison for depth boundary $F_1$ (\%, $\uparrow$).}}
\label{tab:supp:boundary-f1}
\renewcommand{\arraystretch}{1.2}
\begin{tabularx}{\linewidth}{@{}L *{4}{S[table-format=2.2]}@{}}
\toprule
Method & {iBims-1} & {Sintel} & {HAMMER} & {Spring} \\
\midrule
DUSt3R~\cite{wang2024dust3r} & 3.38 & 2.75 & 0.23 & 0.20 \\
Depth Pro~\cite{bochkovskii2024depthpro} & 14.30 & \cellcolor{tablesecond}41.34 & 5.36 & \cellcolor{tablesecond}11.05 \\
MapAnything~\cite{keetha2026mapanything} & 5.50 & 6.85 & 1.14 & 4.80 \\
VGGT~\cite{wang2025vggt} & 3.49 & 13.23 & 0.63 & 4.04 \\
InfiniDepth~\cite{yu2026infinidepth}\textsuperscript{*} & \cellcolor{tablesecond}18.04 & \cellcolor{tablefirst}\bfseries 42.54 & \cellcolor{tablefirst}\bfseries 7.76 & \cellcolor{tablefirst}\bfseries 12.60 \\
VGGT-$\Omega$~\cite{wang2026vggtomega}\textsuperscript{\textdagger} & 4.55 & 15.01 & 0.12 & 5.09 \\
Depth Anything 3~\cite{lin2026depthanything3} & 9.55 & 15.94 & 1.69 & 5.25 \\
PPD~\cite{xu2025ppd}\textsuperscript{*} & 8.58 & 17.12 & 2.24 & 7.38 \\
UniDepthV2~\cite{piccinelli2025unidepthv2} & 11.15 & \cellcolor{tablethird}39.68 & 4.41 & 7.08 \\
$\pi^3$~\cite{wang2026pi3} & 6.07 & 18.58 & 0.66 & 5.09 \\
MoGe-2~\cite{wang2025moge2} & \cellcolor{tablethird}17.89 & 35.23 & \cellcolor{tablethird}5.42 & \cellcolor{tablethird}8.63 \\
MoGe~\cite{wang2025moge} & 14.91 & 24.95 & 5.15 & 7.89 \\
\textbf{\ours{}} & \cellcolor{tablefirst}\bfseries 18.85 & 32.87 & \cellcolor{tablesecond}5.44 & 8.40 \\
\bottomrule
\end{tabularx}
\end{minipage}
\par
\addvspace{\floatsep}
\nointerlineskip
\begin{minipage}{0.85\textwidth}
\centering
\small
\setlength{\tabcolsep}{2.25pt}
\sisetup{table-column-width=50pt}
\caption{\textbf{Depth boundary $F_1$ at matched resolution (\%, $\uparrow$).} All models use MoGe's resolution of \num{2551} encoder tokens, the highest common supported setting.}
\label{tab:supp:boundary-f1-matched}
\renewcommand{\arraystretch}{1.2}
\begin{tabularx}{\linewidth}{@{}L *{4}{S[table-format=2.2]}@{}}
\toprule
Method & {iBims-1} & {Sintel} & {HAMMER} & {Spring} \\
\midrule
MoGe-2~\cite{wang2025moge2} & \cellcolor{tablesecond}17.19 & \cellcolor{tablesecond}32.08 & \cellcolor{tablethird}4.47 & \cellcolor{tablethird}7.31 \\
MoGe~\cite{wang2025moge} & \cellcolor{tablethird}14.91 & \cellcolor{tablethird}24.95 & \cellcolor{tablesecond}5.15 & \cellcolor{tablesecond}7.89 \\
\textbf{\ours{}} & \cellcolor{tablefirst}\bfseries 18.59 & \cellcolor{tablefirst}\bfseries 32.29 & \cellcolor{tablefirst}\bfseries 5.33 & \cellcolor{tablefirst}\bfseries 8.15 \\
\bottomrule
\end{tabularx}
\end{minipage}
\par
\addvspace{\floatsep}
\nointerlineskip
\begin{minipage}{0.85\textwidth}
\centering
\small
\setlength{\tabcolsep}{2.25pt}
\sisetup{table-column-width=50pt}
\caption{\textbf{Comparison with direct normal prediction.} Mean angular error (\si{\degree}, $\downarrow$). Results are not directly comparable: DSINE, E2E-FT, and Dens3R use dataset normal annotations as ground truth, while \ours{} uses normals derived from ground-truth point maps (\MAEnormal{}).}
\label{tab:supp:direct-normal}
\renewcommand{\arraystretch}{1.2}
\begin{tabularx}{\linewidth}{@{}Ll *{2}{S[table-format=2.1]}@{}}
\toprule
Method & Normal source (prediction / ground truth) & {iBims-1} & {Sintel} \\
\midrule
DSINE~\cite{bae2024dsine} & direct / dataset & 17.1 & 34.9 \\
E2E-FT~\cite{martingarcia2024diffusione2eft} & direct / dataset & 15.8 & 33.5 \\
Dens3R~\cite{fang2026dens3r} & direct / dataset & 16.0 & 30.7 \\
\midrule
\textbf{\ours{}} & point map / point map & 16.5 & 24.5 \\
\bottomrule
\end{tabularx}
\end{minipage}
\end{table}

\clearpage

\section{Additional Qualitative Examples}
\label{supp:qualitative}

\begin{figure}[p]
  \centering
  \begingroup
  \newcommand{\sotarowheight}{0.115\textheight}
  \newcommand{\sotamethodwidth}{0pt}

  \newcommand{\sotamethod}[1]{\parbox[c][\sotarowheight][c]{\sotamethodwidth}{\centering
      \rotatebox[origin=c]{90}{\footnotesize\strut #1}}}

  \newcommand{\sotapanel}[1]{\adjustbox{valign=c}{\includegraphics[height=\sotarowheight,width=\linewidth,keepaspectratio]{#1}}}

  \newcommand{\sotaheader}[1]{\makebox[\linewidth][c]{\footnotesize #1}}

  \setlength{\tabcolsep}{1pt}

  \begin{adjustbox}{max totalsize={\linewidth}{0.99\textheight},center}
  \begin{tabular*}{\linewidth}{@{}
                  >{\centering\arraybackslash}m{\sotamethodwidth}
                  @{\extracolsep{\fill}}
                  >{\centering\arraybackslash}m{0.29\linewidth}
                  >{\centering\arraybackslash}m{0.29\linewidth}
                  >{\centering\arraybackslash}m{0.29\linewidth}
                  @{}}
    &
    \adjustbox{valign=b}{\includegraphics[height=\sotarowheight,width=\linewidth,keepaspectratio]{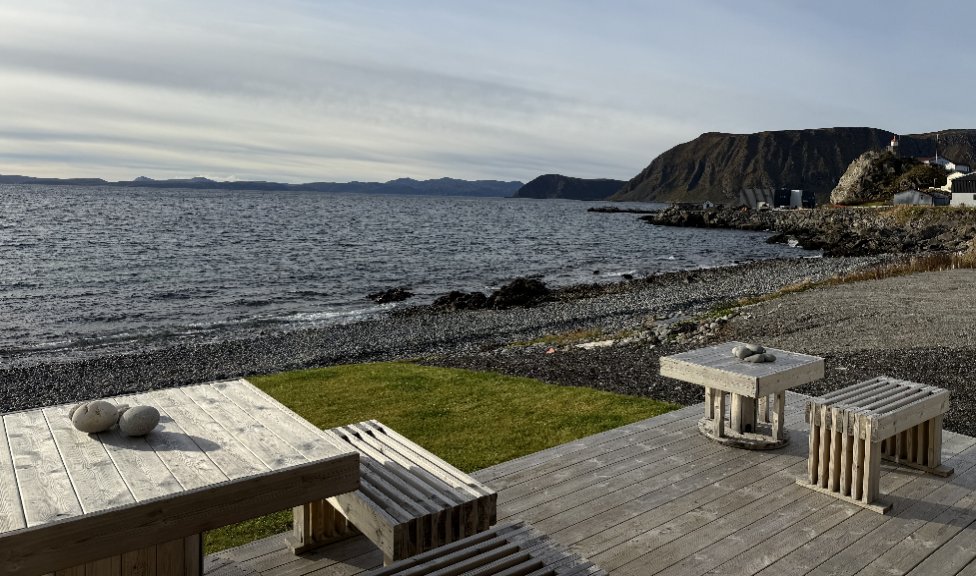}} &
    \sotaheader{Predicted Point Map} &
    \sotaheader{Near View} \\
    \addlinespace[2pt]

    \sotamethod{VGGT~\cite{wang2025vggt}} &
    \sotapanel{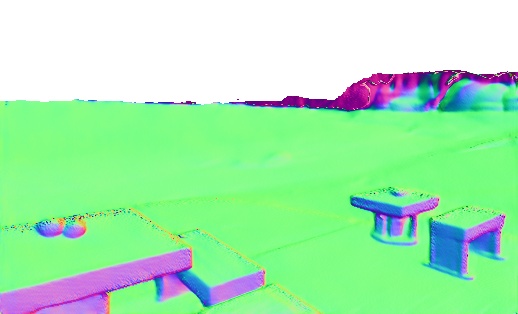} &
    \sotapanel{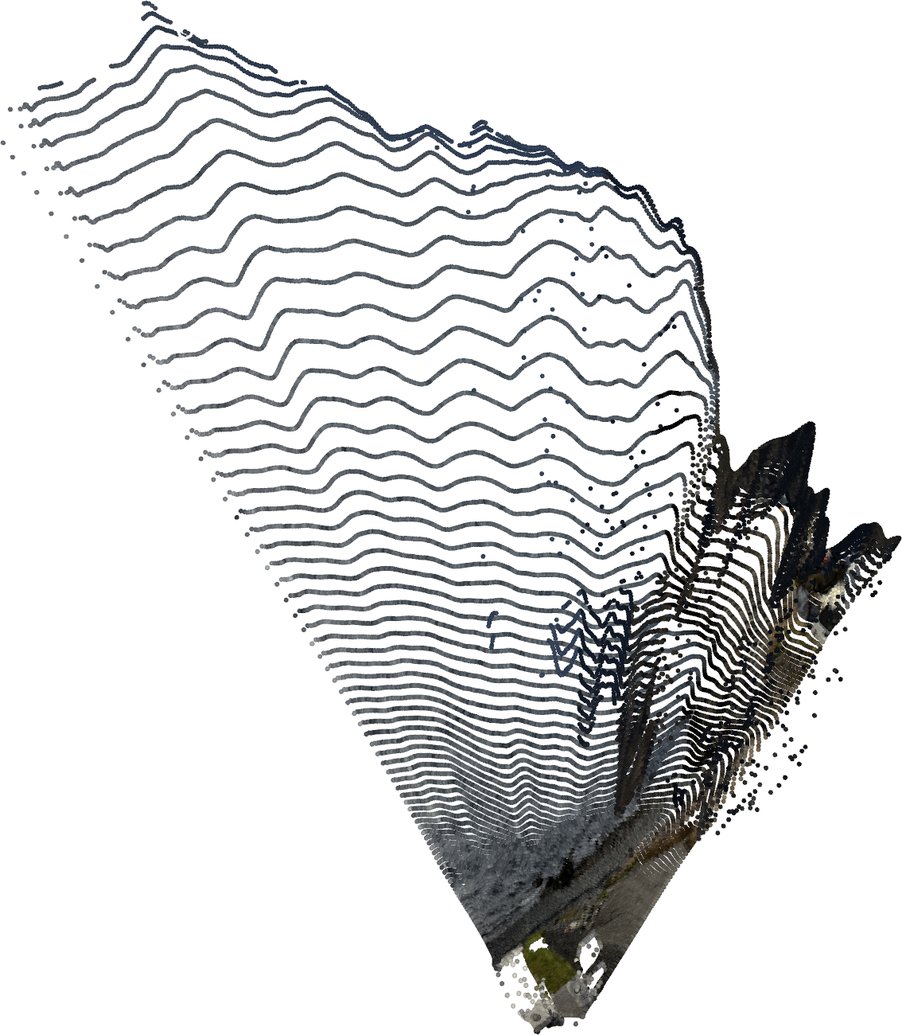} &
    \sotapanel{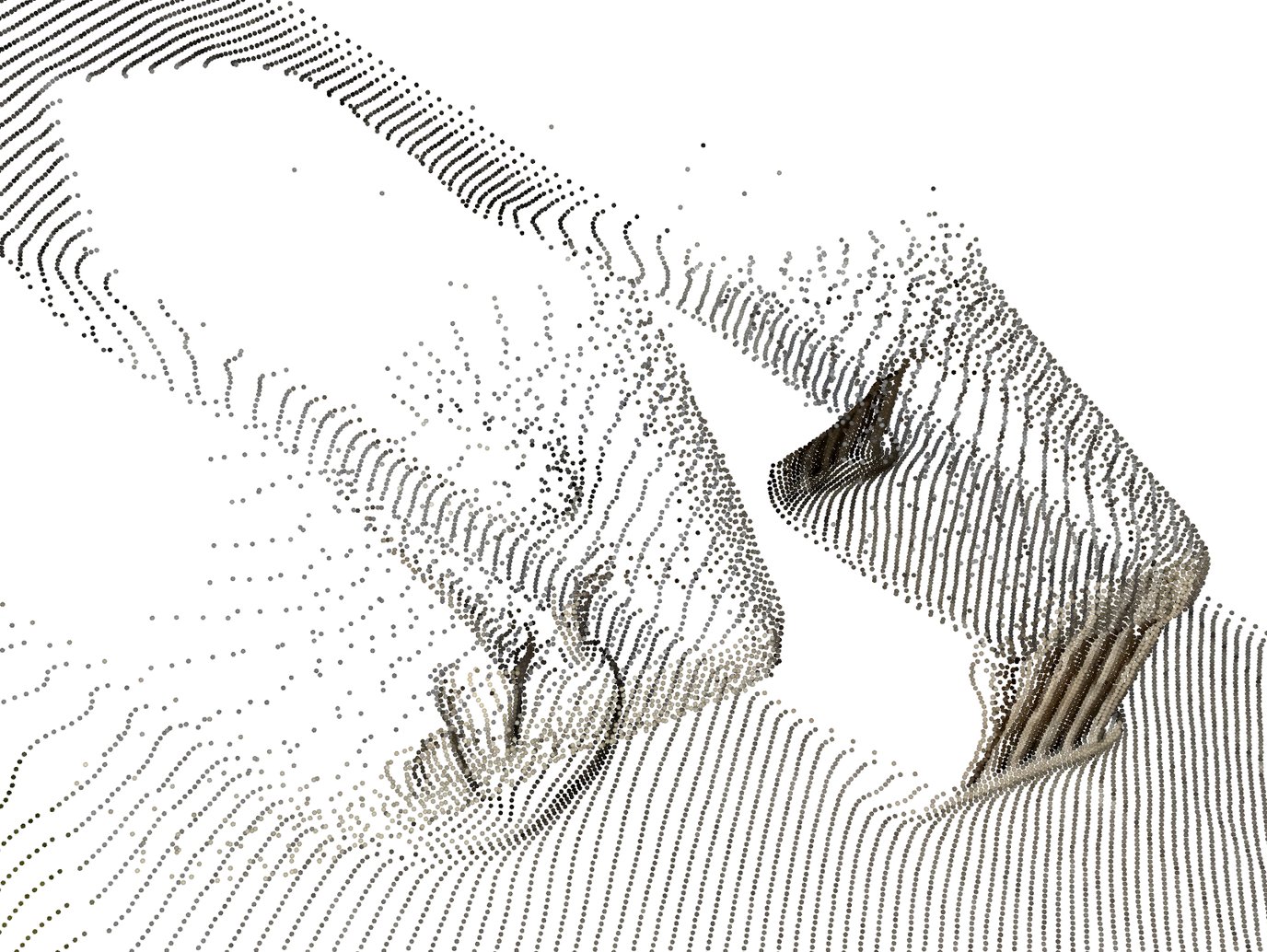} \\
    \addlinespace[2pt]

    \sotamethod{InfiniDepth~\cite{yu2026infinidepth}} &
    \sotapanel{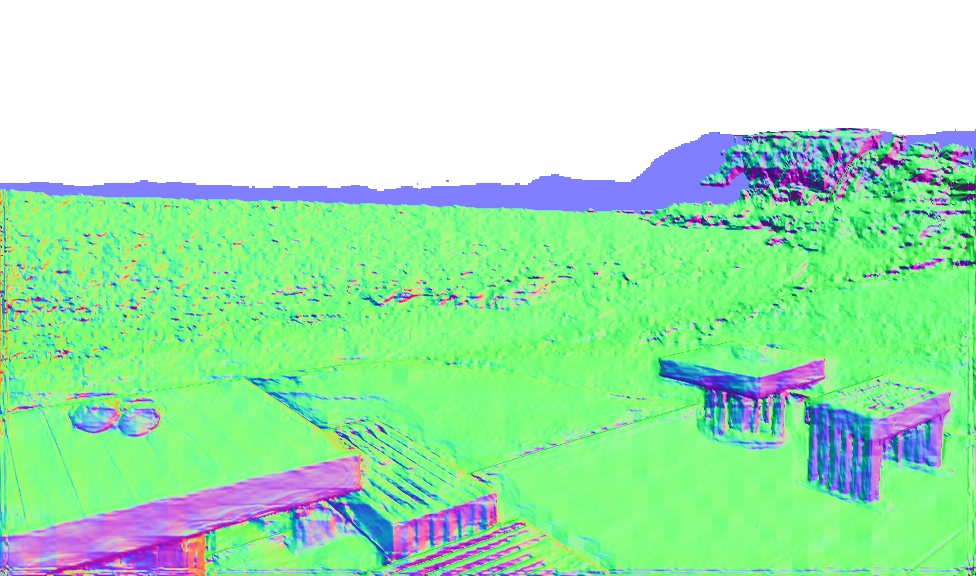} &
    \sotapanel{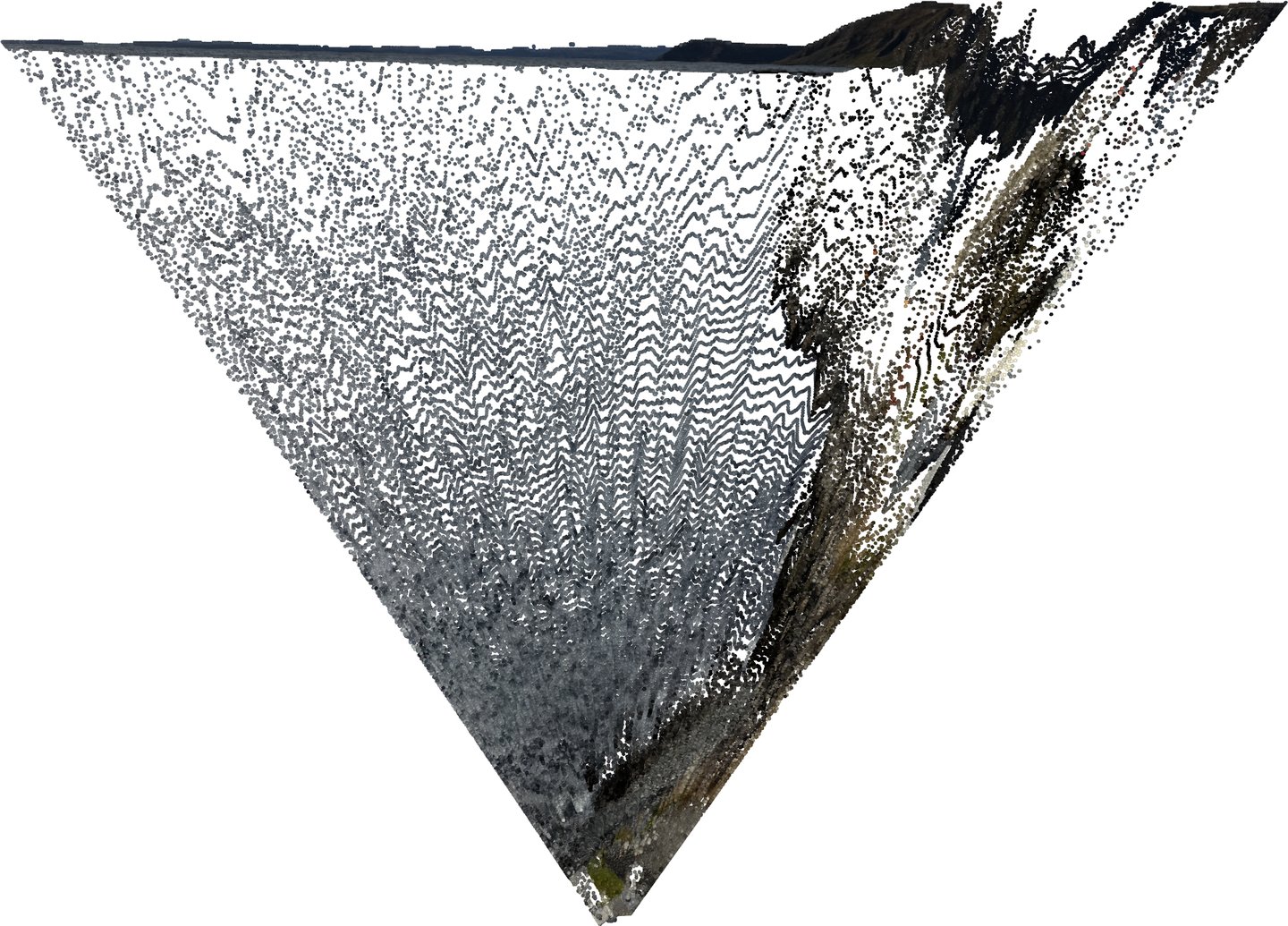} &
    \sotapanel{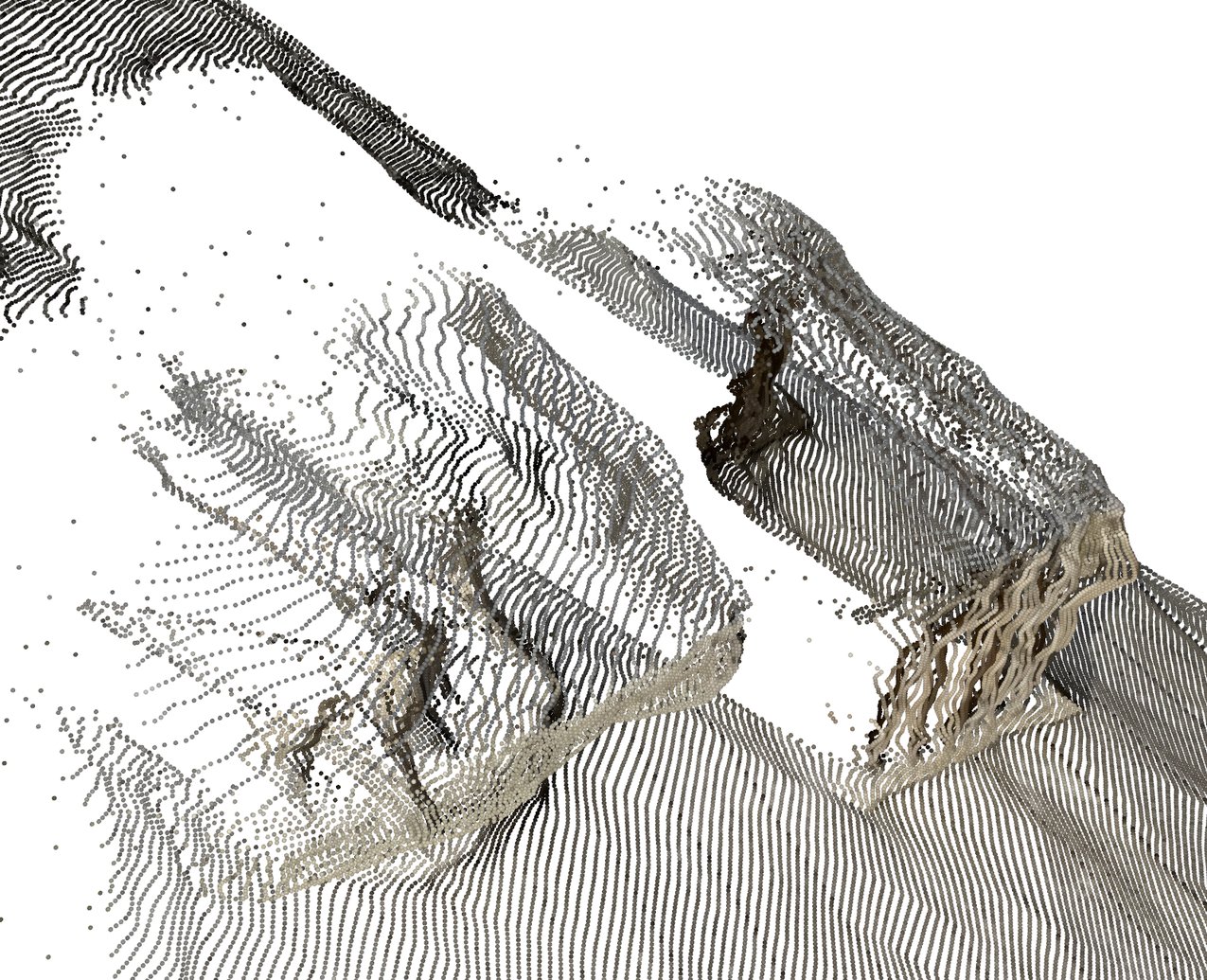} \\
    \addlinespace[2pt]

    \sotamethod{PPD~\cite{xu2025ppd}} &
    \sotapanel{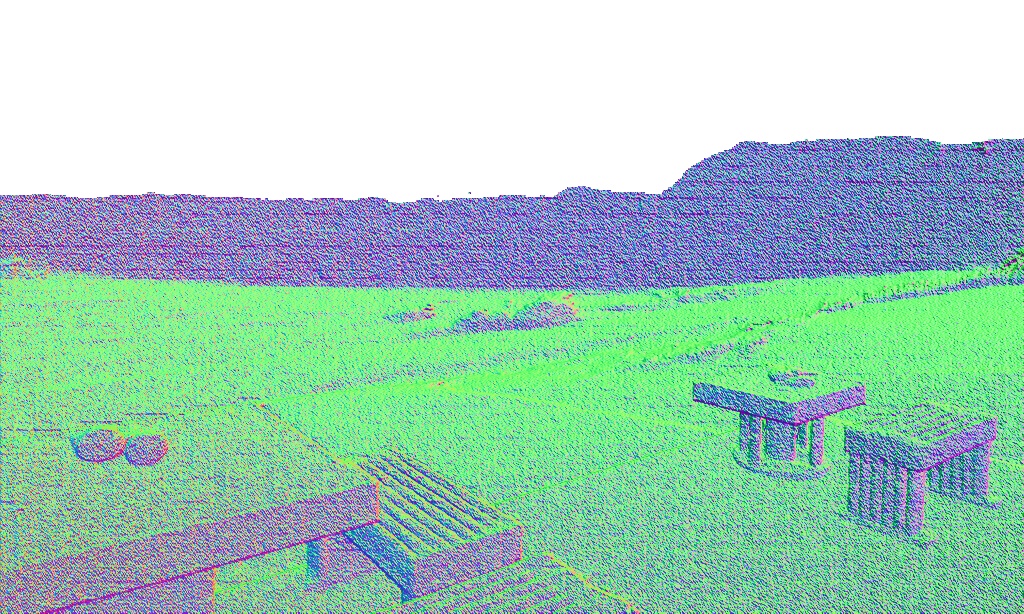} &
    \sotapanel{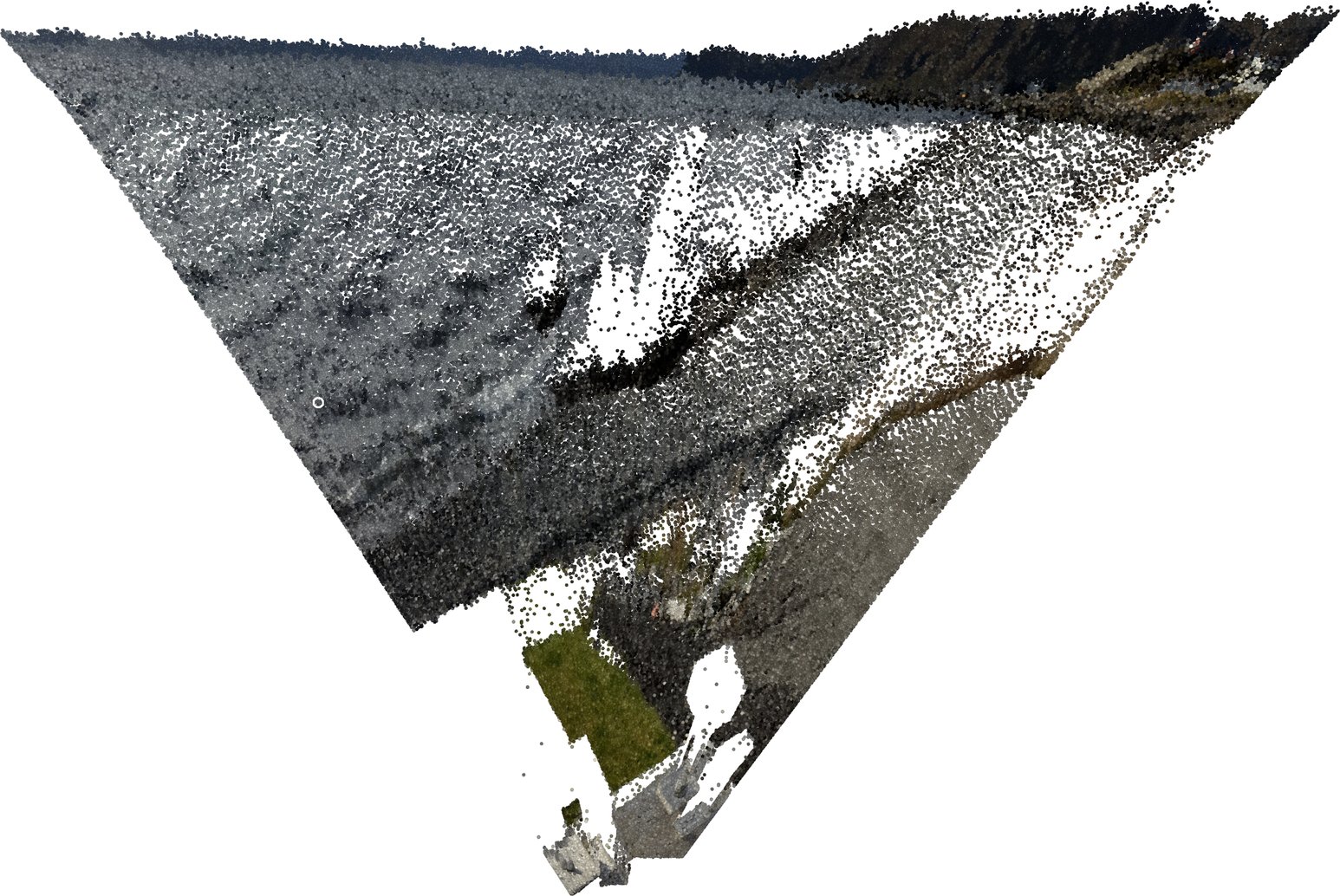} &
    \sotapanel{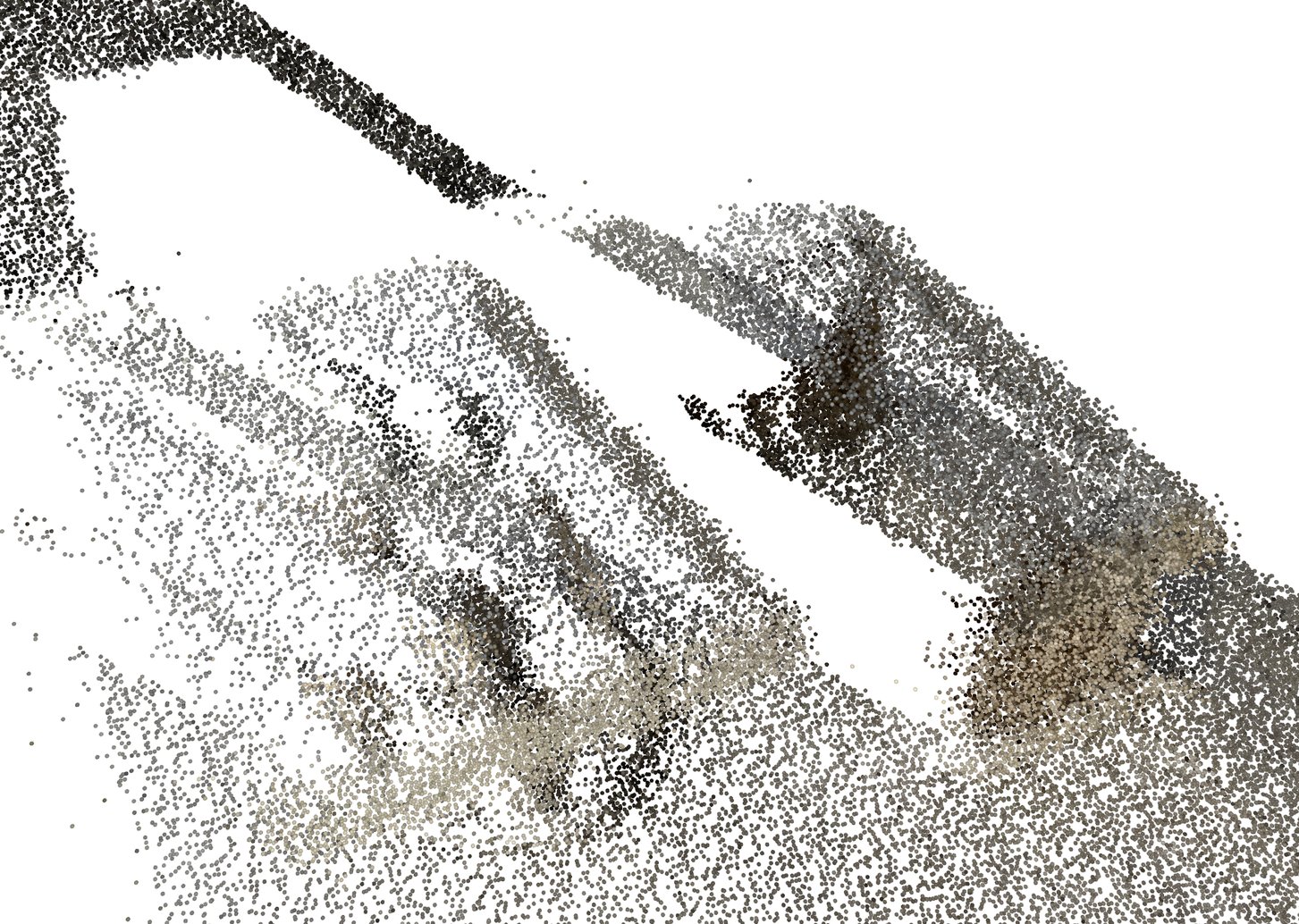} \\
    \addlinespace[2pt]

    \sotamethod{MoGe-2~\cite{wang2025moge2}} &
    \sotapanel{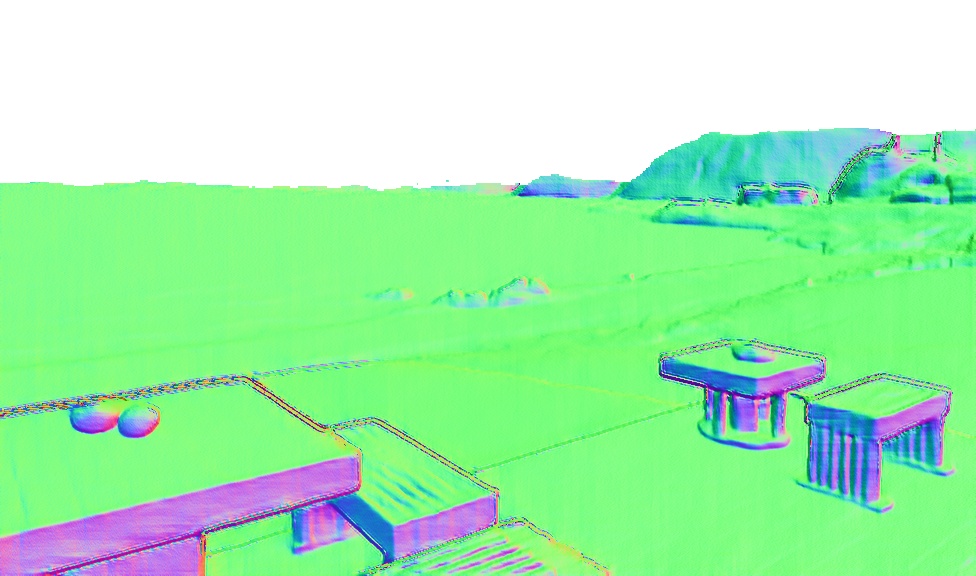} &
    \sotapanel{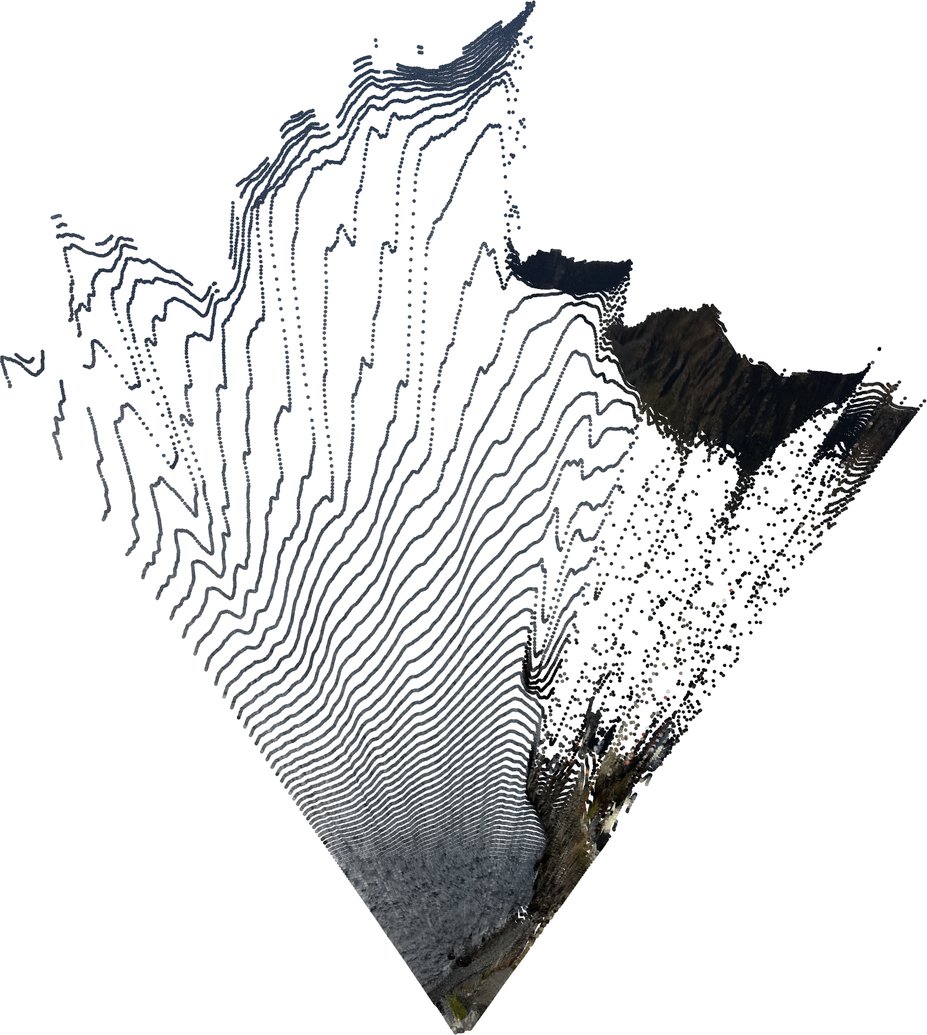} &
    \sotapanel{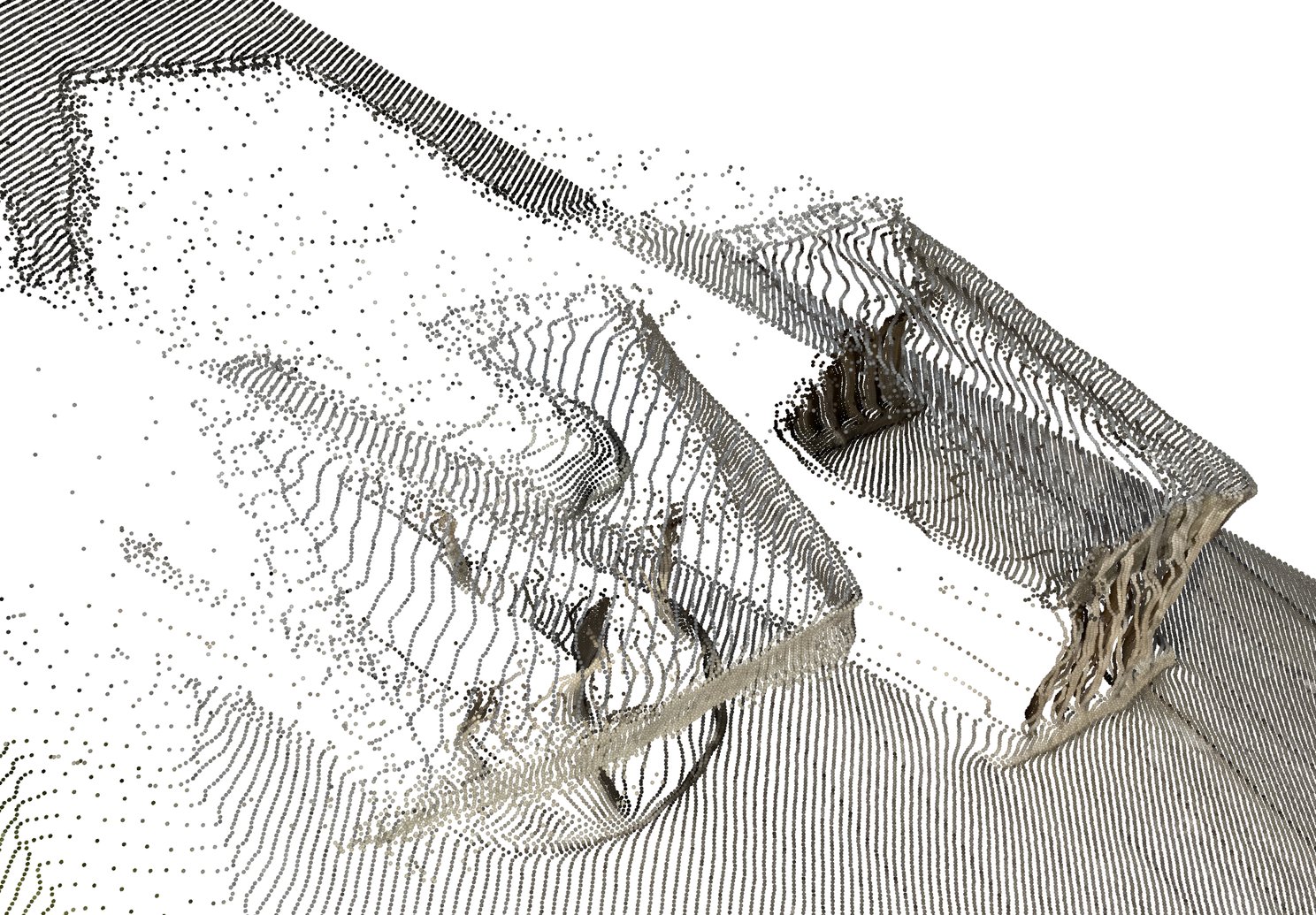} \\
    \addlinespace[2pt]

    \sotamethod{MoGe~\cite{wang2025moge}} &
    \sotapanel{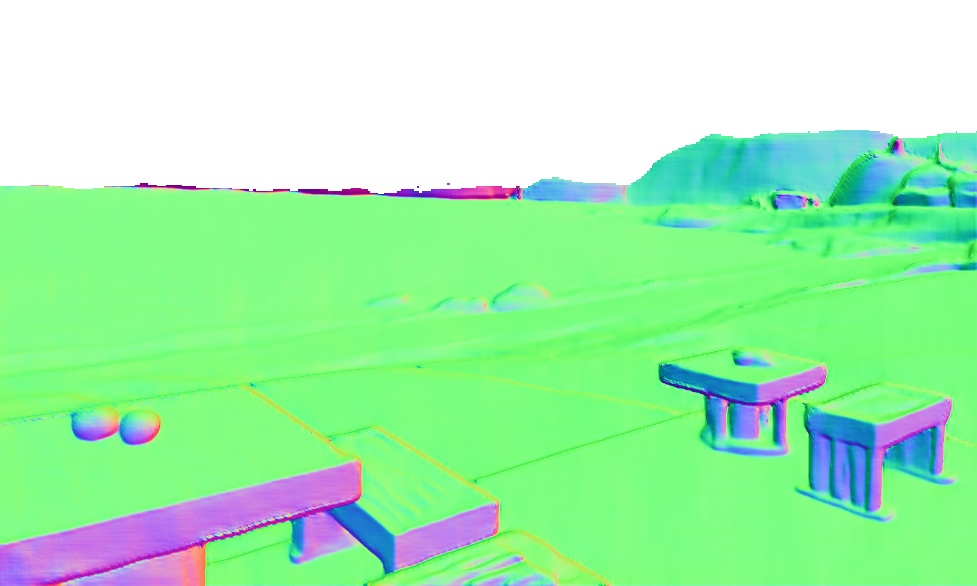} &
    \sotapanel{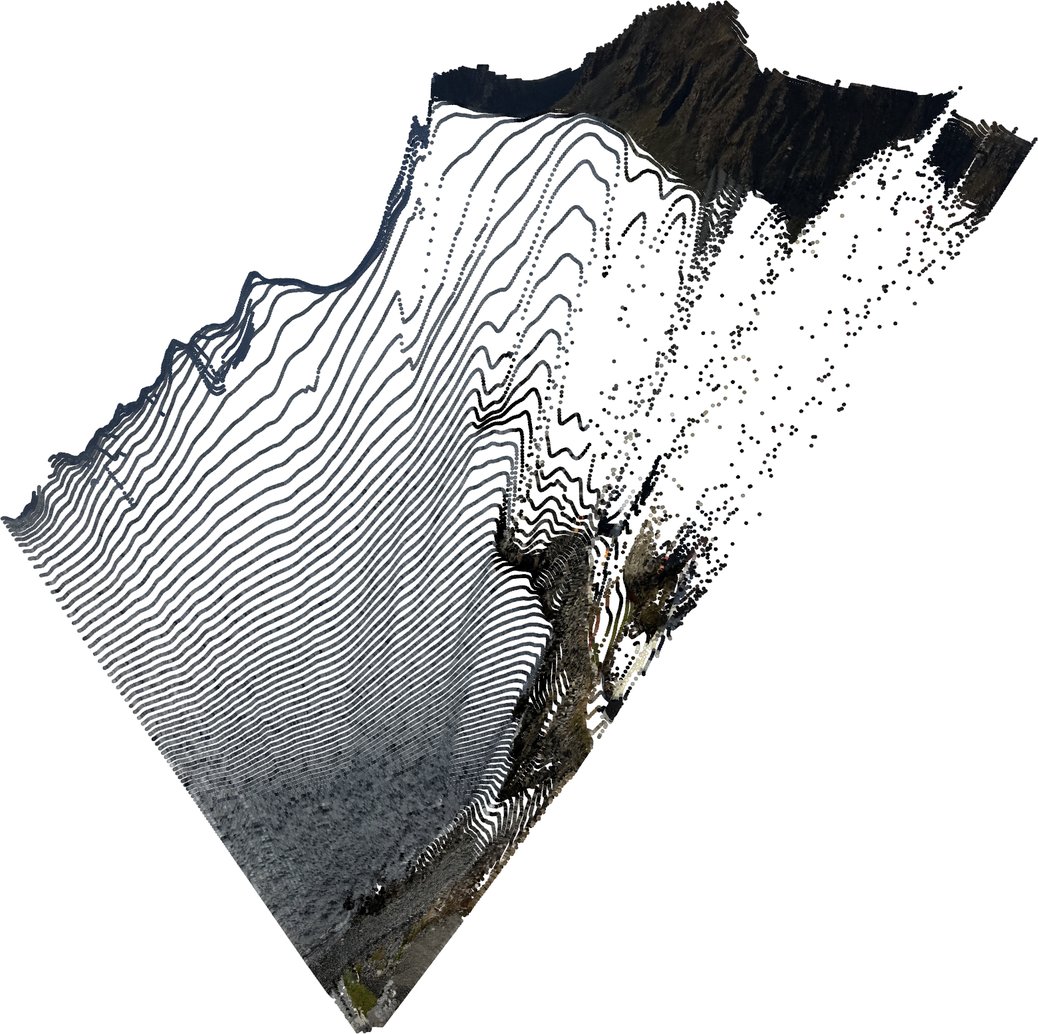} &
    \sotapanel{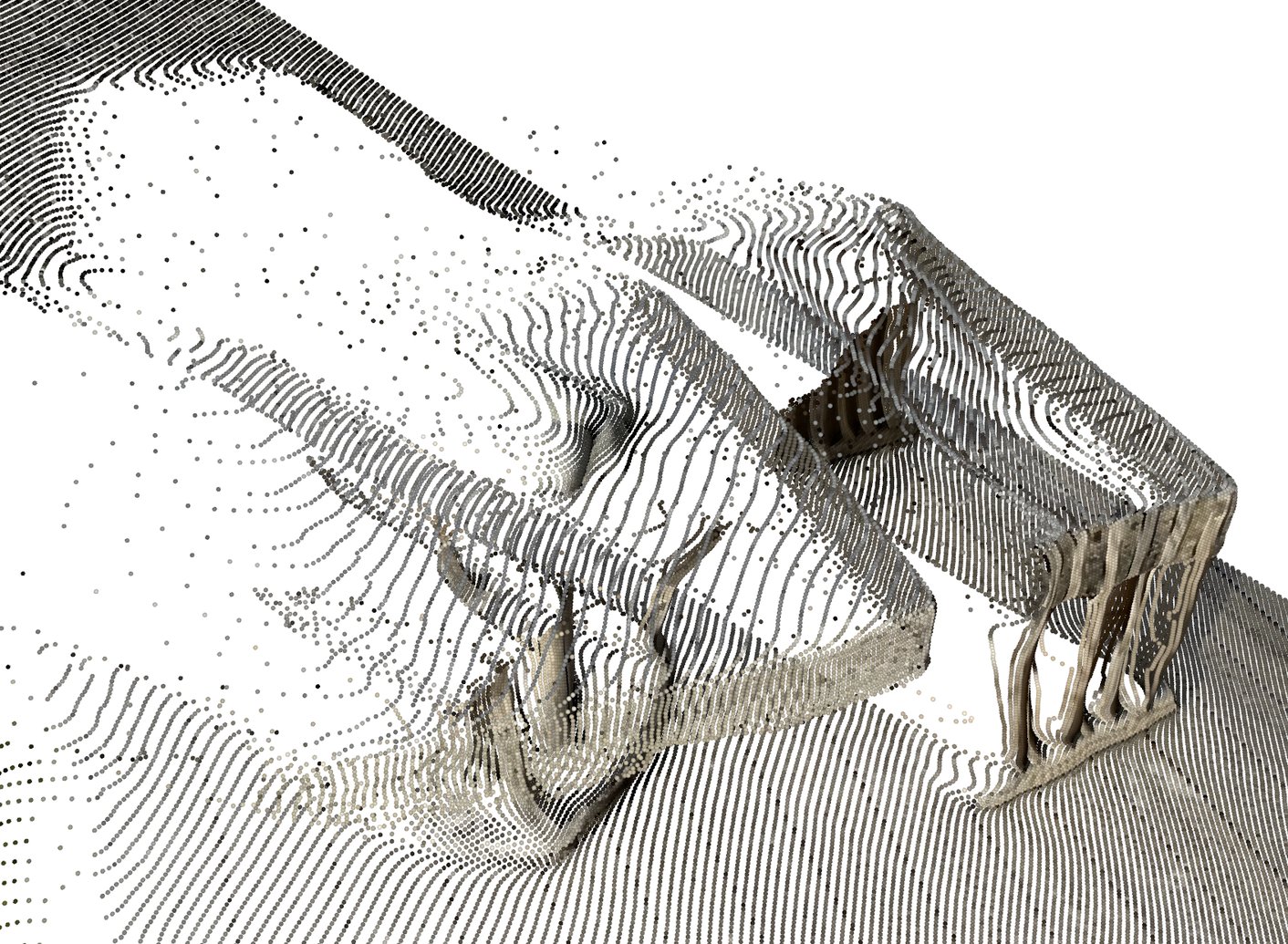} \\
    \addlinespace[2pt]

    \sotamethod{\textbf{\ours{}}} &
    \sotapanel{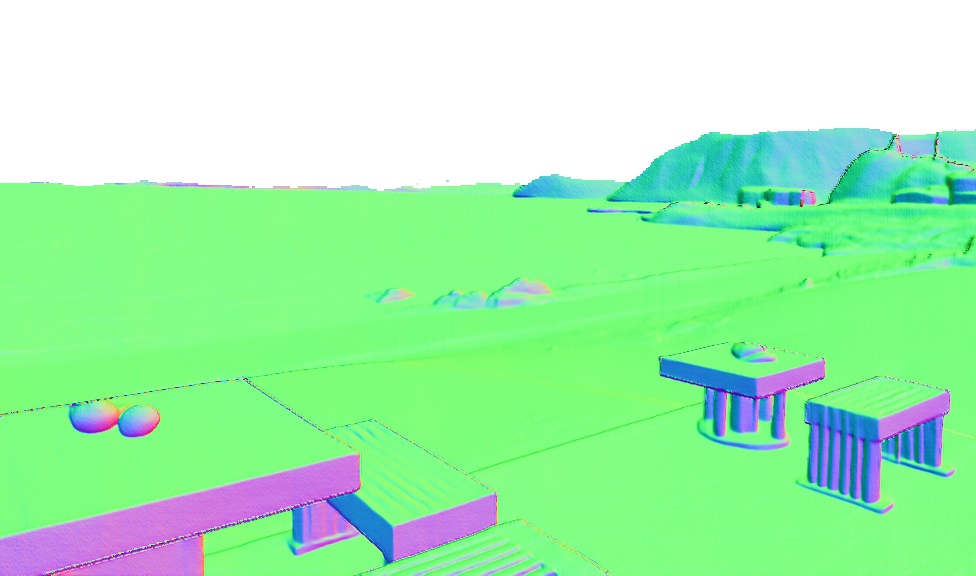} &
    \sotapanel{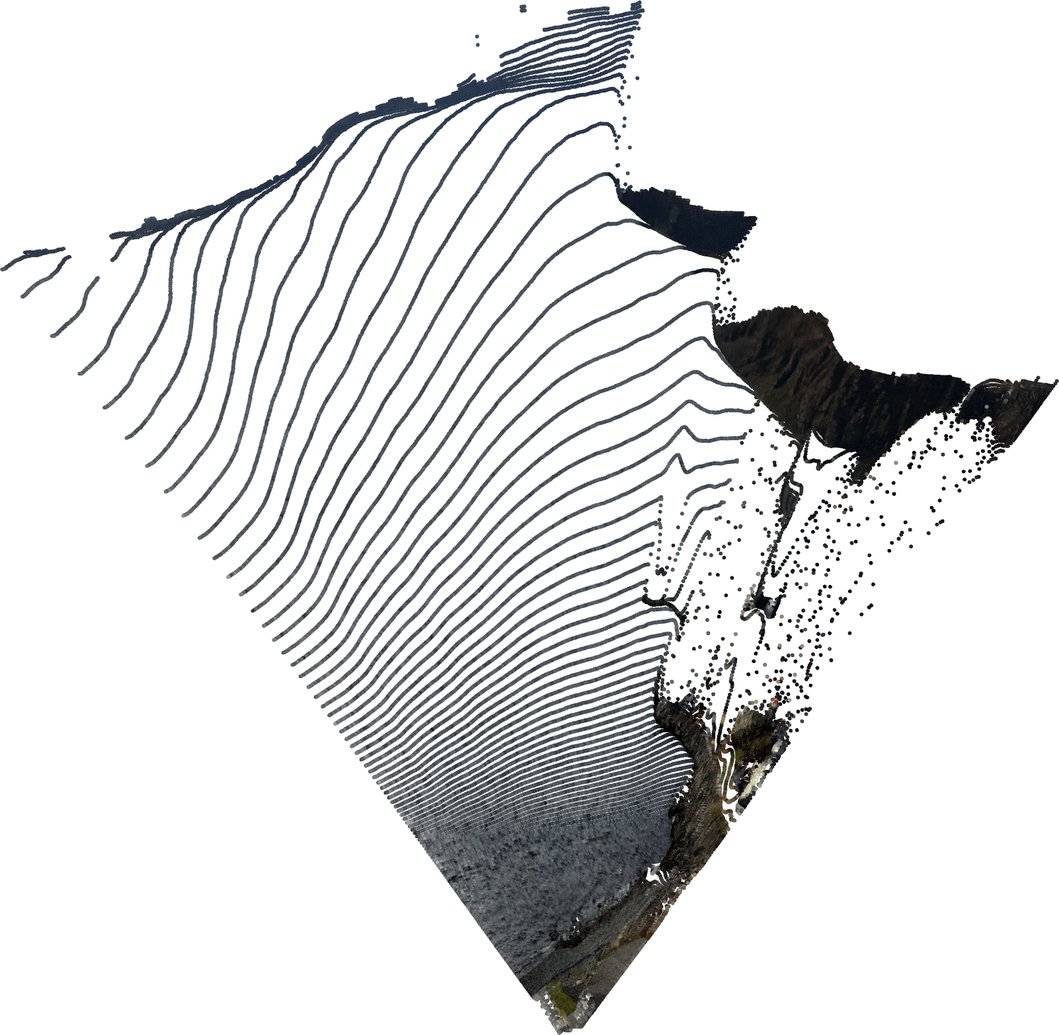} &
    \sotapanel{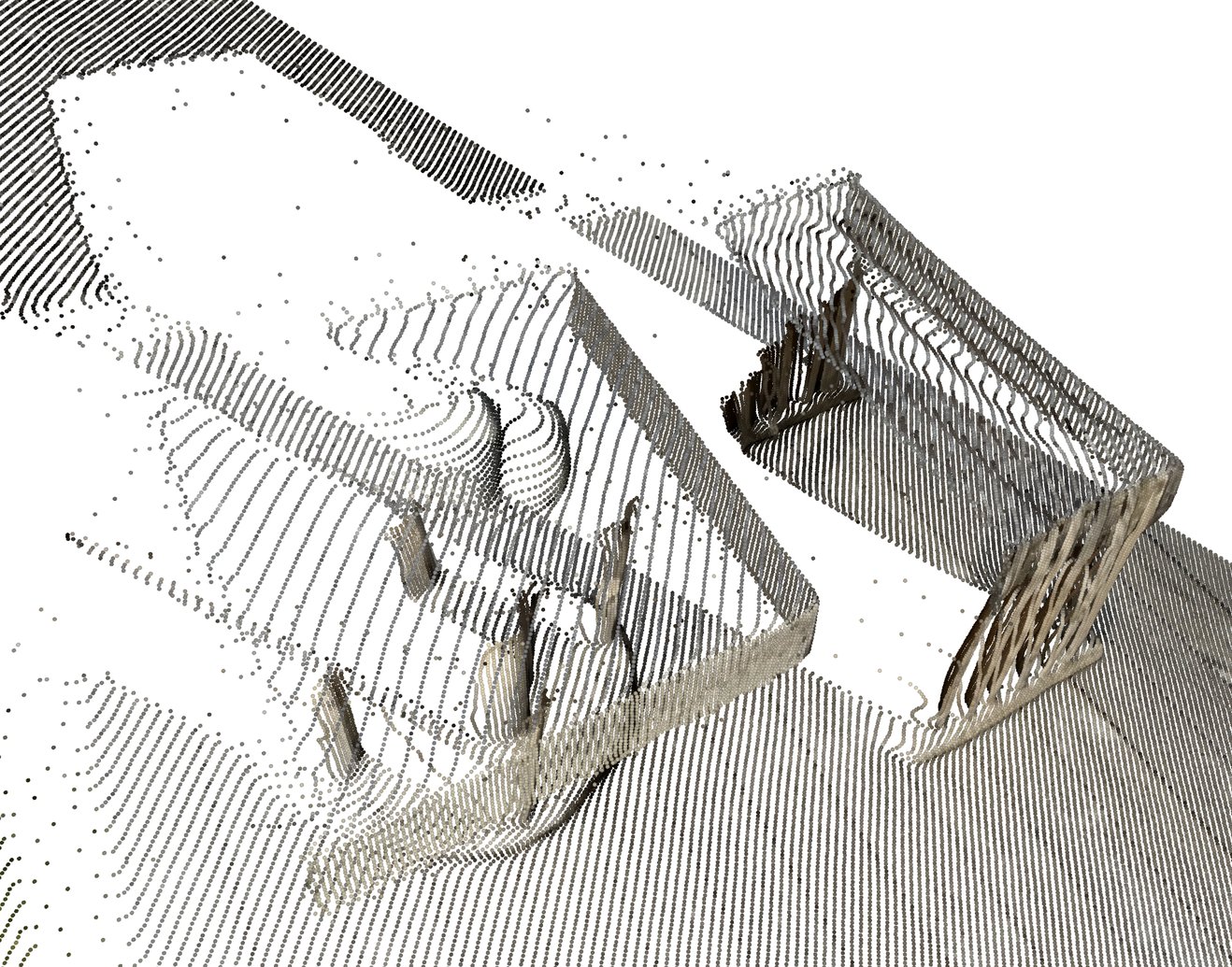} \\
  \end{tabular*}
  \end{adjustbox}
  \endgroup

  \caption{\textbf{Qualitative comparison of state-of-the-art methods in a high-dynamic-range scene.} The top-left panel shows the input image. Each row compares one method; columns show point map normals, followed by bird's-eye and near-camera renderings of the predicted point map. VGGT, InfiniDepth, and PPD struggle with the high dynamic range and collapse distant geometry. VGGT oversmooths surfaces, InfiniDepth, MoGe-2, and MoGe introduce oscillatory or bending artifacts on the table and bench legs, and PPD produces heavy surface noise. \ours{} exhibits the best local surface geometry with sharp edges and straight surfaces. Best viewed zoomed in.}
  \label{fig:sota_norway}
\end{figure}

\begin{figure}[p]
  \centering
  \begingroup
  \newcommand{\sotarowheight}{0.115\textheight}
  \newcommand{\sotamethodwidth}{0pt}

  \newcommand{\sotamethod}[1]{\parbox[c][\sotarowheight][c]{\sotamethodwidth}{\centering
      \rotatebox[origin=c]{90}{\footnotesize\strut #1}}}

  \newcommand{\sotapanel}[1]{\adjustbox{valign=c}{\includegraphics[height=\sotarowheight,width=\linewidth,keepaspectratio]{#1}}}

  \newcommand{\sotaheader}[1]{\makebox[\linewidth][c]{\footnotesize #1}}

  \setlength{\tabcolsep}{1pt}

  \begin{adjustbox}{max totalsize={\linewidth}{0.99\textheight},center}
  \begin{tabular*}{\linewidth}{@{}
                  >{\centering\arraybackslash}m{\sotamethodwidth}
                  @{\extracolsep{\fill}}
                  >{\centering\arraybackslash}m{0.29\linewidth}
                  >{\centering\arraybackslash}m{0.29\linewidth}
                  >{\centering\arraybackslash}m{0.29\linewidth}
                  @{}}
    &
    \adjustbox{valign=b}{\includegraphics[height=\sotarowheight,width=\linewidth,keepaspectratio]{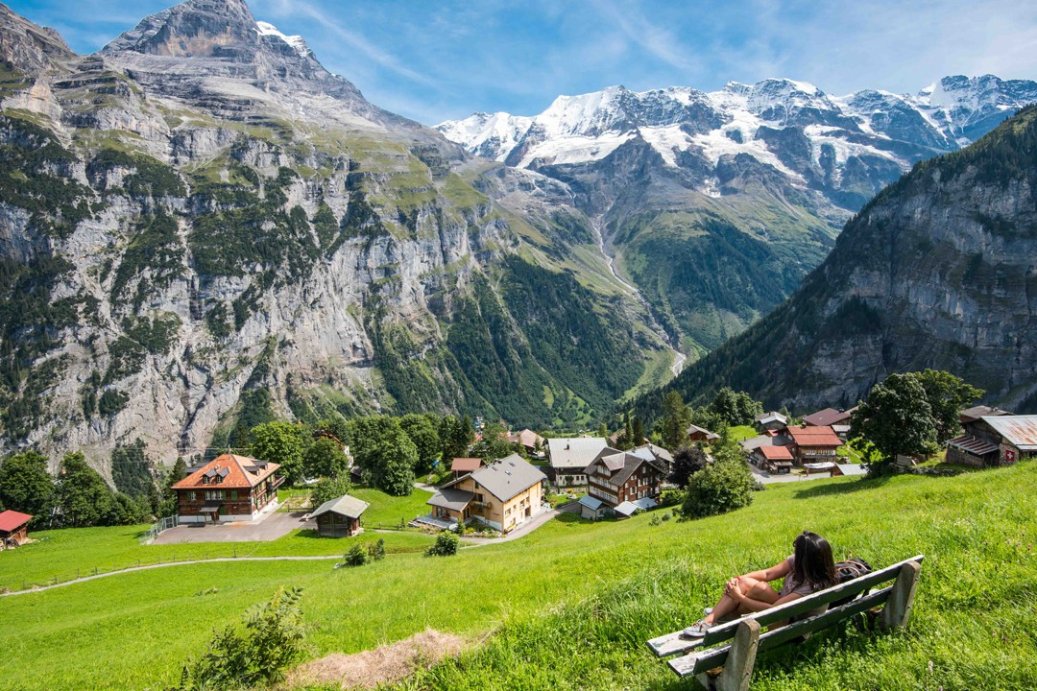}} &
    \sotaheader{Predicted Point Map} &
    \sotaheader{Near View} \\
    \addlinespace[2pt]

    \sotamethod{VGGT~\cite{wang2025vggt}} &
    \sotapanel{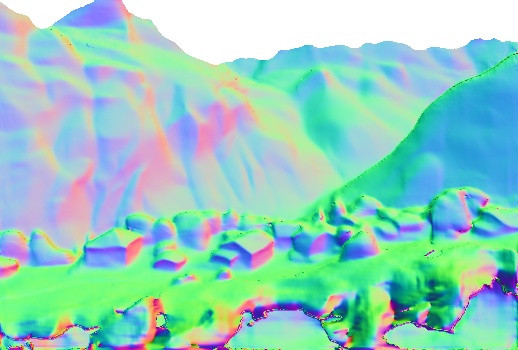} &
    \sotapanel{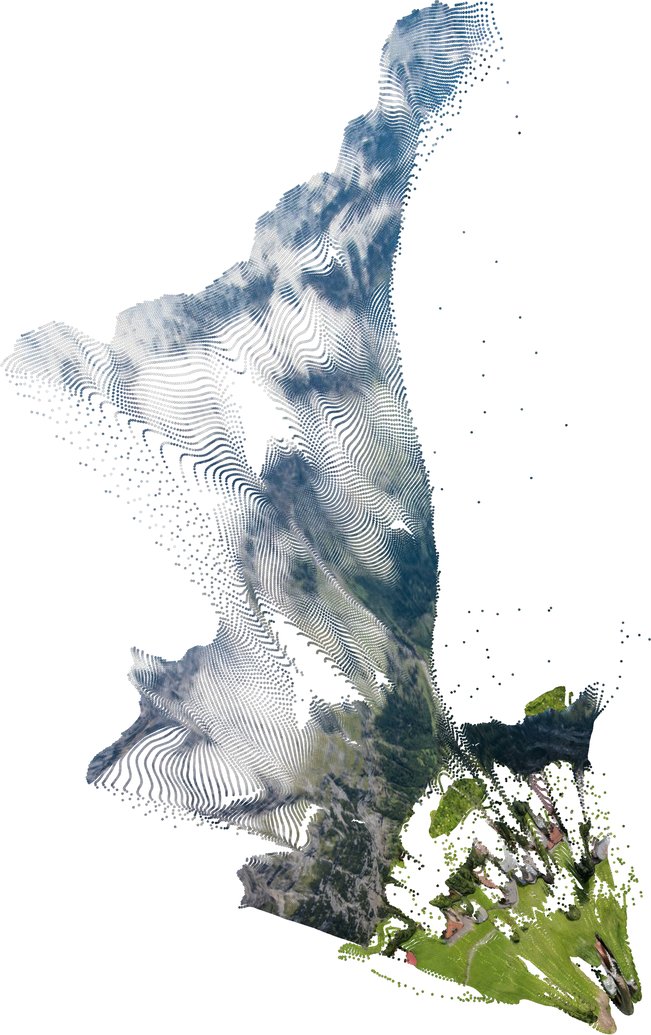} &
    \sotapanel{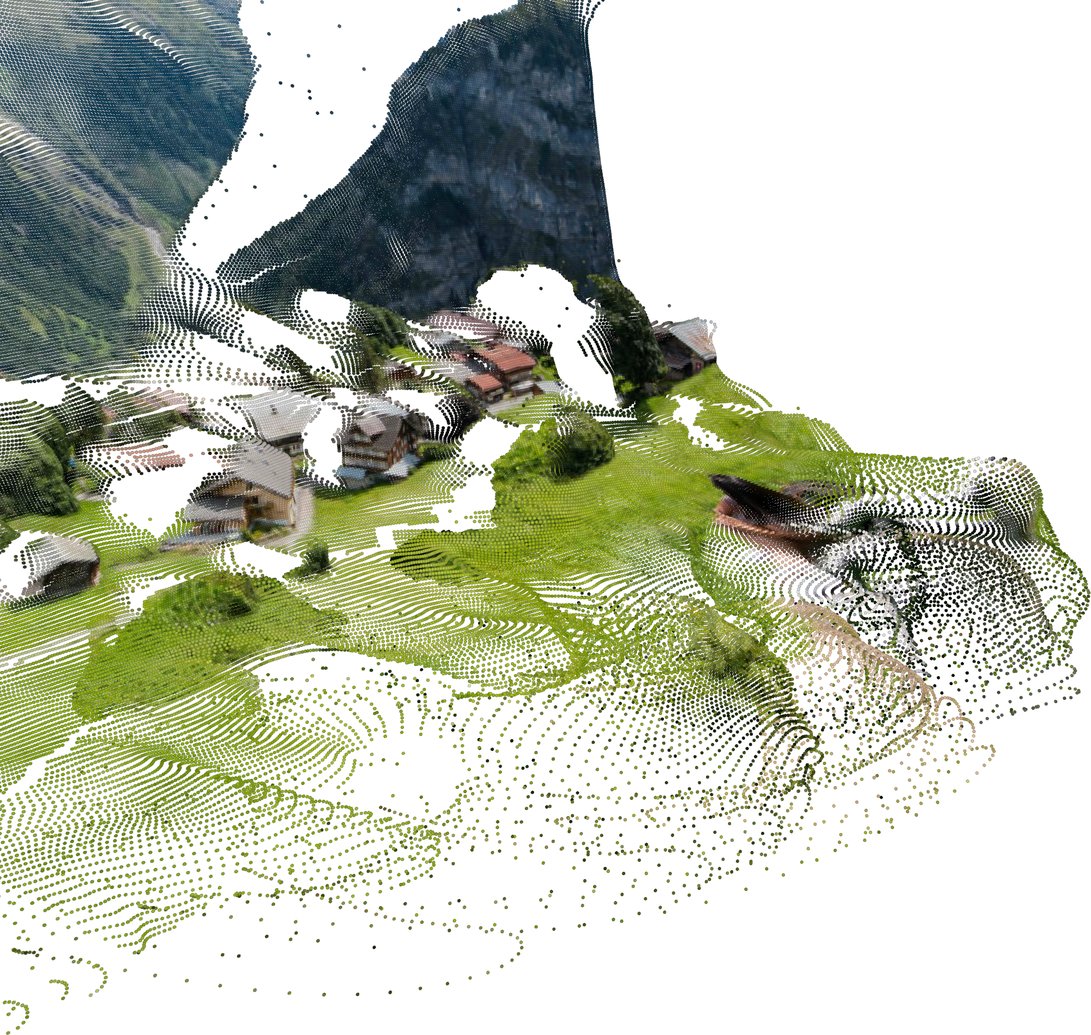} \\
    \addlinespace[2pt]

    \sotamethod{InfiniDepth~\cite{yu2026infinidepth}} &
    \sotapanel{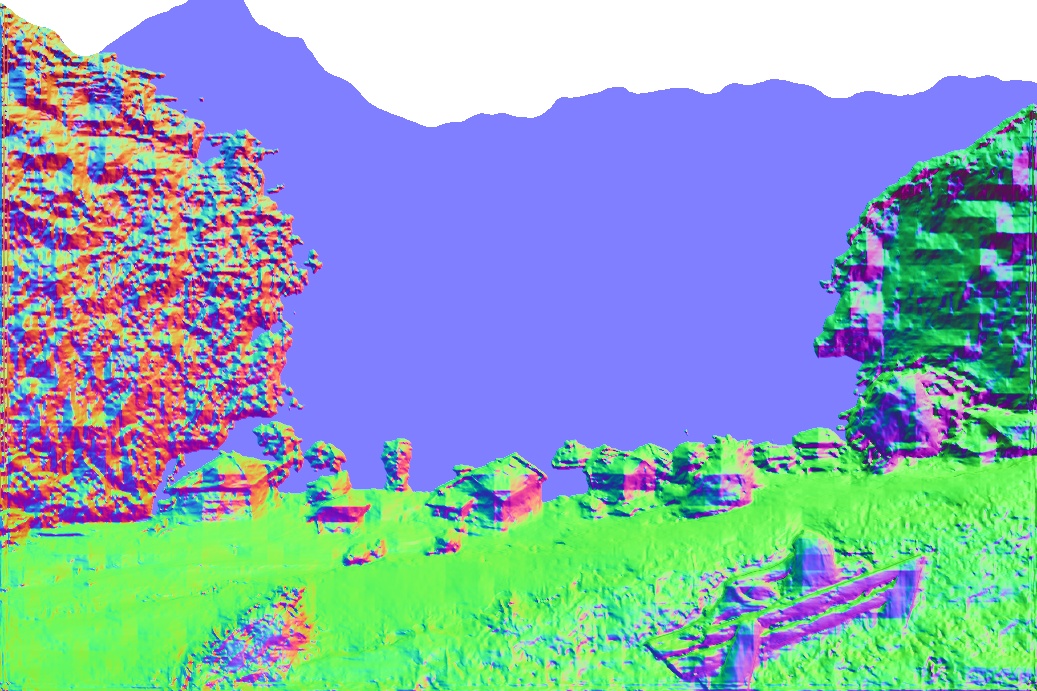} &
    \sotapanel{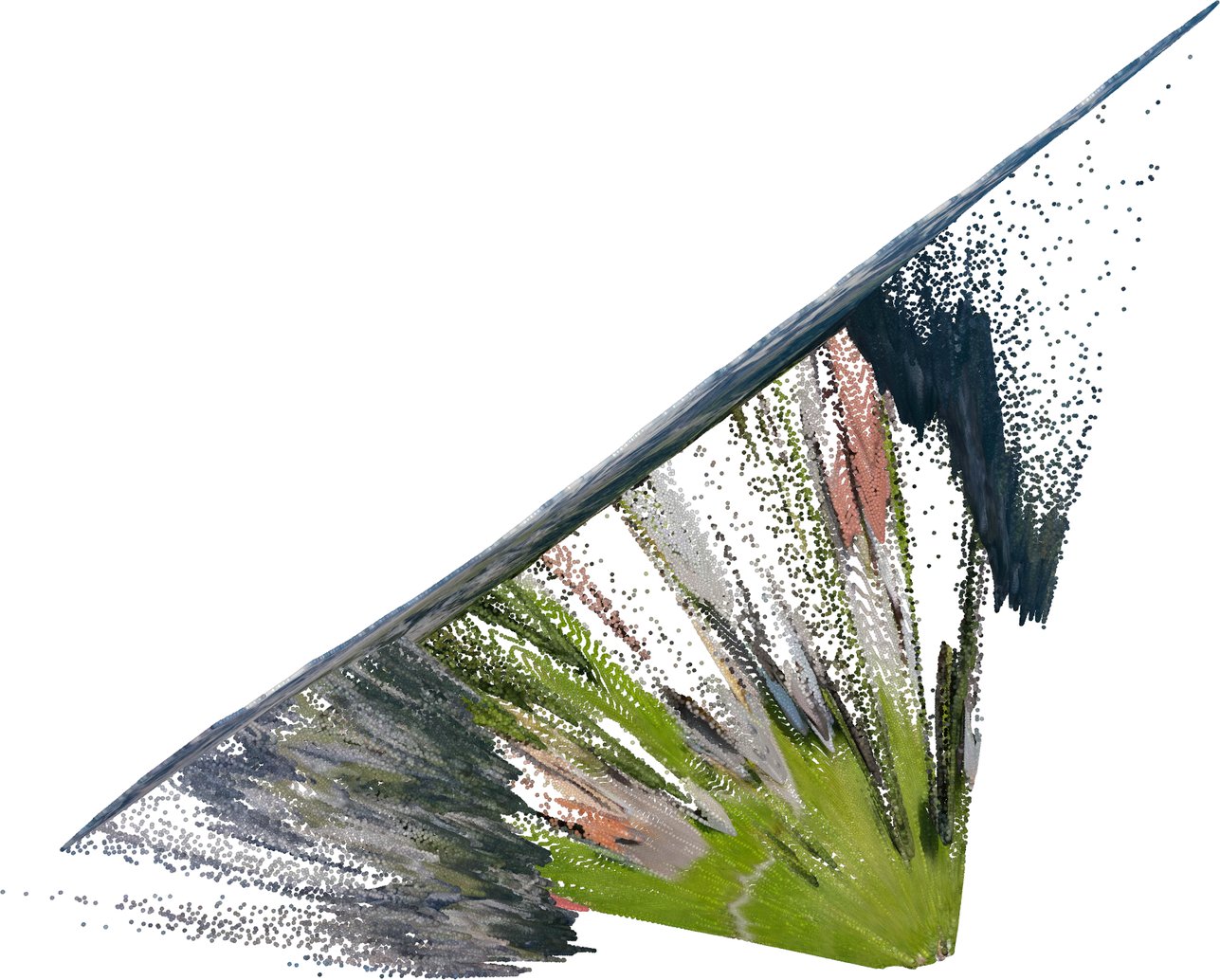} &
    \sotapanel{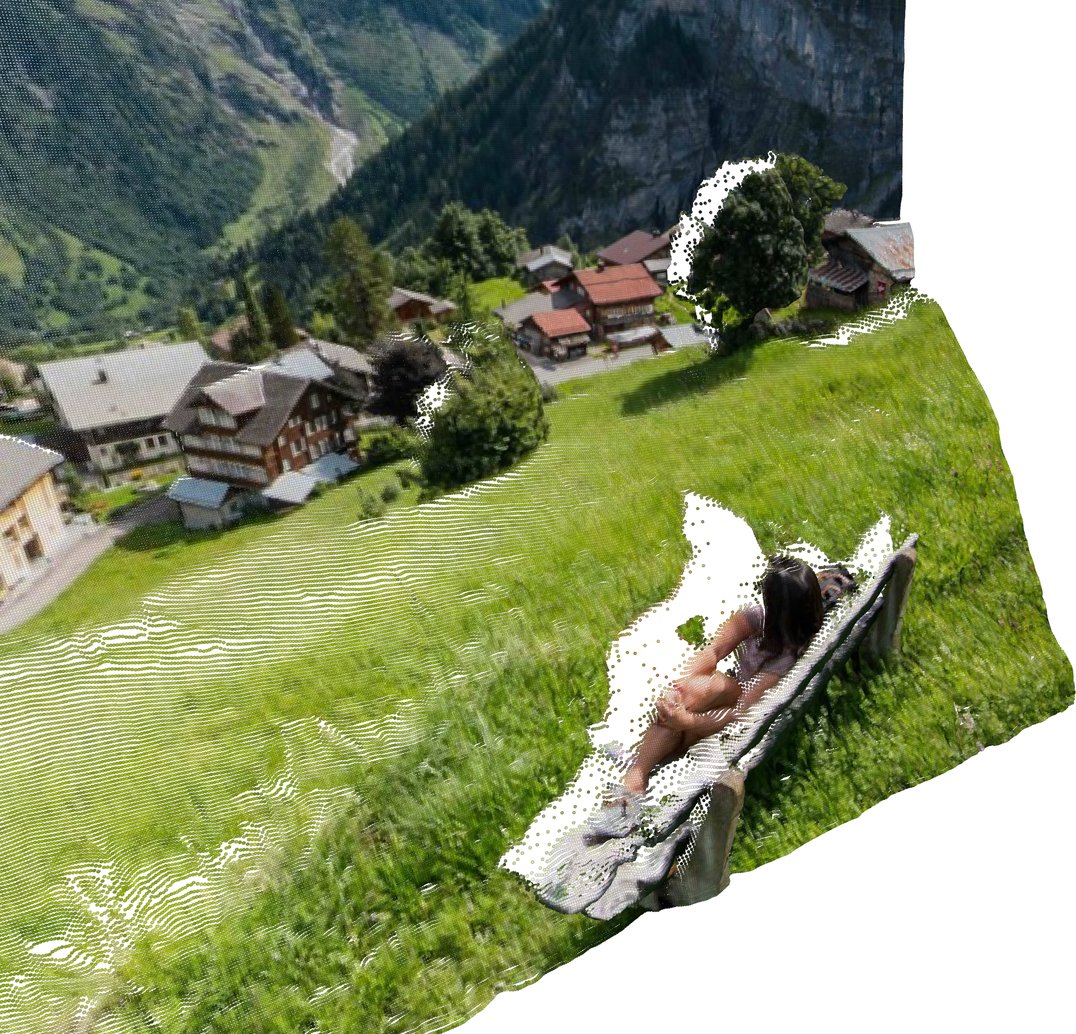} \\
    \addlinespace[2pt]

    \sotamethod{PPD~\cite{xu2025ppd}} &
    \sotapanel{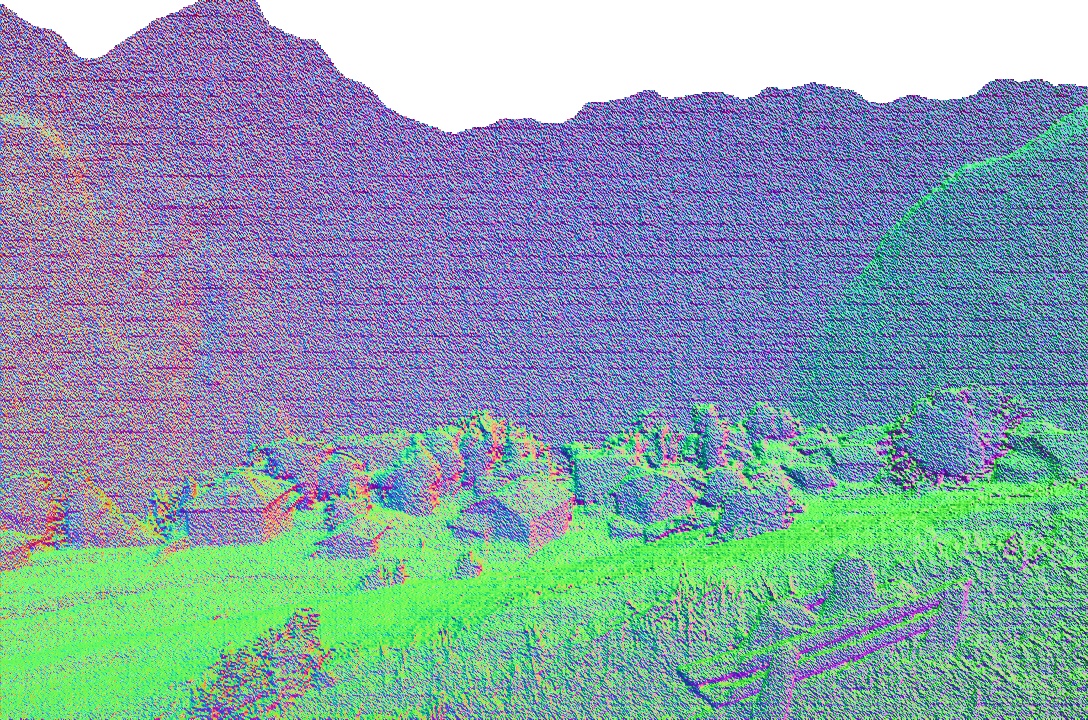} &
    \sotapanel{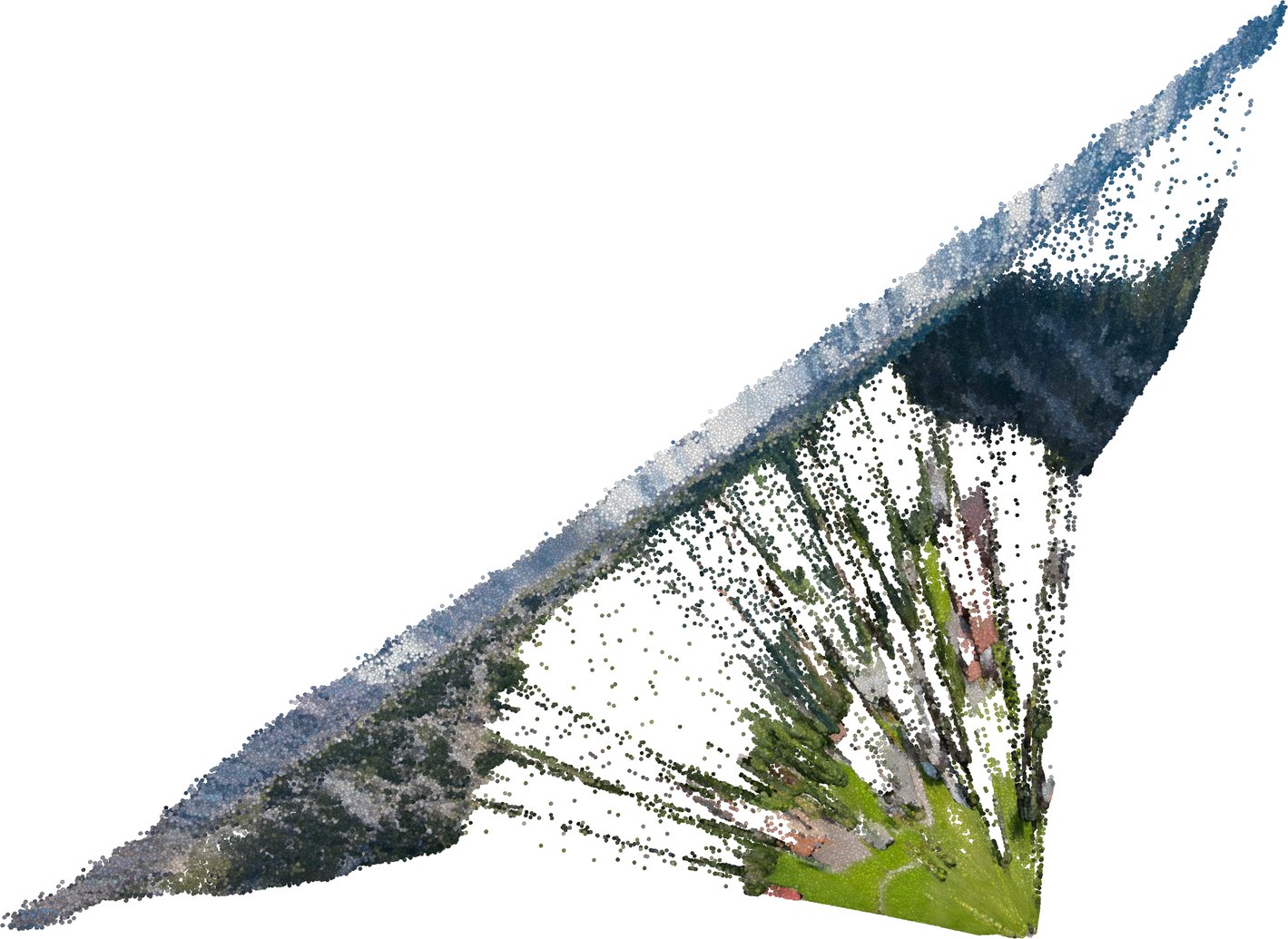} &
    \sotapanel{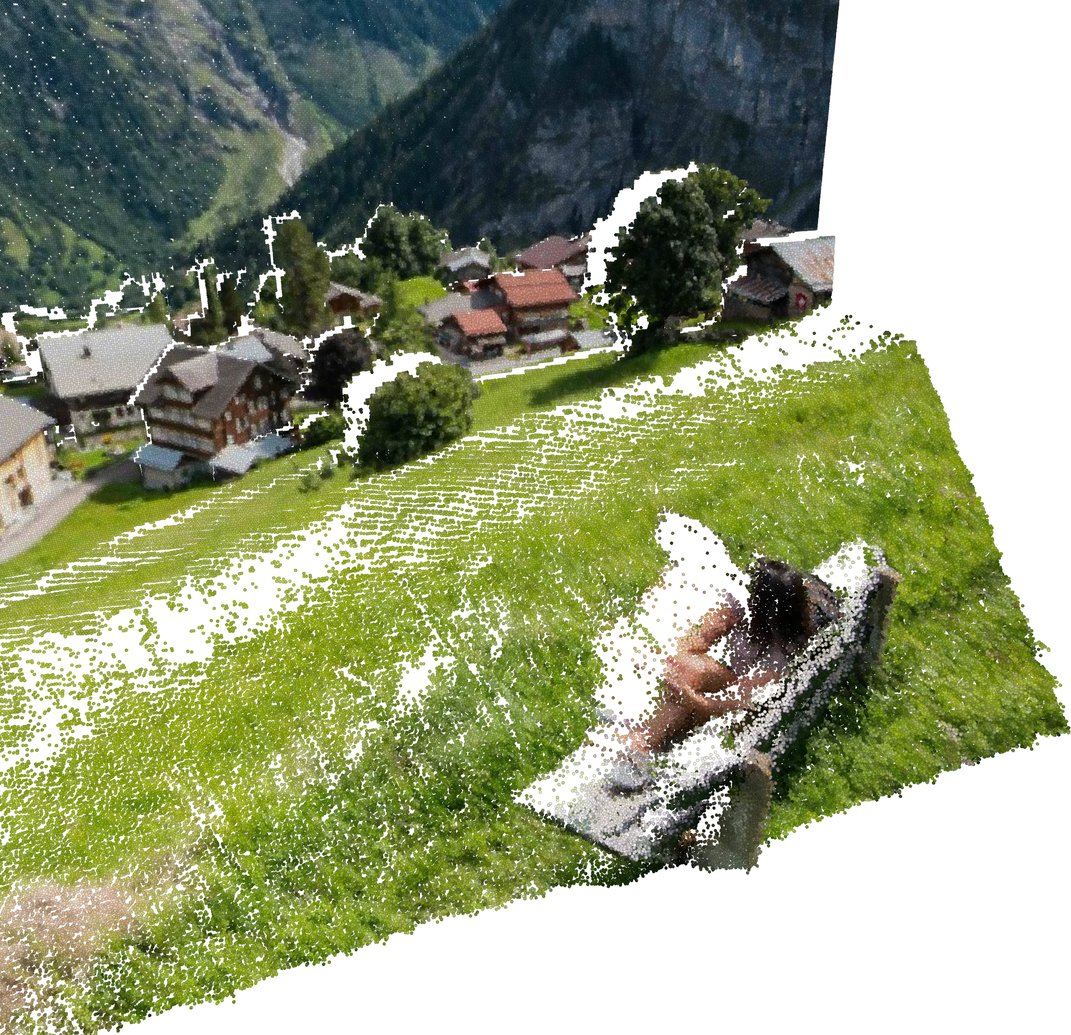} \\
    \addlinespace[2pt]

    \sotamethod{MoGe-2~\cite{wang2025moge2}} &
    \sotapanel{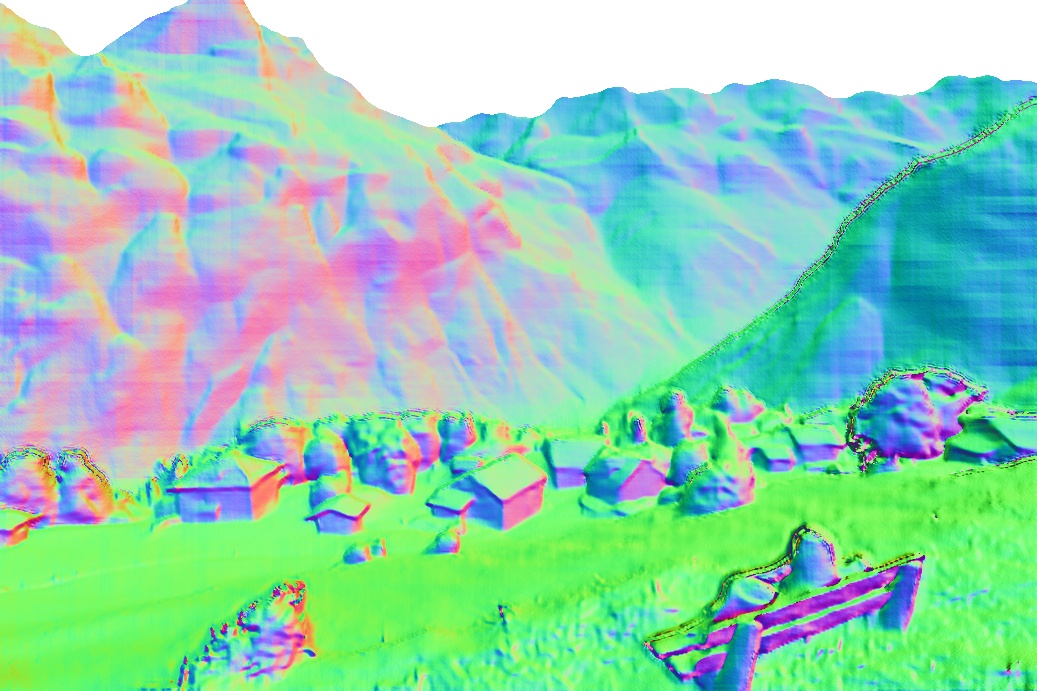} &
    \sotapanel{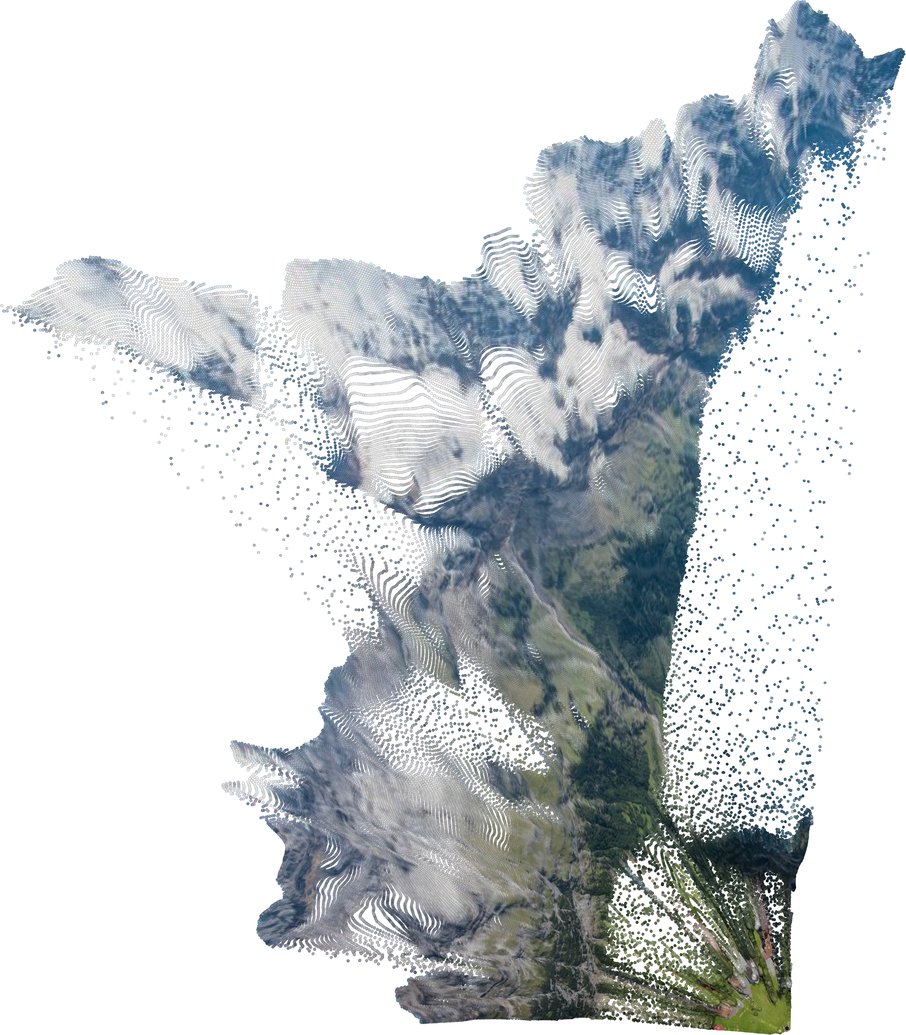} &
    \sotapanel{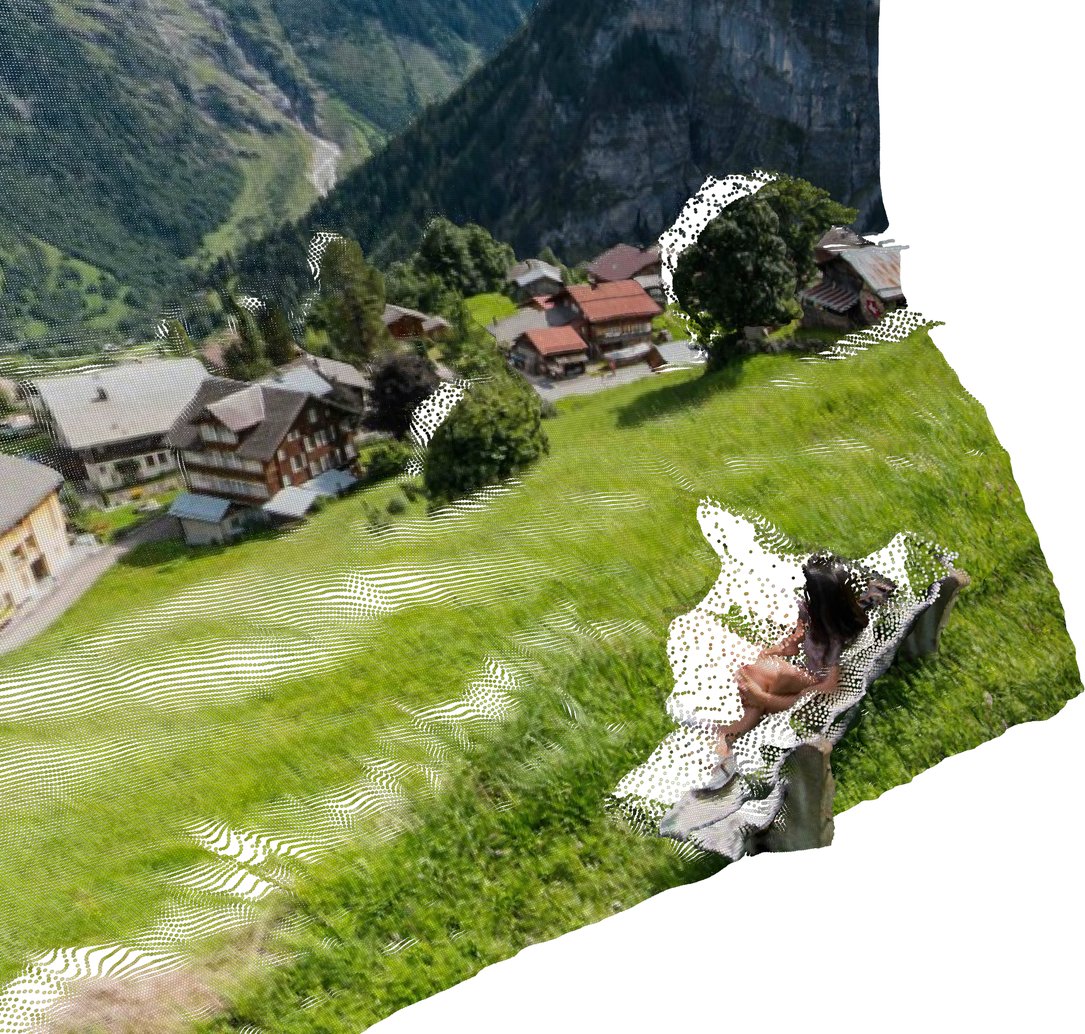} \\
    \addlinespace[2pt]

    \sotamethod{MoGe~\cite{wang2025moge}} &
    \sotapanel{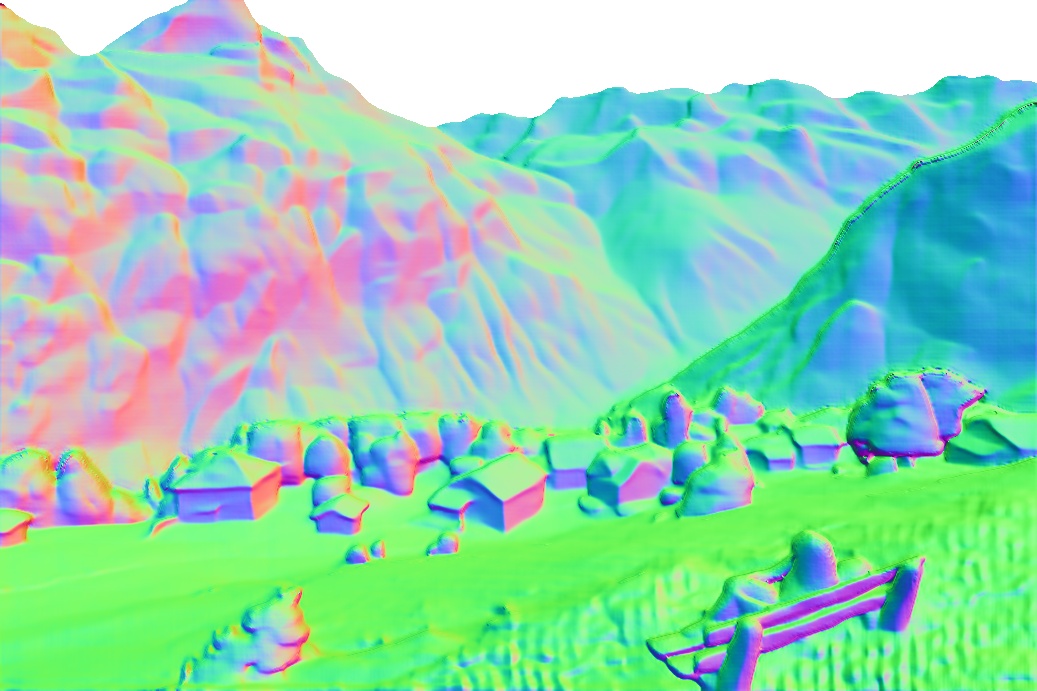} &
    \sotapanel{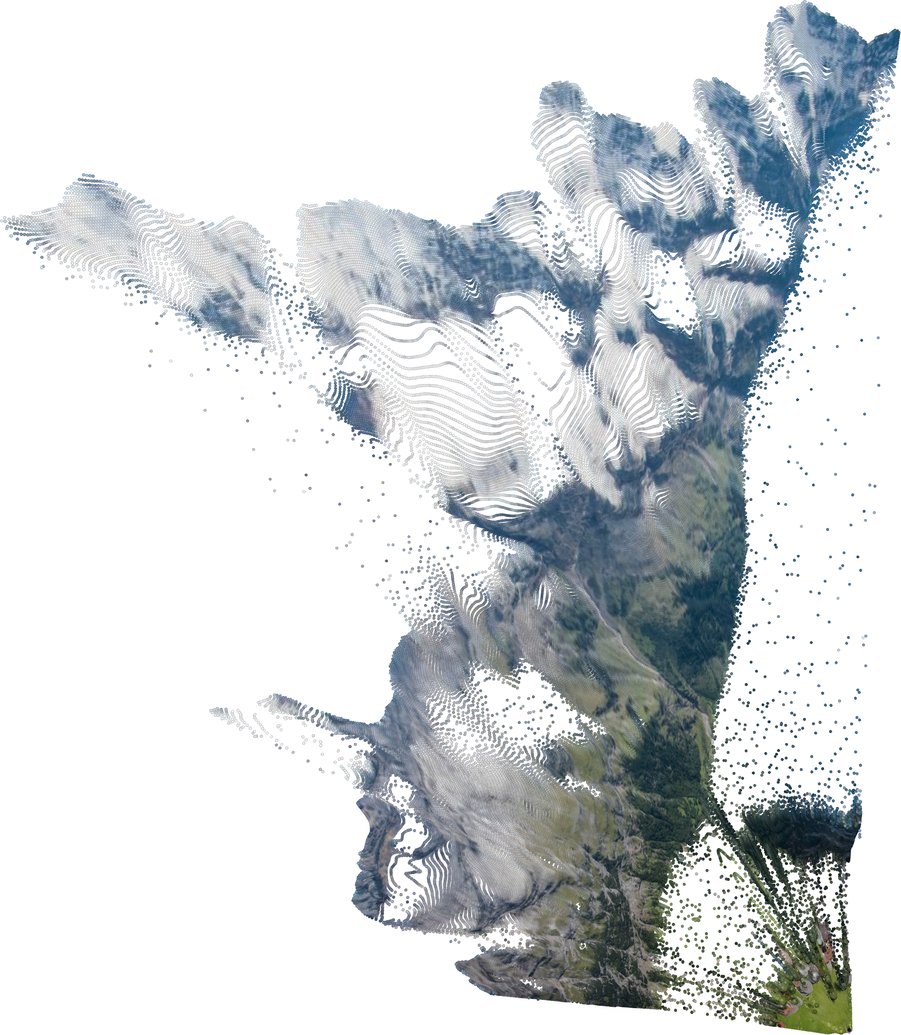} &
    \sotapanel{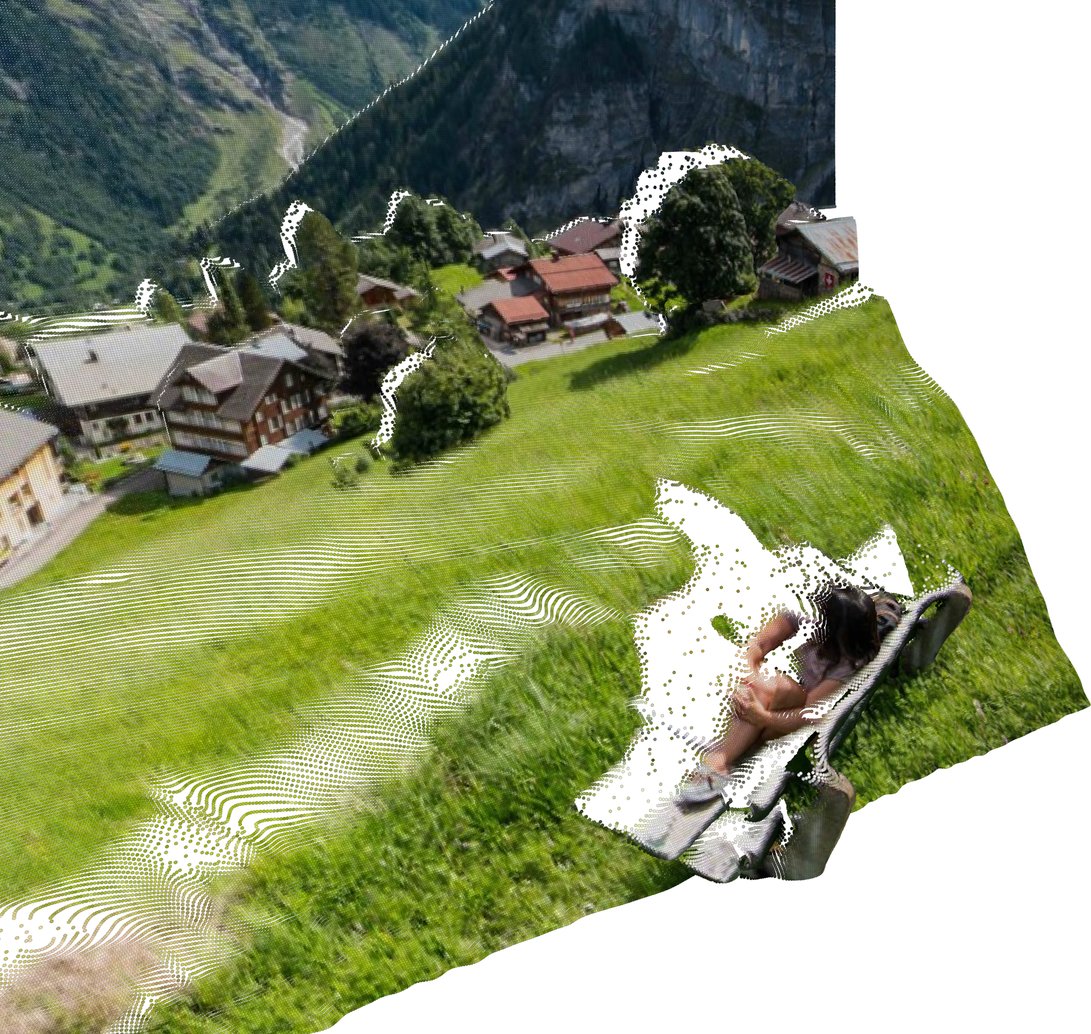} \\
    \addlinespace[2pt]

    \sotamethod{\textbf{\ours{}}} &
    \sotapanel{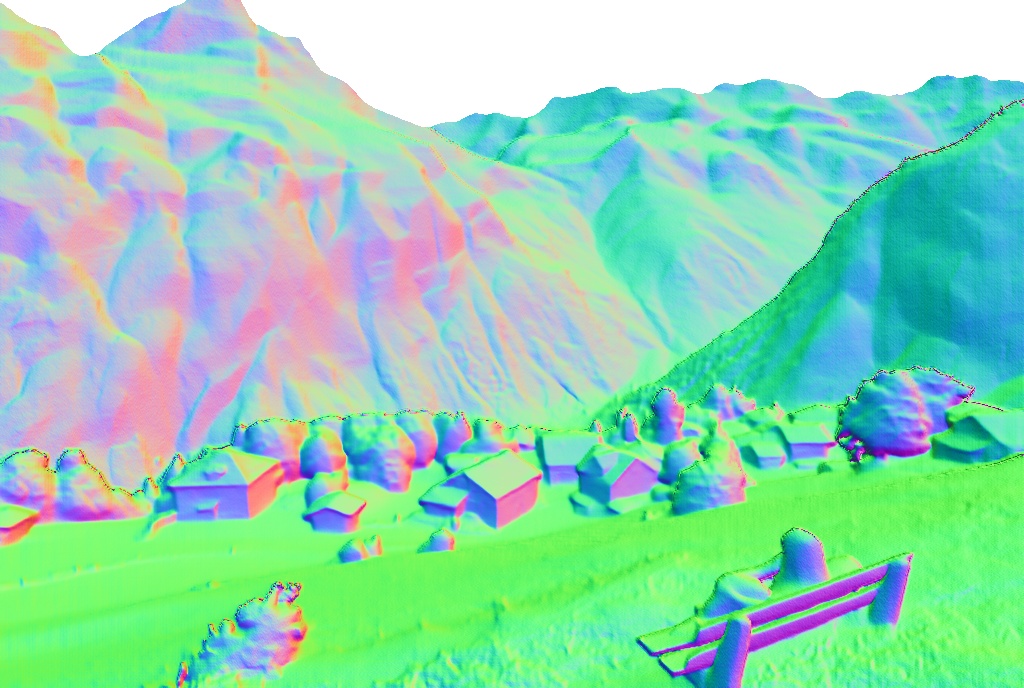} &
    \sotapanel{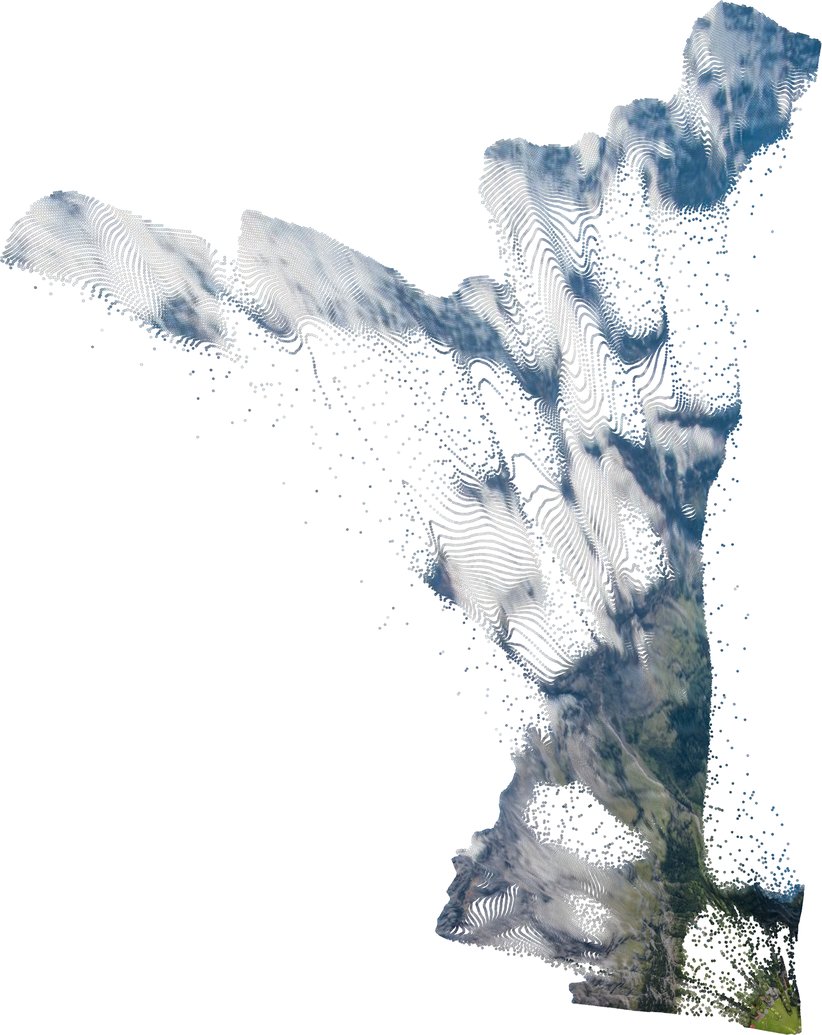} &
    \sotapanel{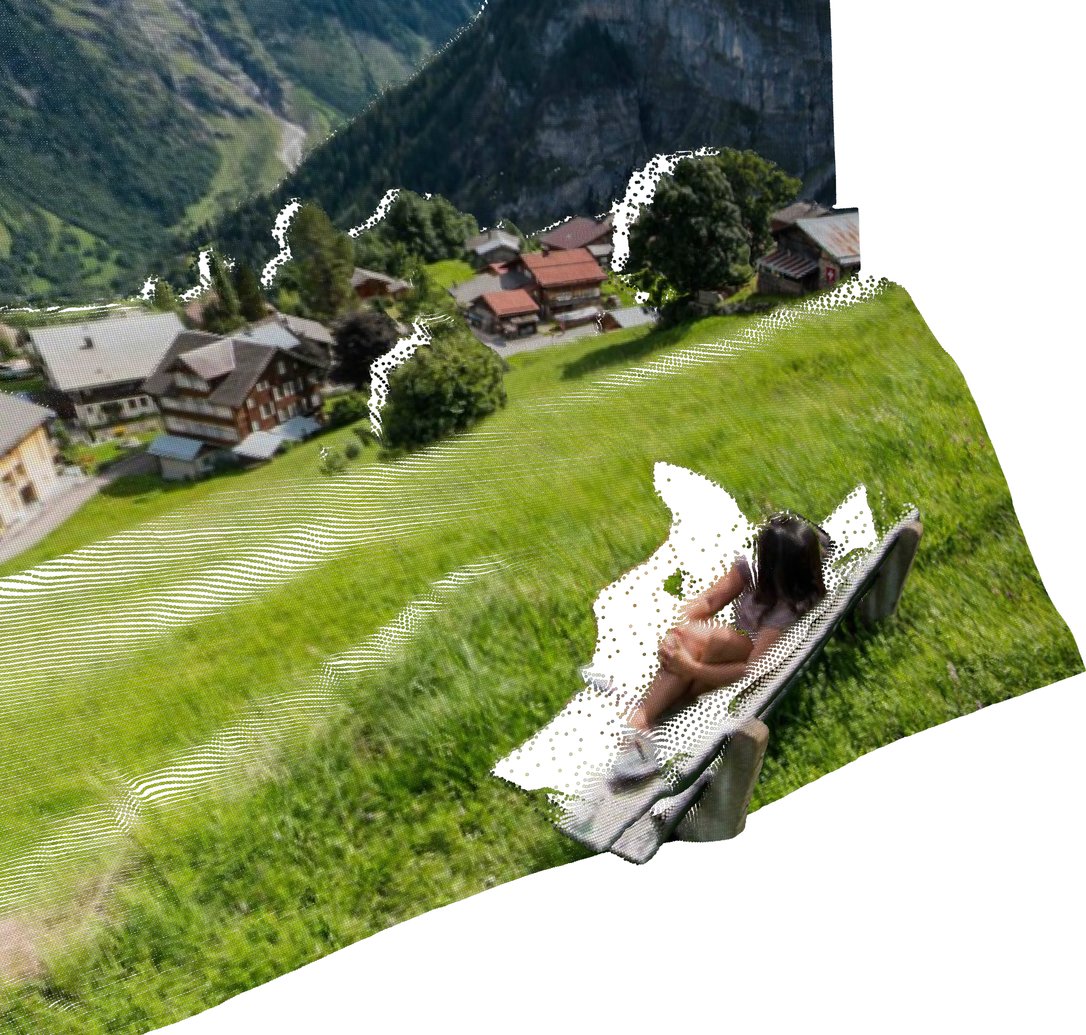} \\
  \end{tabular*}
  \end{adjustbox}
  \endgroup

  \caption{\textbf{Qualitative comparison of state-of-the-art methods in a high-dynamic-range scene.} The top-left panel shows the input image. Each row compares one method; columns show point map normals, followed by bird's-eye and near-camera renderings of the predicted point map. InfiniDepth and PPD struggle with the high dynamic range, VGGT fails to reconstruct geometry close to the camera, and \ours{} produces cleaner local surface geometry than MoGe. Best viewed zoomed in.}
  \label{fig:sota_swiss}
\end{figure}

\Cref{fig:sota_swiss,fig:sota_norway} compare \ours{} with recent methods on in-the-wild scenes with large depth ranges.
PPD~\cite{xu2025ppd} and InfiniDepth~\cite{yu2026infinidepth} struggle most in this setting: both recover plausible near-field geometry but strongly compress distant scene structure.
Their normal maps further show strong surface noise for PPD and visible surface artifacts for InfiniDepth.
VGGT~\cite{wang2025vggt} is less affected but still produces broken near-field geometry or slightly compressed distant structure.
MoGe~\cite{wang2025moge} and MoGe-2~\cite{wang2025moge2} are closest to \ours{} in global layout, but their rendered point maps and normals show less coherent local surface geometry, including bending and oscillatory artifacts.

\Cref{supp:qualitative_voltera_cups,supp:qualitative_stairs_street,supp:qualitative_sandwich_oranges} show additional in-the-wild predictions from \ours{}.

\section{Failure Cases}
\label{supp:failure_cases}

The qualitative comparisons show substantial reductions in bending, oscillation, and surface noise, consistent with the improvements in the local metrics.
However, these artifacts are not entirely eliminated: in complex scenes with many simultaneous high-frequency geometry changes, we still observe occasional local artifacts.
Such scenes place evidence from several surfaces in the same local neighborhoods.
The decoder ablation in~\cref{tab:decoder:local-points-rel,tab:decoder:points-normal-angle-mae,fig:decoder_ablation} suggests that NAD's content-dependent feature mixing helps separate this evidence, but fully disentangling many nearby structures remains challenging.

Floaters around occlusion boundaries are another recurring failure mode, although \ours{} produces substantially fewer than the compared baselines, as visible in~\cref{fig:teaser}.
All qualitative results shown in this work use raw predictions without floater removal or other geometric cleanup.
SMD-Nets~\cite{tosi2021smdnets} and MDA~\cite{bian2026mda} address this ambiguity by estimating multimodal disparity and depth distributions, respectively.
These output formulations are orthogonal to NAD and could be combined with it.

A further limitation comes from the training data.
Real-world datasets are necessary for strong zero-shot accuracy, but their SfM reconstructions often omit moving objects such as pedestrians or absorb them into the background.
Training on these labels can suppress fine structures and cause foreground objects to bleed into the predicted background geometry.
Refining real-world annotations is therefore a complementary direction for improving local geometry.

\clearpage

\tikzset{
  thumb/.style={
    inner sep=0,
    outer sep=0.5em,
  }
}

\begin{figure}
\centering
\begin{tikzpicture}
\node[thumb] (pcl) {\includegraphics[width=12cm]{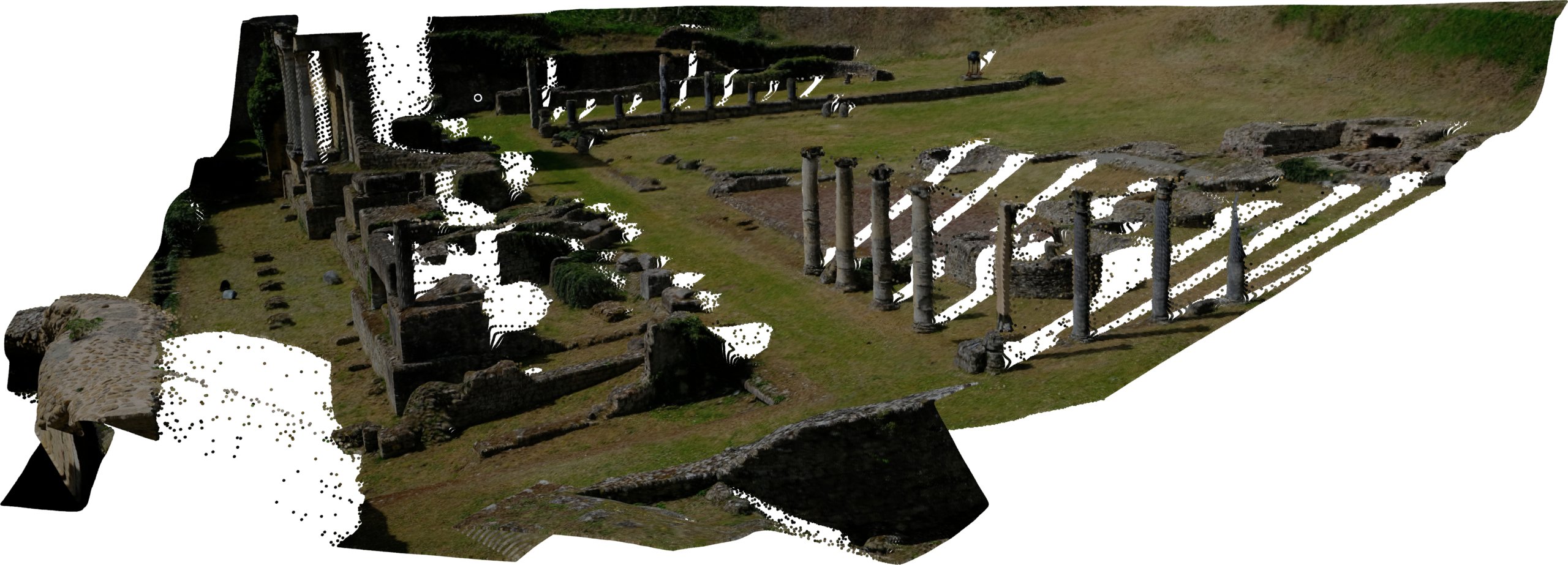}};
\node[thumb, anchor=south] (img-m) at (pcl.north) {\includegraphics[width=4.4cm]{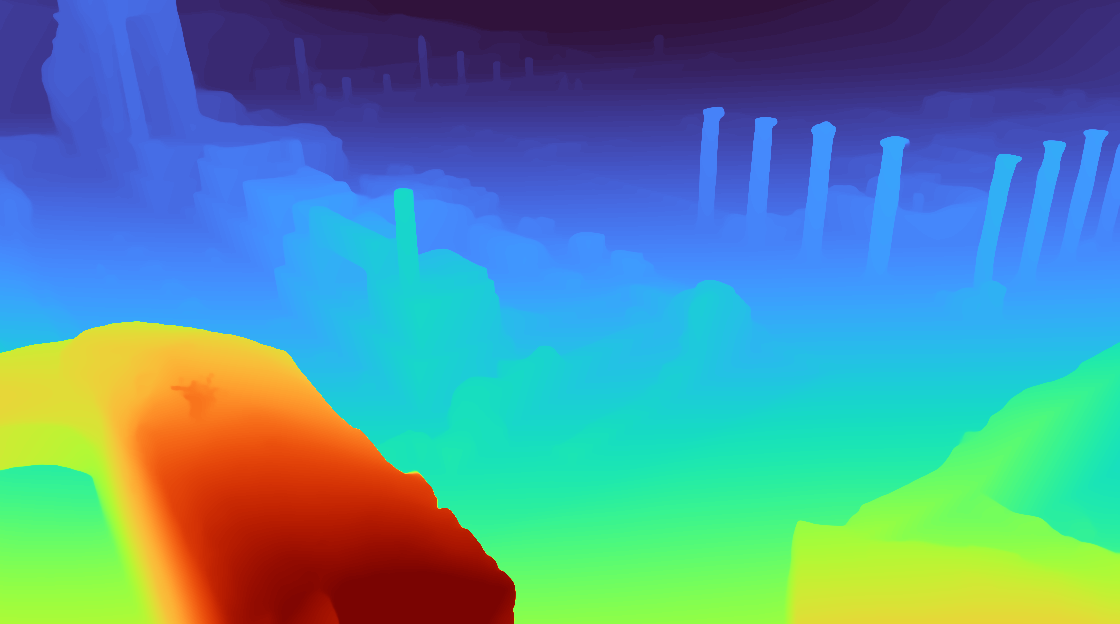}};
\node[thumb, anchor=east] (img-l) at (img-m.west) {\includegraphics[width=4.4cm]{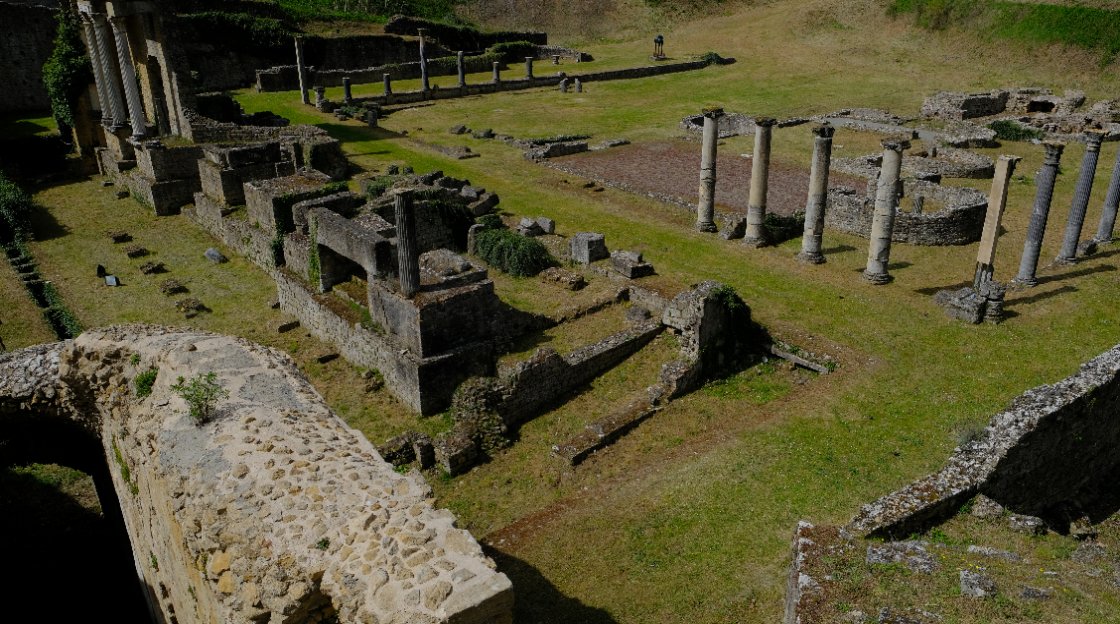}};
\node[thumb, anchor=west] (img-r) at (img-m.east) {\includegraphics[width=4.4cm]{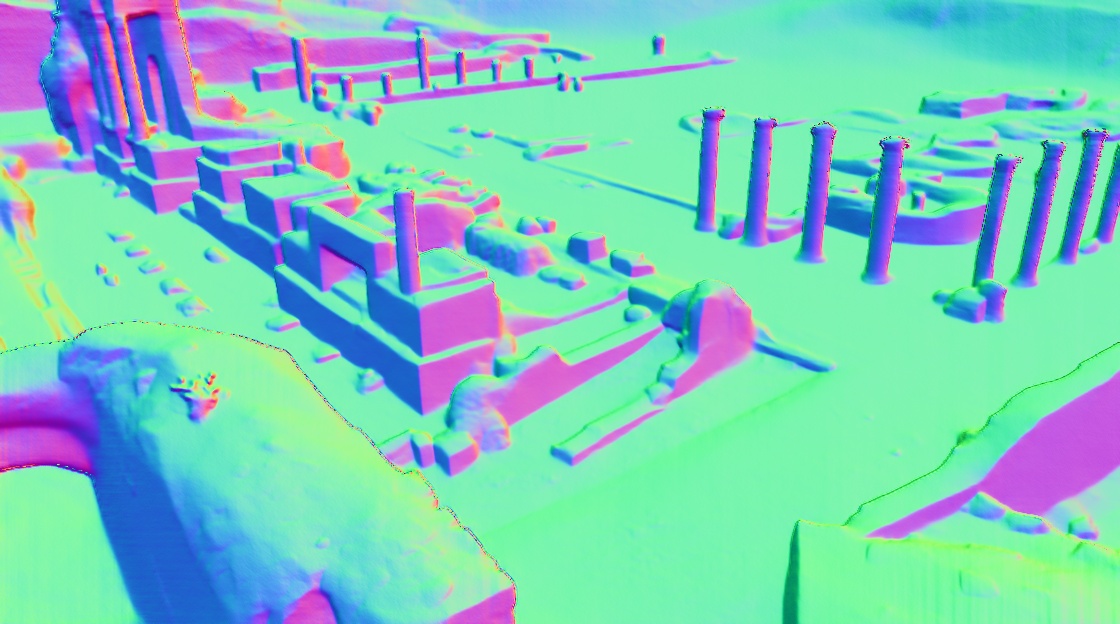}};
\end{tikzpicture}

\vspace{1cm}

\begin{tikzpicture}
\node[thumb, anchor=north] (pcl) at (pcl.south){\includegraphics[width=12cm]{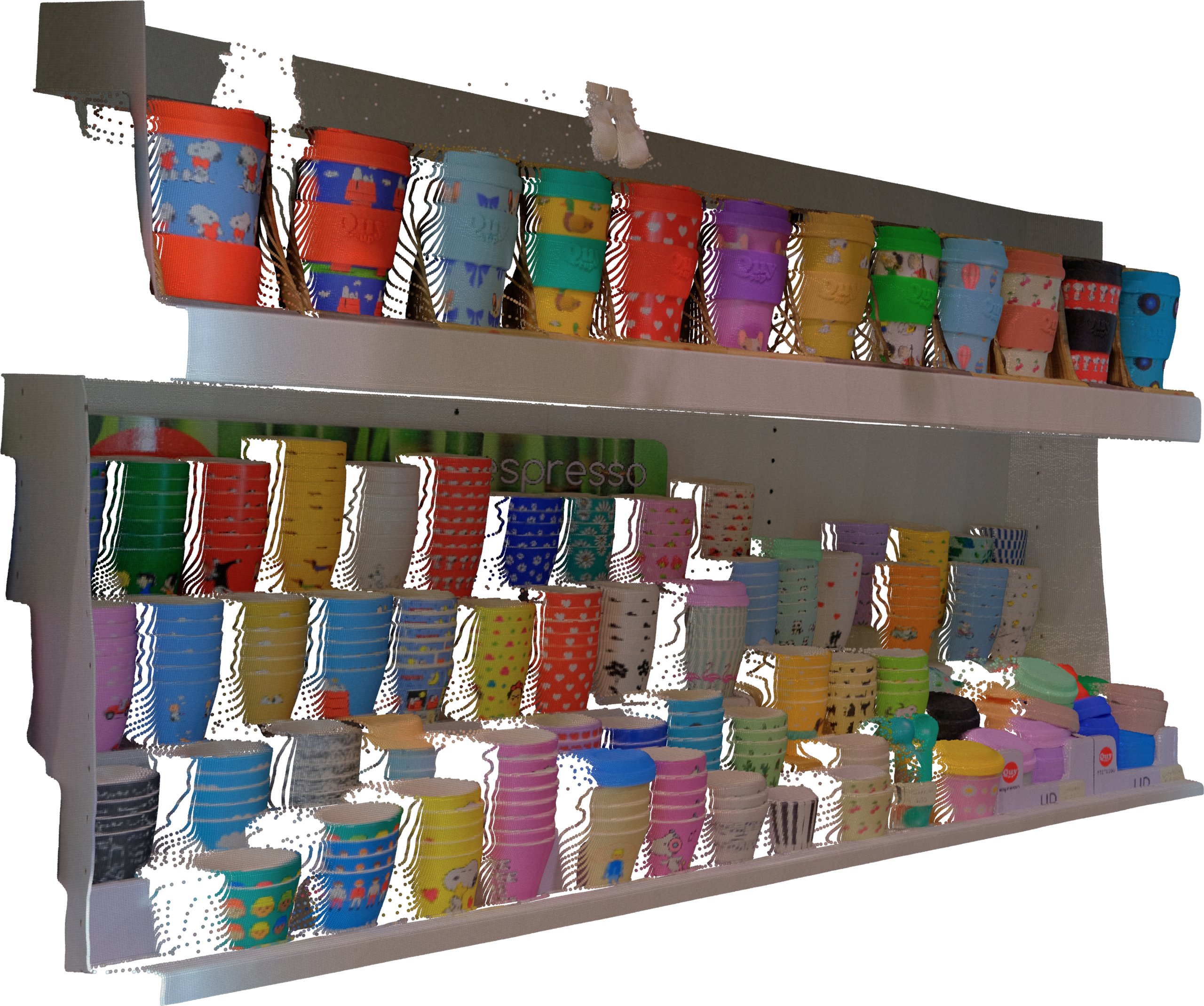}};
\node[thumb, anchor=south] (img-m) at (pcl.north) {\includegraphics[width=4.4cm]{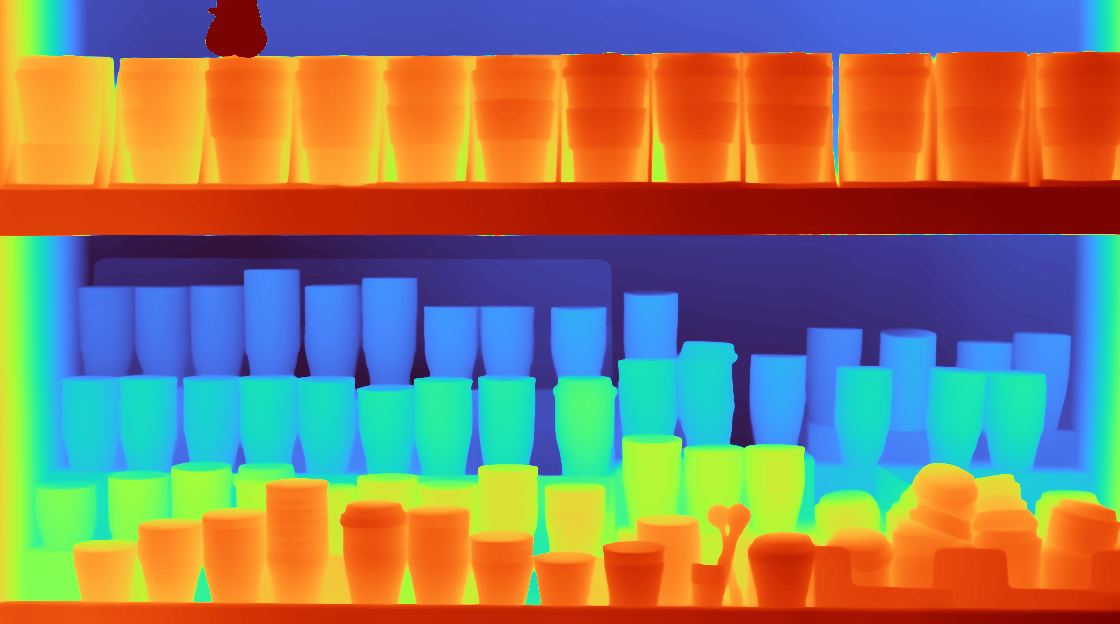}};
\node[thumb, anchor=east] (img-l) at (img-m.west) {\includegraphics[width=4.4cm]{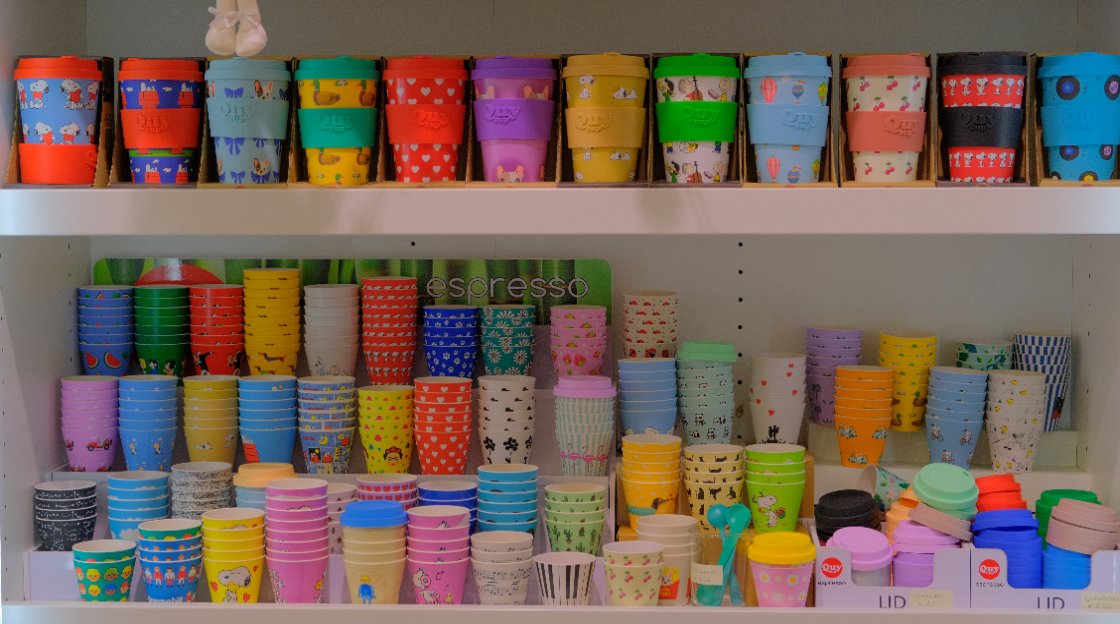}};
\node[thumb, anchor=west] (img-r) at (img-m.east) {\includegraphics[width=4.4cm]{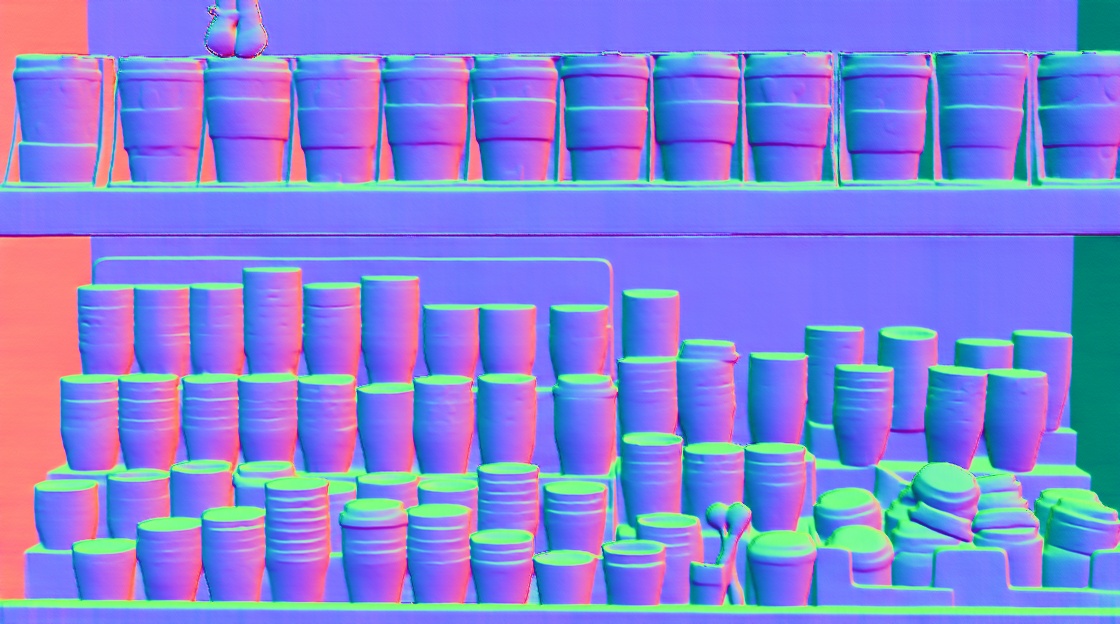}};
\end{tikzpicture}
\caption{\textbf{Additional qualitative results from \ours{}.} Each example includes the RGB input, predicted depth, point map normals, and rendered point map.}
\label{supp:qualitative_voltera_cups}
\end{figure}

\begin{figure}
\centering
\begin{tikzpicture}
\node[thumb] (img-t) {\includegraphics[width=4.55cm]{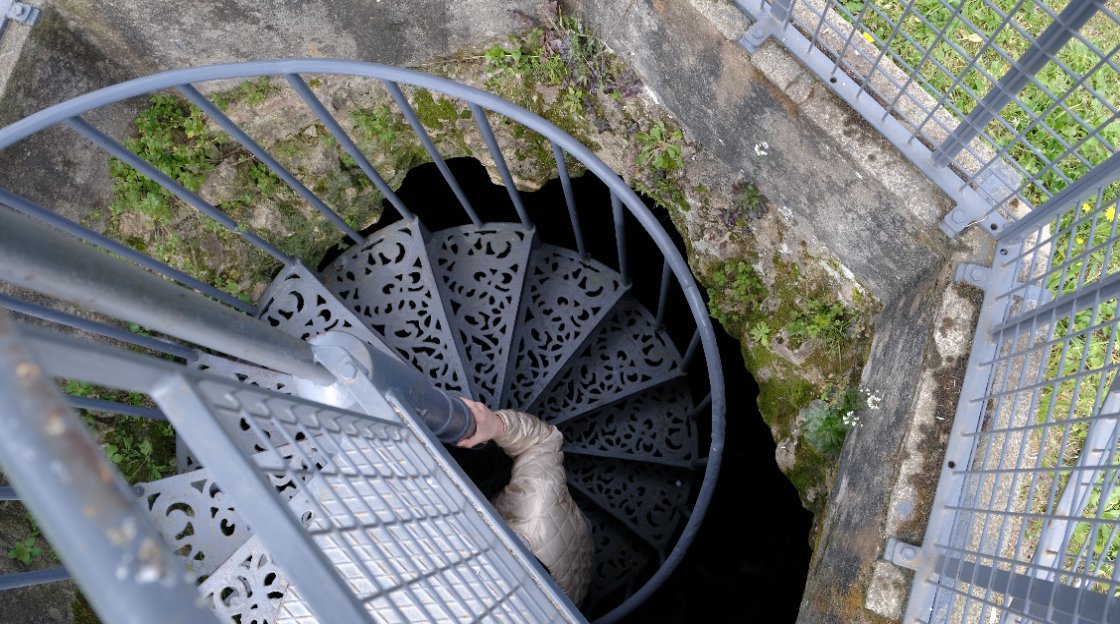}};
\node[thumb, anchor=north] (img-m) at (img-t.south) {\includegraphics[width=4.5cm]{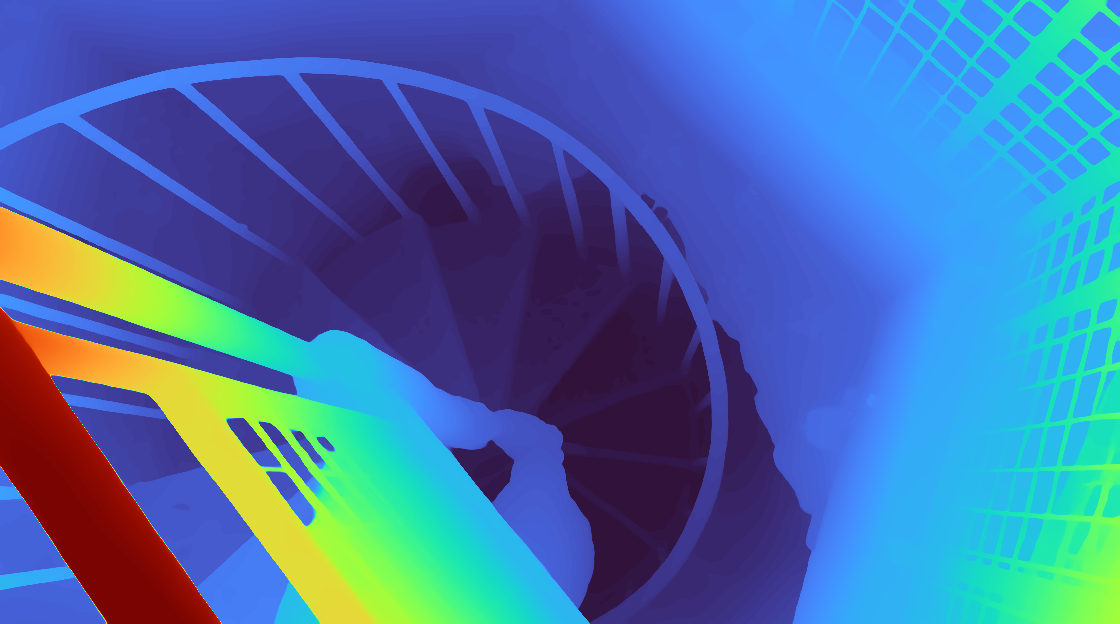}};
\node[thumb, anchor=north] (img-b) at (img-m.south) {\includegraphics[width=4.5cm]{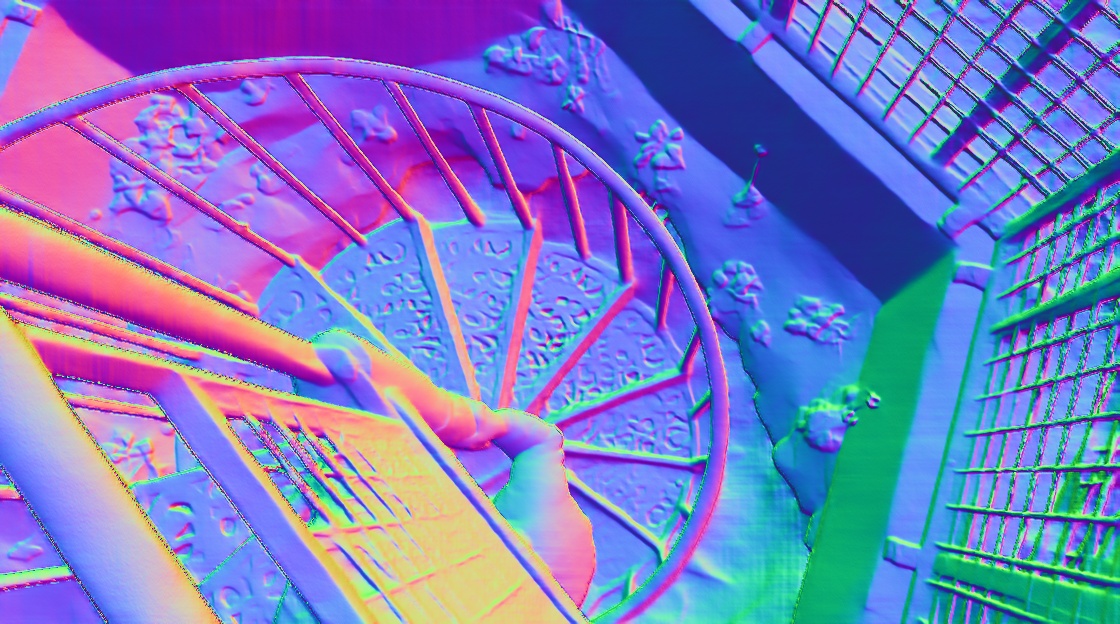}};
\node[anchor=west] (pcl) at (img-m.east) {\includegraphics[width=9cm]{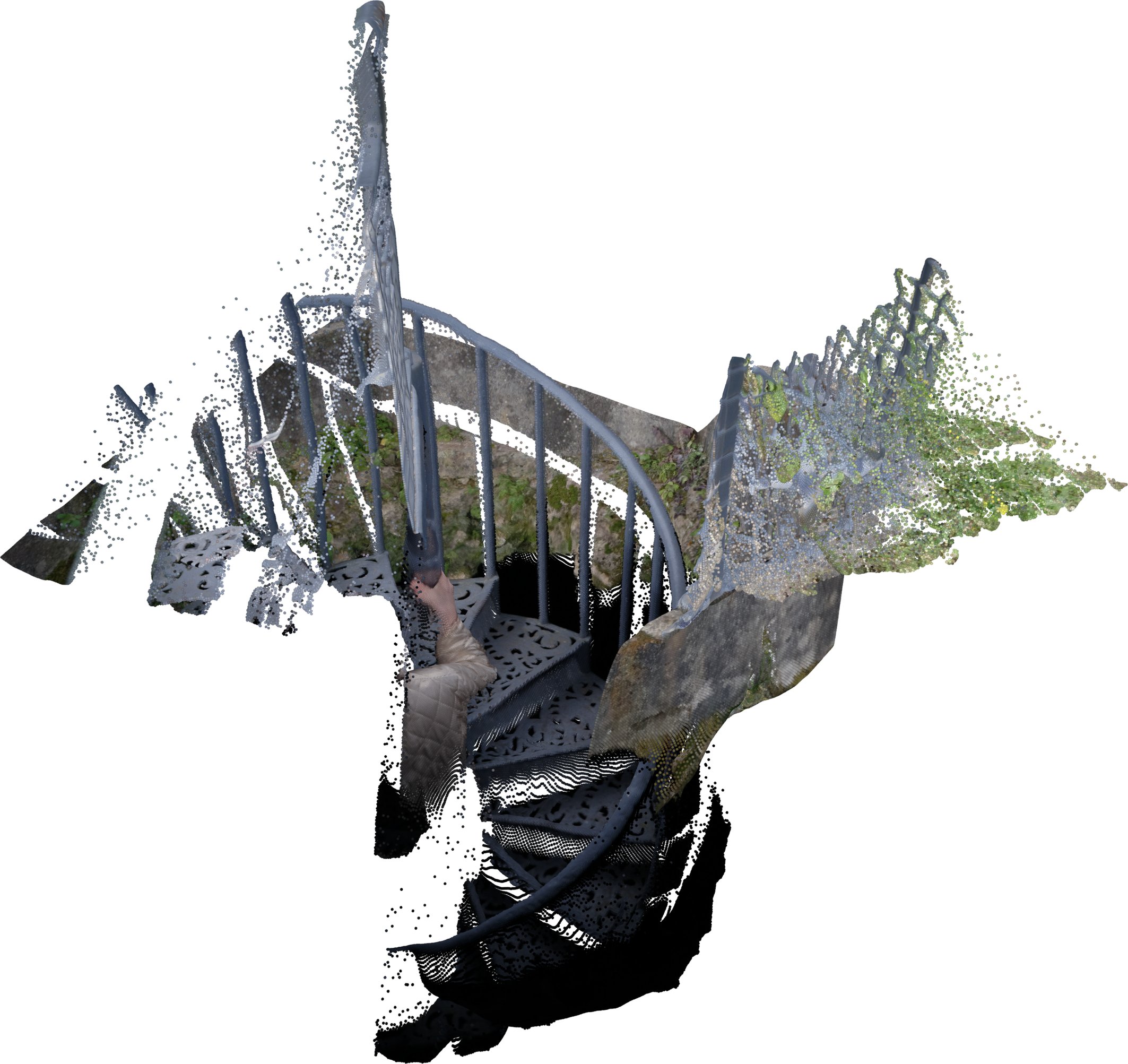}};
\end{tikzpicture}

\begin{tikzpicture}
\node[thumb] (pcl) {\includegraphics[width=10cm]{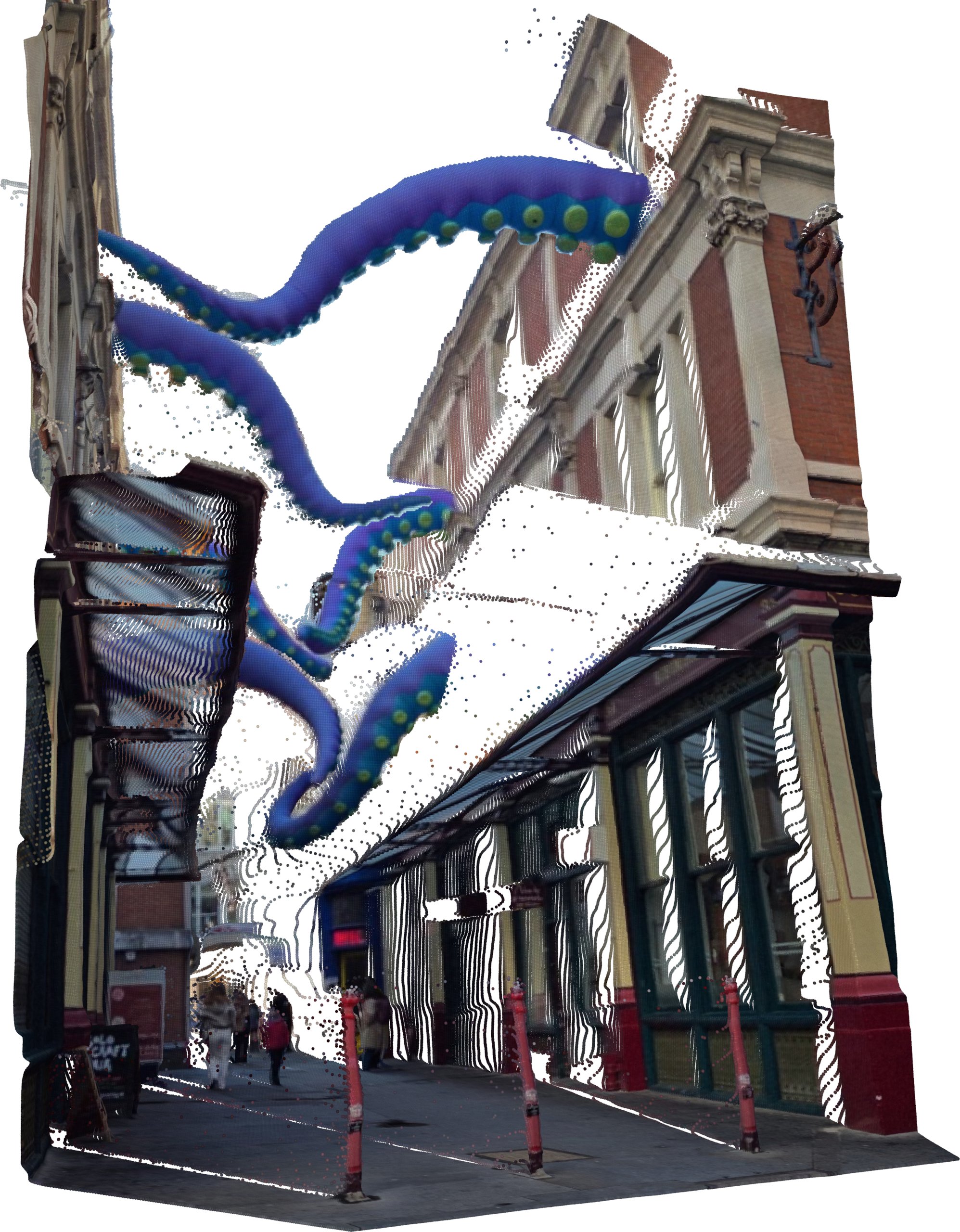}};
\node[thumb, anchor=west] (img-m) at (pcl.east) {\includegraphics[width=3cm]{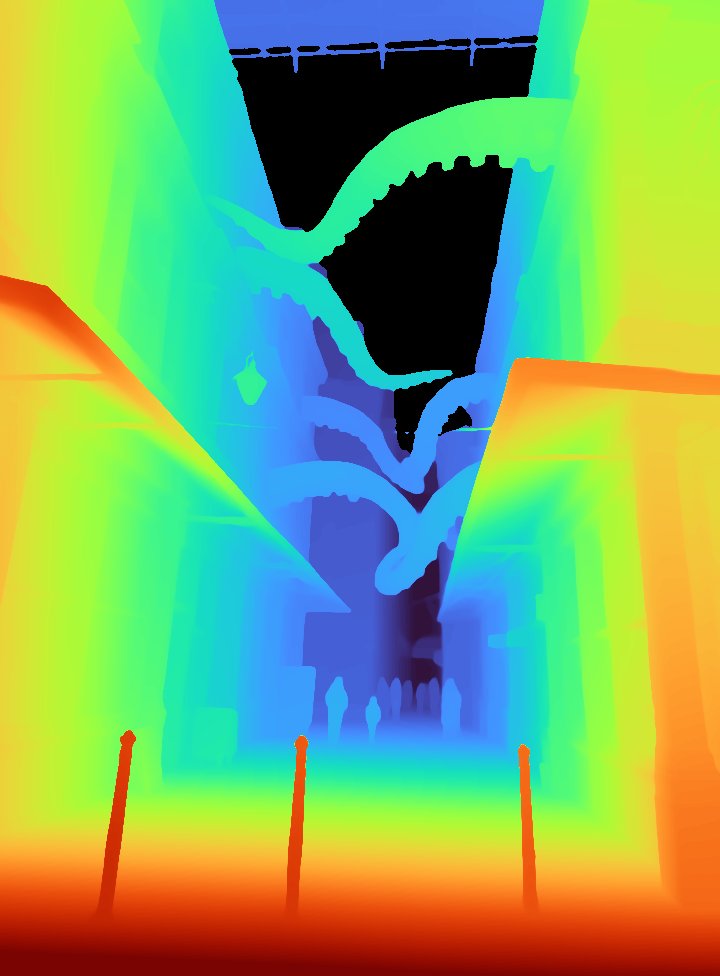}};
\node[thumb, anchor=south] (img-t) at (img-m.north) {\includegraphics[width=3cm]{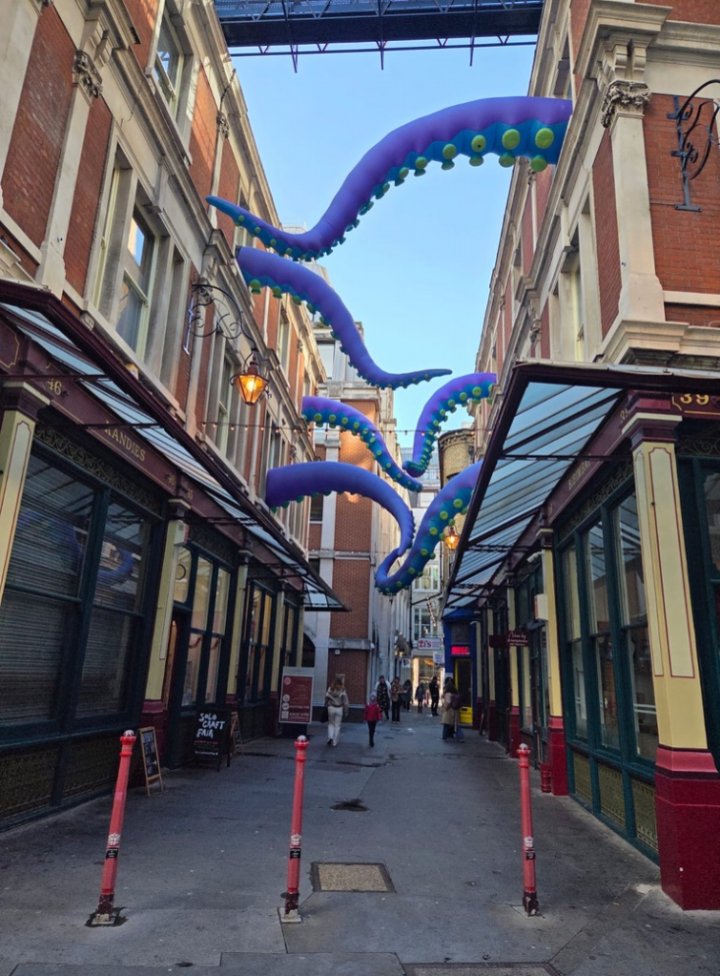}};
\node[thumb, anchor=north] (img-b) at (img-m.south) {\includegraphics[width=3cm]{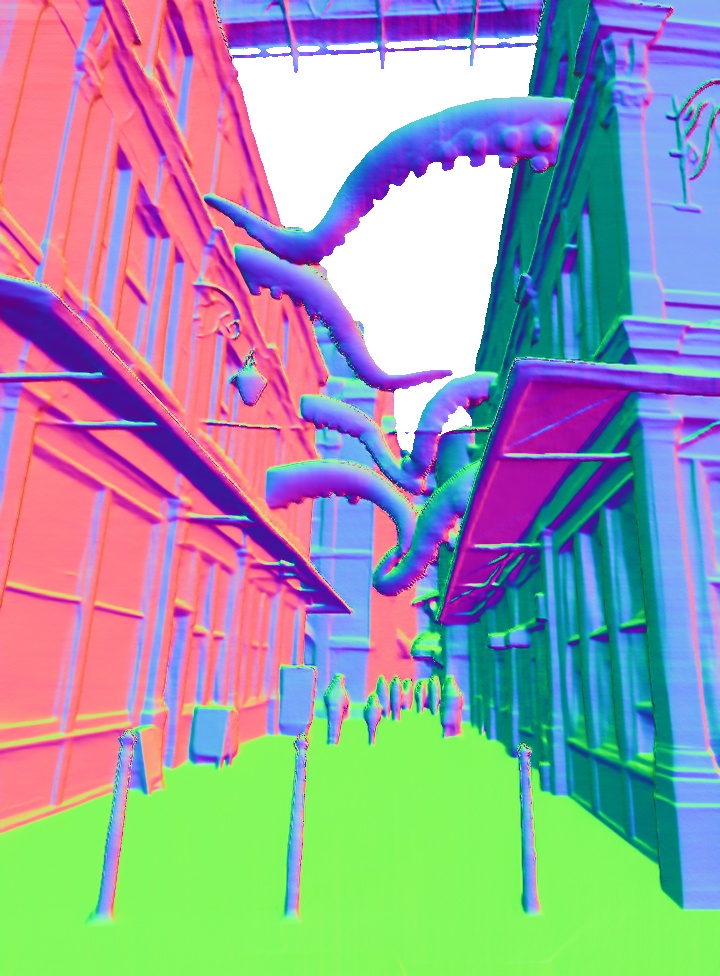}};
\end{tikzpicture}
\caption{\textbf{Additional qualitative results from \ours{}.} Each example includes the RGB input, predicted depth, point map normals, and rendered point map.}
\label{supp:qualitative_stairs_street}
\end{figure}

\begin{figure}
\centering
\begin{tikzpicture}
\node[thumb] (pcl) {\includegraphics[width=10cm]{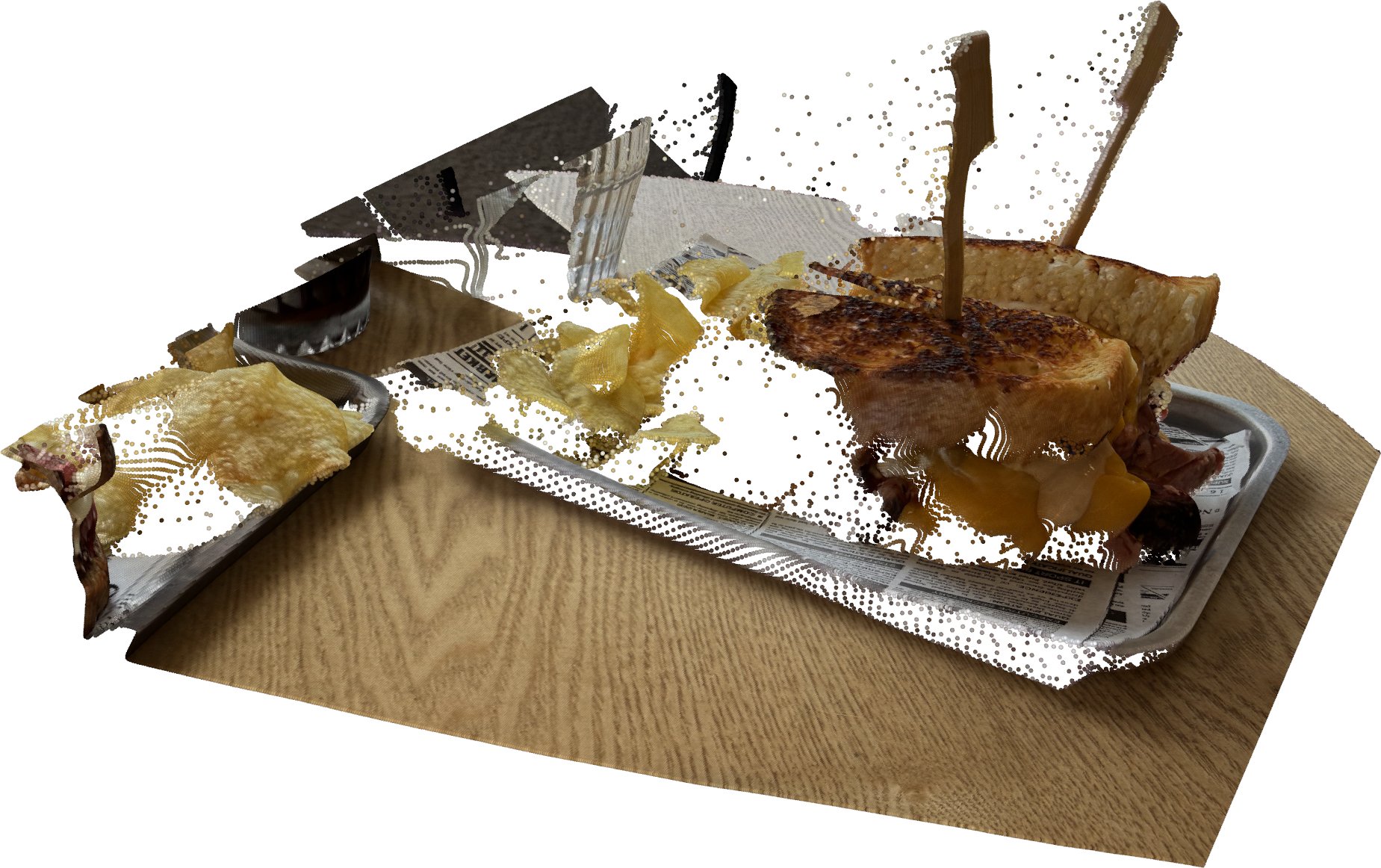}};
\node[thumb, anchor=south] (img-m) at (pcl.north) {\includegraphics[width=4.4cm]{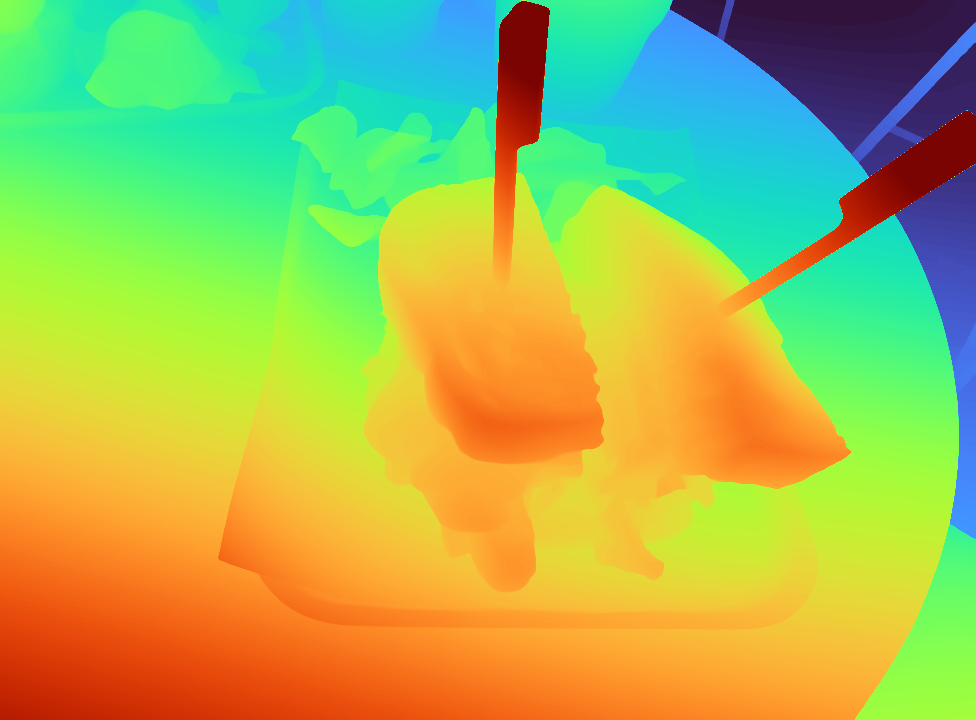}};
\node[thumb, anchor=east] (img-l) at (img-m.west) {\includegraphics[width=4.4cm]{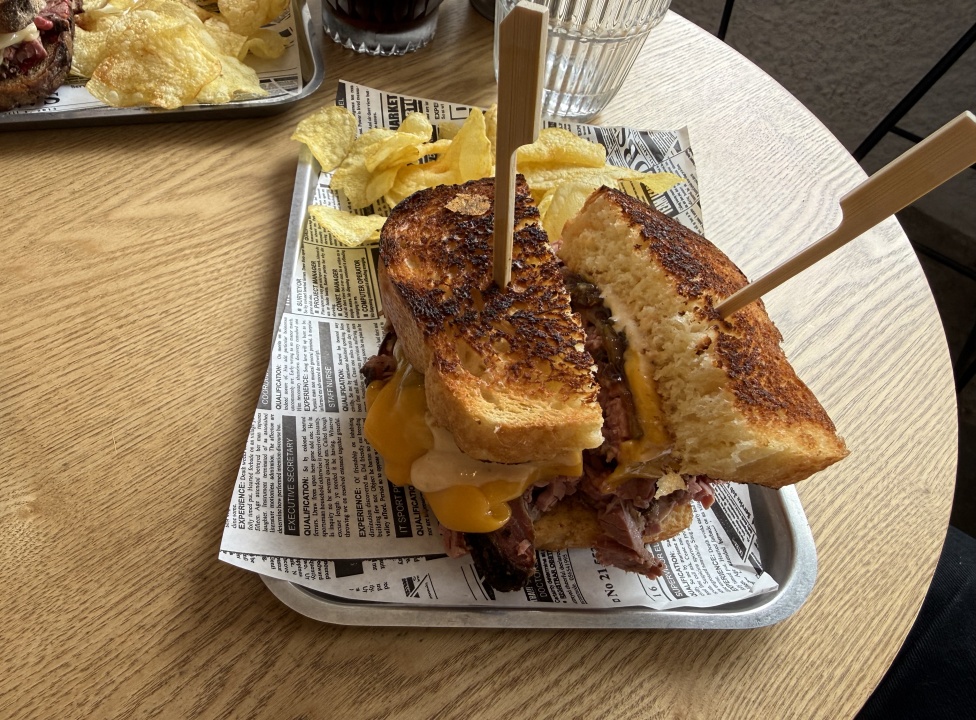}};
\node[thumb, anchor=west] (img-r) at (img-m.east) {\includegraphics[width=4.4cm]{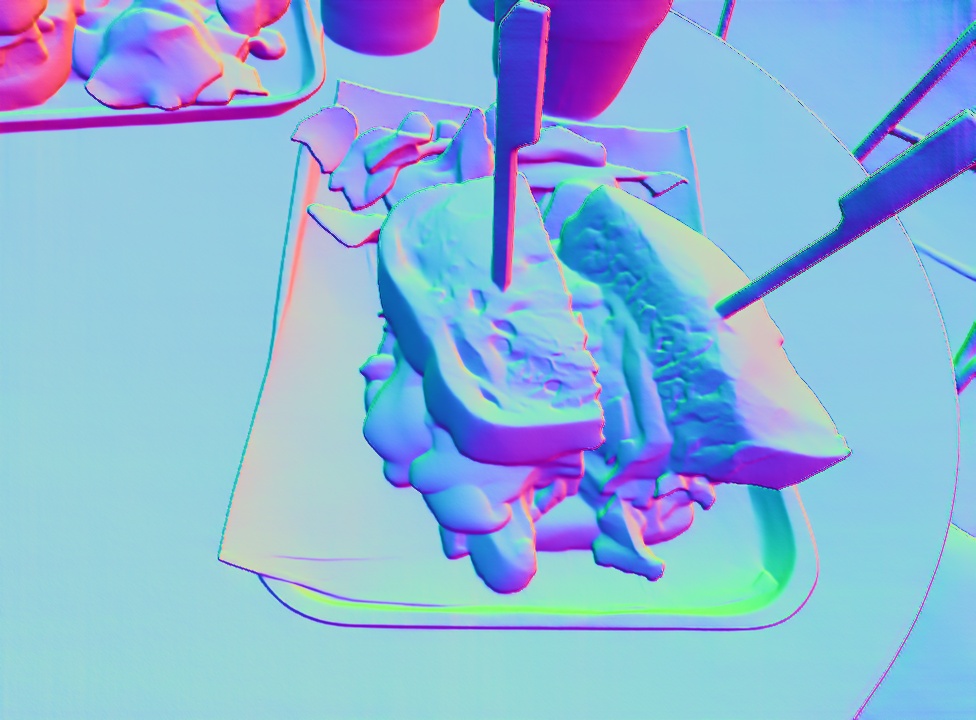}};
\end{tikzpicture}

\vspace{1cm}

\begin{tikzpicture}
\node[thumb, anchor=north] (pcl) at (pcl.south){\includegraphics[width=10cm]{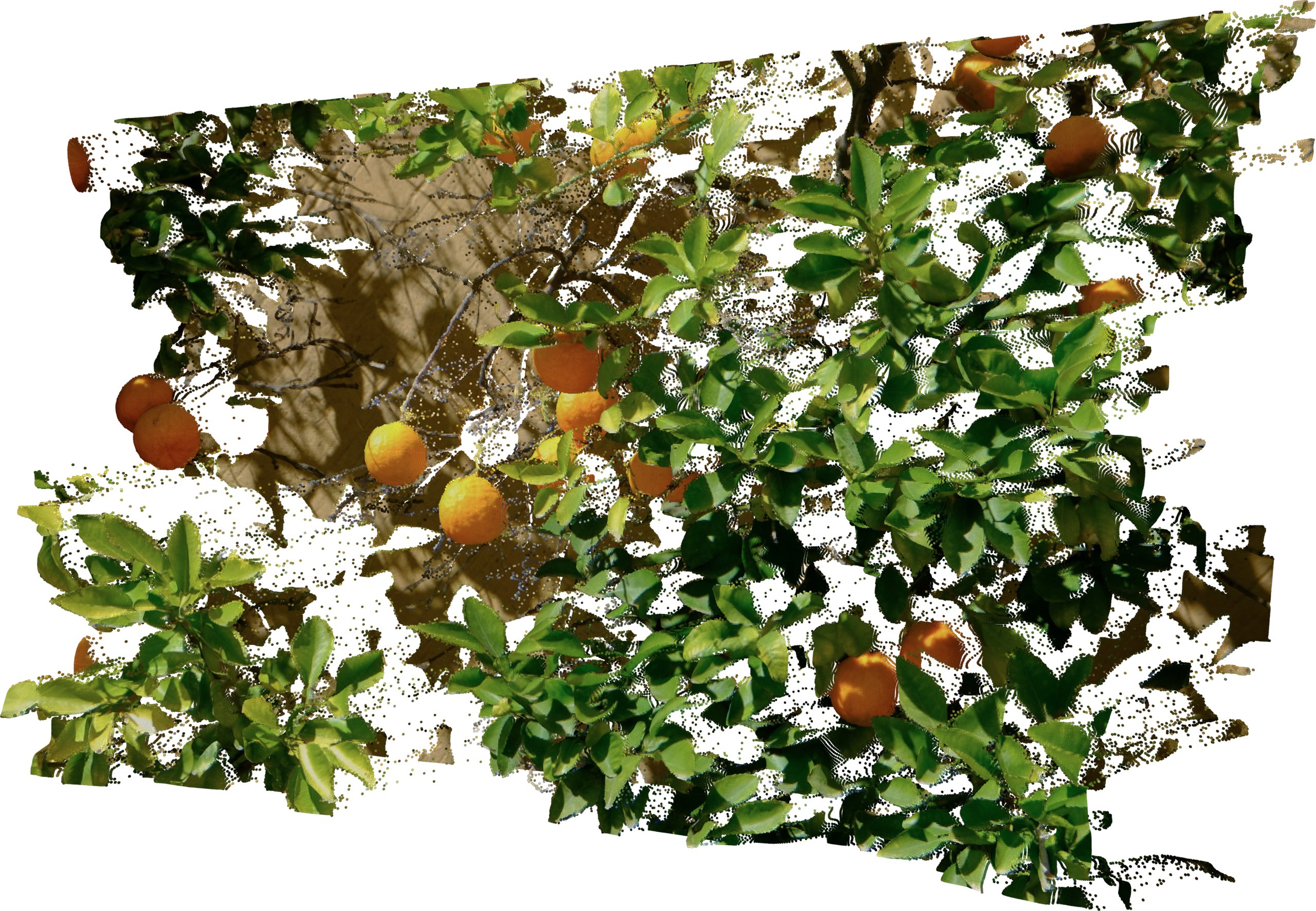}};
\node[thumb, anchor=south] (img-m) at (pcl.north) {\includegraphics[width=4.4cm]{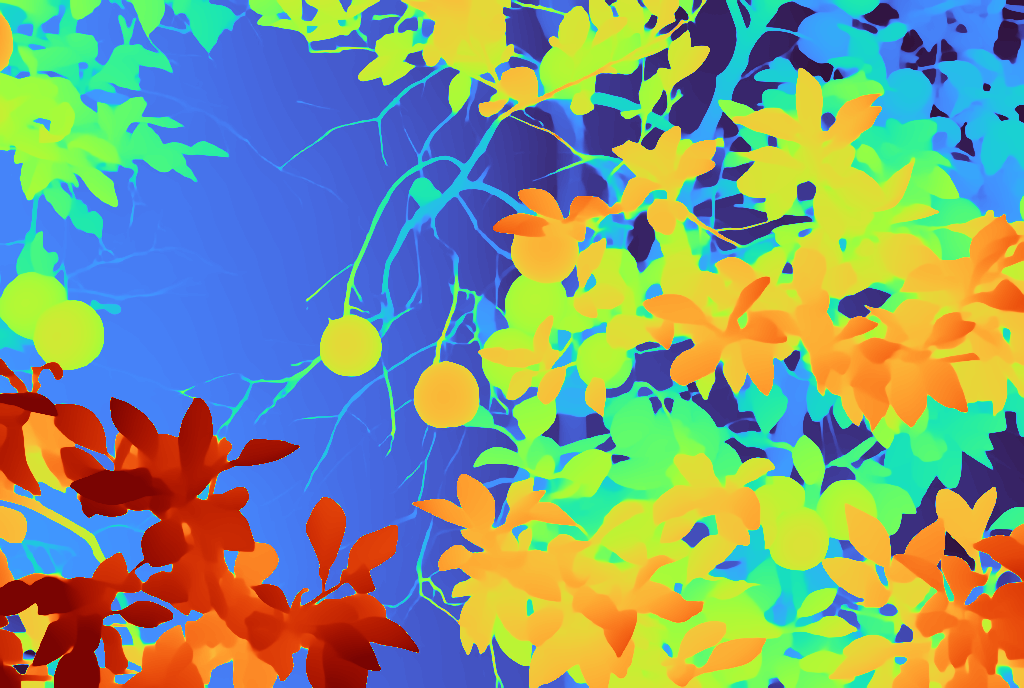}};
\node[thumb, anchor=east] (img-l) at (img-m.west) {\includegraphics[width=4.4cm]{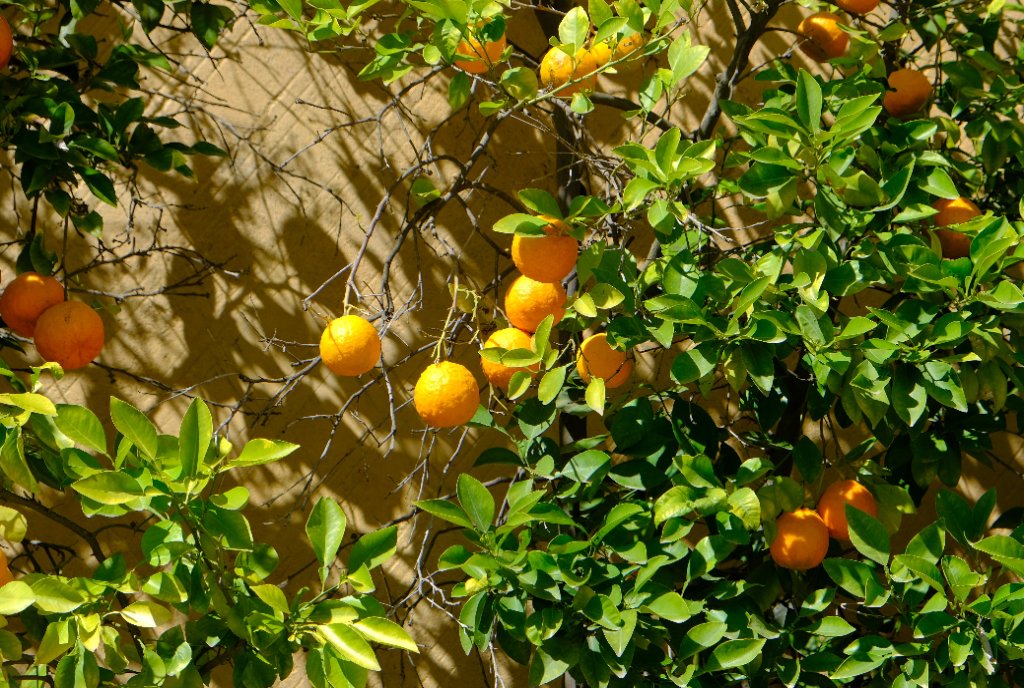}};
\node[thumb, anchor=west] (img-r) at (img-m.east) {\includegraphics[width=4.4cm]{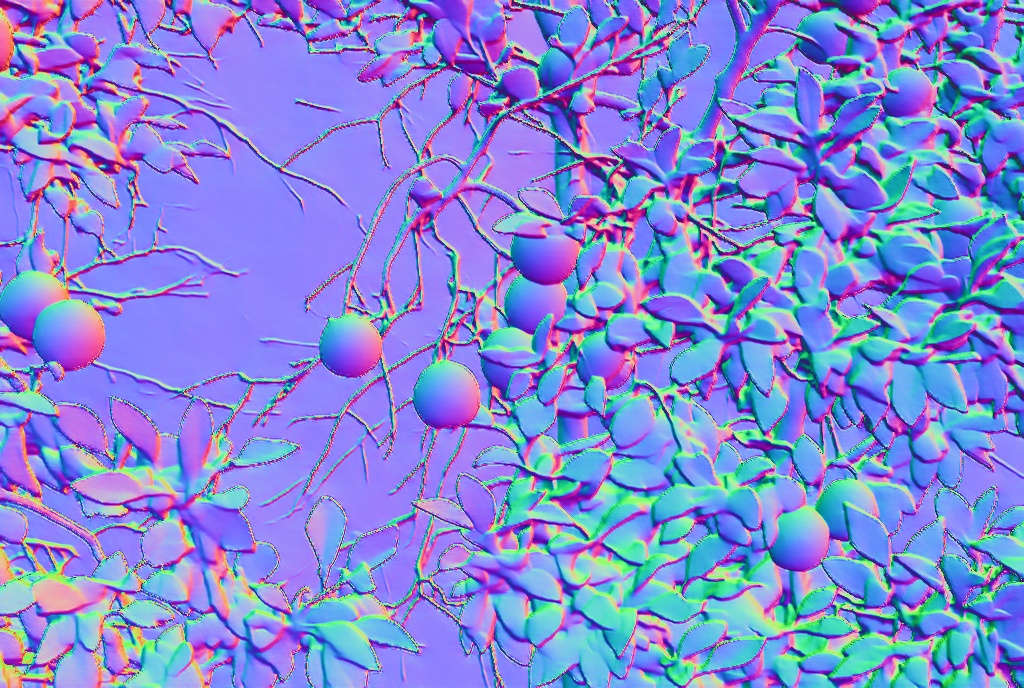}};
\end{tikzpicture}
\caption{\textbf{Additional qualitative results from \ours{}.} Each example includes the RGB input, predicted depth, point map normals, and rendered point map.}
\label{supp:qualitative_sandwich_oranges}
\end{figure}

\end{document}